\documentclass[twoside,11pt]{article}

\usepackage{jmlr2e}
\usepackage{microtype}
\usepackage{graphicx}
\usepackage{subfigure}
\usepackage{booktabs}
\usepackage{wrapfig}
\usepackage{multirow}
\usepackage{amsmath,amsfonts,bm}
\usepackage[algo2e,ruled,vlined]{algorithm2e}
\usepackage{algorithm}
\usepackage{algorithmic}
\usepackage{enumitem}

\newcommand{\KL}{D_{\mathrm{KL}}}
\newcommand{\softmax}{\mathrm{softmax}}
\DeclareMathOperator*{\argmax}{arg\,max}

\newcommand{\E}{\mathbb{E}}

\newcommand{\I}{\mathbb{I}}
\newcommand{\calN}{\mathcal{N}}

\jmlrheading{1}{2000}{1-48}{4/00}{10/00}{meila00a}{Yang Li and Junier B. Oliva}

\ShortHeadings{Distribution Guided Active Feature Acquisition}{Li and Oliva}
\firstpageno{1}

\begin{document}

\title{Distribution Guided Active Feature Acquisition}

\author{\name Yang Li \email yangli95@cs.unc.edu \\
       \addr Department of Computer Science\\
       University of North Carolina at Chapel Hill\\
       Chapel Hill, NC 27514, USA
       \AND
       \name Junier B.\ Oliva \email joliva@cs.unc.edu \\
       \addr Department of Computer Science\\
       University of North Carolina at Chapel Hill\\
       Chapel Hill, NC 27514, USA}

\editor{}

\maketitle

\begin{abstract}
Human agents routinely reason on instances with incomplete and muddied data (and weigh the cost of obtaining further features). In contrast, much of ML is devoted to the unrealistic, sterile environment where all features are observed and further information on an instance is obviated. Here we extend past static ML and develop an active feature acquisition (AFA) framework that interacts with the environment to obtain new information on-the-fly and can: 1) make inferences on an instance in the face of incomplete features, 2) determine a plan for feature acquisitions to obtain additional information on the instance at hand. We build our AFA framework on a backbone of understanding the information and conditional dependencies that are present in the data. First, we show how to build generative models that can capture dependencies over arbitrary subsets of features and employ these models for acquisitions in a greedy scheme. After, we show that it is possible to guide the training of RL agents for AFA via side-information and auxiliary rewards stemming from our generative models. We also examine two important factors for deploying AFA models in real-world scenarios, namely interpretability and robustness. Extensive experiments demonstrate the state-of-the-art performance of our AFA framework.
\end{abstract}


\section{Introduction}
Truly robust intelligence is a far cry from simple machine pattern detection; instead, robust intelligent systems are expected to make critical decisions with incomplete and uncertain data. Moreover, they are expected to weigh the cost of obtaining further information to improve their predictions and decisions. Humans routinely use these capabilities to navigate decisions. For instance, doctors often face a critical decision when diagnosing a patient: diagnose that the patient has/does not have a certain disease based on the information at hand; or obtain additional information before making a diagnosis. Such decisions must weigh the cost of an incorrect prediction (a false positive or false negative) and the cost (in time, risk, or money) of obtaining additional information. Incongruently to the needs of many real-world applications, much of machine learning (ML) is devoted to the unrealistic, sterile environment where all features are observed and further information on an instance is obviated. I.e., much of machine learning is focused on developing models for predictions that are passively given all features. As a consequence, the current paradigm in ML is unprepared for the future of automation, which will be highly interactive and able to obtain further information from the environment when making predictions. The goal of this work is to equip machines with the capabilities of \emph{interacting with the environment to obtain new information on-the-fly as it is making decisions}.

As a motivating example, let us reconsider a doctor making a diagnosis on a patient (an instance). The doctor usually has not observed all the possible measurements (such as blood samples, x-rays, etc.) from the patient. The doctor is also not forced to make a diagnosis based on the currently observed measurements; instead, she may dynamically decide to take more measurements to help determine the diagnosis. The next measurement to make (feature to observe), if any, will depend on the values of the already observed features; thus, the doctor may determine a different set of features to observe from patient to patient (instance to instance) depending on the values of the features that were observed. For example, a low value from a blood test may lead a doctor to ask for a biopsy, whereas a high value may lead to an MRI. Hence, each patient will not have the same subset of features selected (as would be the case with typical feature selection). Furthermore, acquiring features typically involves some cost (in time, money and risk), and intelligent systems are expected to automatically balance between cost and improvement in performance. 

As a second motivating example, consider an automated education system that is assessing a student’s grasp on a particular topic (e.g., algebra) as quantified via a score on a future comprehensive test. The automated teacher has access to a large bank of questions that it may ask in order to assess the student (the instance). A typical ML approach would be to collect all the available features for a student (scores on each question in the bank) and use a classifier on the vector of scores to predict the future grade on the comprehensive test. This approach may prove quite expensive and wasteful, however, since the student is required to answer (and be graded on) all possible bank questions. Instead, a more efficient route is to dynamically choose the next question(s) to ask based on the student’s score on previous questions. The ideal system would be able to personalize the questions asked based on the particular student’s history of responses with minimal supervision (e.g., without having to label sub-topics covered in a question).

In this work, we study how to sequentially acquire informative features at inference time through a backbone that is guided by an understanding of conditional dependencies among features. The conditional dependencies among features play a vital role in assessing what new information would be most useful in light of what is currently known (what has been observed) about an instance. E.g., if one is \emph{certain} of the value of an unobserved feature (a candidate for acquisition) given what is currently observed (previously acquired features), then it would be wasteful to pay the cost to acquire that feature. Similarly, if the acquisition of an unobserved feature is unlikely to update (improve) one's prediction over the target output, then it would also be wasteful to pay the cost to acquire that feature. 

We develop a distribution guided approach to actively acquiring features as follows. In Section~\ref{sec:problem}, we formulate the problem and describe the challenges and requirements for solving this problem. In Section~\ref{sec:acflow}, we model arbitrary conditional distributions among subsets of features, which underpin much of the AFA problem. Based on the arbitrary conditionals, we derive a greedy acquisition policy in Section~\ref{sec:afa_greedy}. Section~\ref{sec:afa_rl} formulates the AFA problem as a Markov Decision Process (MDP) and proposes to solve it using reinforcement learning (RL) approaches that are guided by auxiliary information and rewards from a generative surrogate model. We also develop an action space grouping scheme and propose a hierarchical acquisition policy to help the AFA agent deal with a potentially large number of candidate features. Section~\ref{sec:explain_afa} and \ref{sec:robust_afa} further extend the AFA framework to encourage explainability of the acquisition process and robustness to out-of-distribution (OOD) instances respectively. In Section~\ref{sec:related_works}, we review the related works. Section~\ref{sec:implement} presents some important implementation details. Section~\ref{sec:experiments} elaborates the experimental setting and presents the numerical results. In Section~\ref{sec:conclusion}, we conclude our work.

This work is an extension of our conference paper ``Active Feature Acquisition with Generative Surrogate Models'' \citep{li2021active}, but with substantial improvements. First, in order to scale the AFA model for high-dimensional data, we develop a hierarchical acquisition policy based on an internal clustering of candidate features. The action space grouping aids the agent through the high dimensional action space thus assisting exploration and scalability. Second, we study the interpretability of the acquisitions and propose a goal based acquisition policy for better explainable acquisitions. The acquisition sequences can be explained by the goal they strive to achieve. Third, we explore the cases where out-of-distribution instances may be encountered when performing active feature acquisition. We develop an OOD detection model that can deal with partially observed instances. The detector is then utilized to encourage the AFA model to acquire robust features.

\begin{figure*}
    \centering
    \subfigure[Supervised AFA]{
    \includegraphics[width=0.48\linewidth]{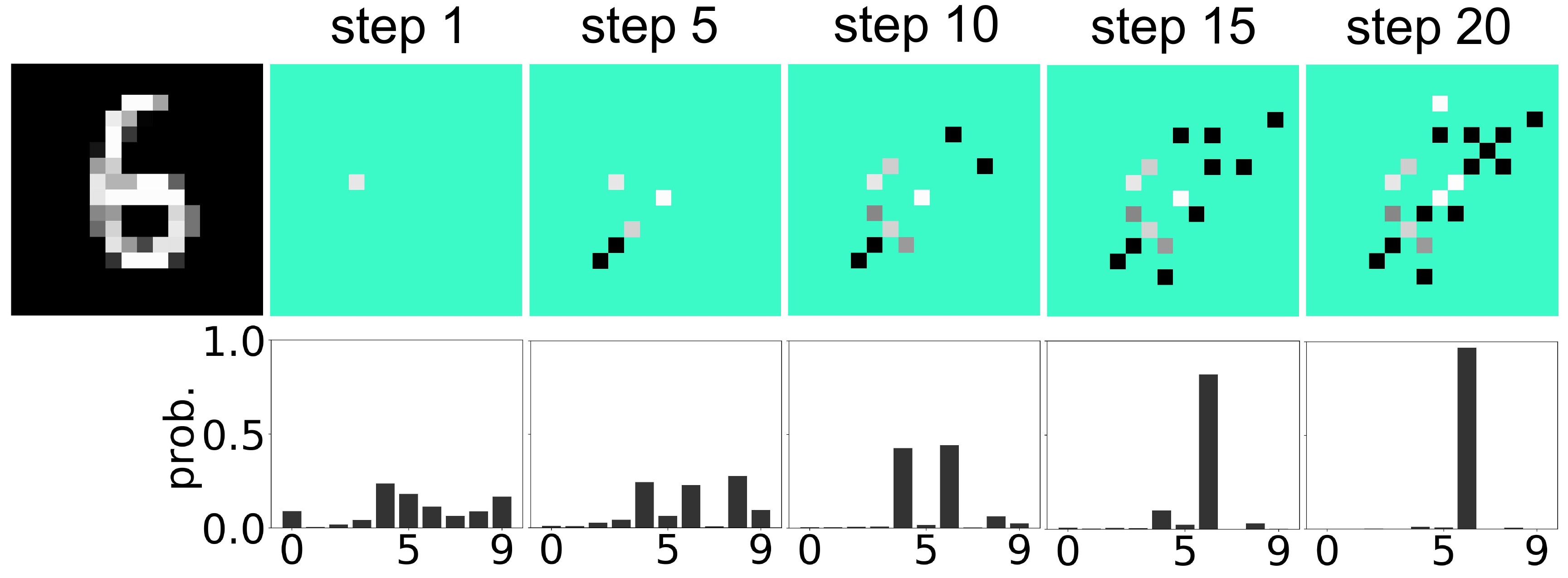}}
    \hfill
    \subfigure[Unsupervised AFA]{
    \includegraphics[width=0.48\linewidth]{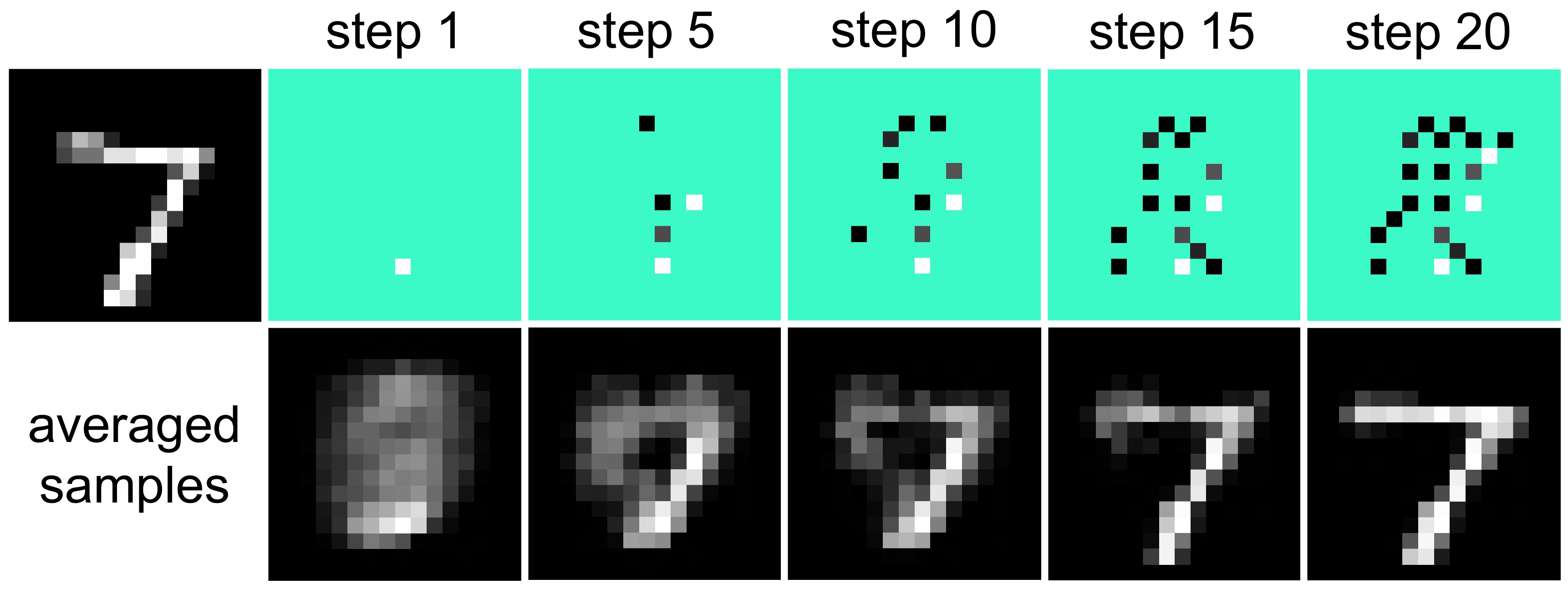}}
    \vspace{-5pt}
    \caption{Acquisition process for supervised and unsupervised AFA tasks. Top: acquisition process where one pixel value is acquired at each step and the green masks indicate the unobserved features. Bottom: the prediction probabilities and averaged inpaintings for supervised and unsupervised tasks respectively.}
    \label{fig:afa_examples}
    \vspace{-15pt}
\end{figure*}

\section{Problem Formulation}\label{sec:problem}
In this section, we formulate the active feature acquisition problem from a cost sensitive perspective, where acquiring the feature values incurs some cost. Our goal is to achieve the best possible performance for the downstream task while minimizing the cost of feature acquisition.

\paragraph{AFA for Supervised Tasks}
Consider a discriminative task with features $x \in \mathbb{R}^d$ and target $y$. Instead of predicting the target by first collecting all the features, we let the model determine what features to acquire for a given instance. There is typically a cost associated with features and the goal is to maximize the task performance while minimizing the acquisition cost, i.e.,
\begin{equation}\label{eq:goal_sup}
    \underset{o \subseteq \{1,\ldots,d\}}{\text{minimize}}~ \mathcal{L}(\hat{y}(x_o), y) + \alpha\, \mathcal{C}(o),
\end{equation}
where $\mathcal{L}(\hat{y}(x_o), y)$ represents the loss function between the prediction $\hat{y}(x_o)$ and the target $y$. Note that the prediction is made with the acquired feature subset $x_o, o \subseteq \{1,\ldots,d\}$. Since we expect the model to acquire different features for different instances, the model should be able to predict with arbitrary subsets $\hat{y}(x_o)$. $\mathcal{C}(o)$ represents the cost of the acquired features $o$. The hyperparameter $\alpha$ controls the trade-off between prediction loss and acquisition cost.

\paragraph{AFA for Unsupervised Tasks}
In addition to the supervised case, where the goal is to acquire new features to predict a target variable $y$. In some cases, however, there may not be a single target variable, but instead, the target of interest may be the remaining unobserved features themselves. That is, rather than reducing the uncertainty with respect to some desired output response (that cannot be directly queried and must be predicted), we now pursue an unsupervised case, where the task is to query as few features as possible that allow the model to correctly uncover the remaining unobserved features. Similarly, the goal can be formulated as 
\begin{equation}\label{eq:goal_unsup}
    \underset{o \subseteq \{1,\ldots,d\}}{\text{minimize}}~ \mathcal{L}(\hat{x}(x_o), x) + \alpha\, \mathcal{C}(o),
\end{equation}
where $\hat{x}(x_o)$ represents the reconstructed features based on the observed ones.

\paragraph{Dynamic Sequential Acquisition}
Note that a (retrospective) minimization of the objectives \eqref{eq:goal_sup} and \eqref{eq:goal_unsup} requires knowledge of all features of $x$ (consider, for example, an exhaustive search over subsets $o \subseteq \{1,\ldots,d\}$). Instead, we wish to minimize of the objectives \eqref{eq:goal_sup} and \eqref{eq:goal_unsup} \emph{whilst} only ever acquiring (and having access to) a small number of the feature-values $x_1, \ldots, x_d$. That is, the final selected subset, $o$, must itself be a function of only the values in $x_o$ (since having $o$ be a function of features in $\{1, \ldots, d\} \setminus o$ implies having access to features that have not been `acquired'.
Thus, without loss of generality, we consider a sequential acquisition of features, where the next feature (index) to acquire may depend on the feature value of the previously acquired features. (Please refer to Fig.~\ref{fig:afa_examples} for an illustration of AFA for both supervised and unsupervised tasks.) 
By virtue of acquiring features sequentially, the later acquisitions depend on information gathered from earlier acquisitions. Hence, this allows for the subset that is selected to depend on feature values, but only for those feature values that are acquired.


\paragraph{Contrast to Active Learning} Below, we contrast AFA with two conceptually similar tasks, beginning with active learning.
Active learning \citep{fu2013survey,konyushkova2017learning,yoo2019learning} is a related approach in ML to gather more information when a learner can query an oracle for the true label, $y$, of a complete feature vector $x \in \mathbb{R}^d$ to build a better estimator. However, AFA considers queries to the environment for the feature value corresponding to an unobserved feature dimension, $i$, in order to provide a better prediction on the current instance. Thus, while the active learning paradigm queries an oracle \emph{during training} to build a classifier with complete features, AFA paradigm queries the environment \emph{at evaluation} to obtain missing features of a current instance to help its current assessment.

\paragraph{Contrast to Feature Selection} Next, we contrast AFA to traditional feature selection.
Feature selection \citep{miao2016survey,li2017feature,cai2018feature} ascertains a static subset of important features to eliminate redundancies, which can help reduce computation and improve generalization. Feature selection methods choose a \emph{fixed} subset of features $s \subseteq \{1,\dots, d\}$, and always predict $y$ using this same subset of feature values, $x_s$. 
Thus, feature selection can be seen as a special case of AFA (perhaps a stubborn one) that selects the same next features to observe regardless of the previous feature values it has encountered.
In contrast, AFA considers a \emph{dynamic} subset of features that are sequentially chosen and personalized on an instance-by-instance basis to increase useful information. 
A dynamic strategy will be helpful whenever features are indicative of other features, which is often the case in real-world examples\footnote{A simple case that illustrates the dominance of a dynamic acquisition strategy is as follows. Suppose that there are $d-1$ independent features $x_i$ for $i\in \{1, \ldots, d-1\}$, and one ``guiding'' feature, $x_d$, which decides what independent feature determines the output $y$. E.g., if $x_d$ is in some range then $y$ is a function of only $x_1$, if $x_d$ is in another range then $y$ is a function of $x_2$, etc. If all independent features are used with equal probability, then a typical feature selection algorithm will have to select all $d$ to obtain perfect accuracy (since it must use a fixed subset). In contrast, by dynamically selecting features, one can observe only two features: $x_d$, and the corresponding independent feature.}.
It is worth noting that AFA may be applied after a feature selection preprocessing step to reduce the search space.

\section{Modeling Arbitrary Conditionals}\label{sec:acflow}
As previously noted, the conditional dependencies among features play a crucial role in the AFA task. The conditional distributions between the already acquired features, $x_o$, and missing features $x_u$, $p(x_u \mid x_o)$ ($u \cap o = \emptyset$), is informative of dependencies that may be exploited. For instance, if missing features may be predicted with high certainty, then acquiring them would be wasteful. Moreover, the dependencies between the target label and features, $p(y \mid x_o, x_u)$ can indicate if missing features will not be informative to $y$ (and therefore are not worth acquiring). In this work, we propose to explicitly model these conditional dependencies among features to aid AFA approaches. Below, 
we propose a framework, arbitrary conditioning flow models (ACFlow) \citep{li2019flow}, to construct generative models that yield tractable (analytically available) conditional likelihoods $p(x_u \mid x_o)$ of an arbitrary subset of covariates, $x_u$, given the remaining observed covariates $x_o$ (where $u, o \subseteq \{1,\ldots,d\}$ and $u \cap o = \emptyset$).

Modeling the dependencies of \emph{arbitrary} subsets of observed and unobserved features ($o$ and $u$, respectively) is necessary for AFA. This is because features shall be dynamically acquired in an instance-by-instance basis. That is, the features that are acquired may vary greatly from one instance to another. Therefore, one must be able to provide conditional dependencies for a wide set of possibly observed features. This, however, poses several challenges.
A majority of generative modeling approaches are focused solely on the joint distribution $p(x)$ and utilize models where it is intractable to obtain the conditional distribution of some arbitrary subset of features $x_u$ given the rest of the observed covariates $x_o$, $p(x_u \mid x_o)$, due to intractable marginalization.
Therefore, they are opaque in the conditional dependencies that are carried among subsets of features. 
Existing conditional generative models are mostly conditioned on a \emph{fixed} set of covariates, such as class labels \citep{kingma2018glow}, image descriptions \citep{van2016conditional}, or other data points \citep{li2019forest,norouzi2020exemplar}.
Learning a separate model for each different subset of observed covariates quickly becomes infeasible as it requires an exponential number of models with respect to the dimensionality of the input space.
However, capturing the variability in the domains and dependencies covered across different conditionals $p(x_u \mid x_o)$ in a single model is challenging.
For example, some subsets $u, o$ may lead to unimodal distribution with linear dependencies, while other subsets may lead to multi-modal distributions with non-linear dependencies.
Furthermore, the dimensionality of $x_u$ and $x_o$ will vary, and dealing with arbitrary dimensionality is a largely unexplored topic in current generative models.  
Here we develop ACFlow \citep{li2019flow}, a method that is capable of estimating \emph{all} conditional distributions $p(x_u \mid x_o)$ (for arbitrary $x_u$ and $x_o$) via tractable conditional likelihoods with a single model.

ACFlow builds on Transformation Autoregressive Networks (TANs) \citep{oliva2018transformation}, a flow-based model that combines transformation of variables with autoregressive likelihoods. 
The change of variable theorem \eqref{eq:changevariables} is the cornerstone of flow-based generative models, where $q$ represents an invertible transformation that transforms covariates from input space $\mathcal{X}$ into a latent space $\mathcal{Z}$.
\begin{equation}\label{eq:changevariables}
    p_{\mathcal{X}}(x) = \left| \det \frac{dq}{dx} \right| p_{\mathcal{Z}}(q(x))
\end{equation}
Typically, a flow-based model transforms the covariates to a latent space with a simple base distribution, like a standard Gaussian \citep{dinh2014nice,dinh2016density,kingma2018glow,JainiSY19,durkan2019neural,papamakarios2021normalizing}. However, TANs provide additional flexibility by modeling the latent distribution with an autoregressive approach \citep{larochelle2011neural}. This alters the earlier equation \eqref{eq:changevariables}, in that $p_{\mathcal{Z}}(q(x))$ is now represented as the product of $d$ conditional distributions.
\begin{equation}\label{eq:autoreg}
    p_{\mathcal{X}}(x) = \left| \det \frac{dq}{dx} \right| \prod_{i=1}^d p_{\mathcal{Z}}(z^i \mid z^{i-1},...,z^{1})
\end{equation}
Since flow models give the exact likelihood, they can be trained by directly optimizing the log likelihood. In addition, thanks to the invertibility of the transformations, one can draw samples by simply inverting the transformations over a set of samples from the latent space.

In order to model the arbitrary conditionals $p(x_u \mid x_o)$, we propose a conditional extension to the change of variable theorem:
\begin{equation}\label{eq:condtrans}
    p_{\mathcal{X}}(x_u \mid x_o) = \left| \det \frac{dq_{x_o}}{dx_u} \right| p_{\mathcal{Z}}(q_{x_o}(x_u) \mid x_o),
\end{equation}
where $q_{x_o}$ is a transformation on the unobserved covariates $x_u$ with respect to the observed covariates $x_o$. 
At a high level, we propose to utilize neural networks to map the conditioning features, $x_o$, to the parameters of invertible transformations $\theta$. In essence, this approach maps the conditioning features to the change of variables \emph{function} that shall be applied to the unobserved features, $x_u$, $x_o \mapsto q_{\theta(x_o)}(\cdot)$, where $q_{\theta(x_o)}: \mathbb{R}^{|u|} \mapsto \mathbb{R}^{|u|}$. In order to deal with the challenges of varying dimensionalities of inputs and outputs, we propose careful masking schemes. For example, to handle the varying cardinality of inputs $x_o$ with a single network, we propose to feed in a zero imputed $d$-dimensional feature vectors (where observed features are filled in with $x_o$ values, and others are zero); we further concatenate a $d$-dimensional indicator vector, where $b_i = \mathbb{I}(i \in o)$ for $i\in\{1, \ldots, d\}$ to disambiguate zero values. Similar masking schemes are used to handle varying dimensionality of outputs. Using this approach we develop linear and non-linear change of variables;
please refer to \citet{li2019flow} for additional details about the conditional transformations.

The conditional likelihoods in latent space $p_{\mathcal{Z}}(z_u \mid x_o) := p_{\mathcal{Z}}(q_{x_o}(x_u) \mid x_o)$ can be computed by either a base distribution, like a Gaussian, or an autoregressive model as in TANs. The Gaussian base likelihood is generally less flexible than using a learned autoregressive one:
\begin{equation}
    p_{\mathcal{Z}}(z_u \mid x_o) = \prod_{i=1}^{|u|} p(z_u^i \mid z_u^{i-1},...,z_u^{1}, x_o).
\end{equation}
Incorporating the autoregressive likelihood into Eq. \eqref{eq:condtrans} yields:
\begin{equation}\label{eq:condtans}
    p(x_u \mid x_o) = \left| \det \frac{dq_{x_o}}{dx_u} \right| \prod_{i=1}^{|u|} p(z_u^i \mid z_u^{i-1},...,z_u^{1}, x_o),
\end{equation}
where $|u|$ is the cardinality of the unobserved covariates. Please refer to \citet{li2019flow} for additional details.

For supervised AFA tasks, we desire to make a prediction to the target $y$ using the acquired subset of features $x_o$. Therefore, we additionally require the model to capture the dependencies between features and the target: $p(x_u, y \mid x_o)$. For real-valued target variables, we can concatenate $x$ and $y$ and treat $y$ as extra dimensions. Since normalizing flows have difficulty in dealing with discrete variables, we employ Bayes's rule for discrete target variables and train a conditional version of ACFlow that is conditioned on the target, i.e., $p(x_u \mid x_o, y)$. Then, the prediction given an arbitrary subset of features $x_o$ can be derived as 
\begin{equation}\label{eq:afa_pred_cls}
P(y \mid x_o) = \frac{p(x_o \mid y)P(y)}{\sum_{y'} p(x_o \mid y')P(y')} = \softmax_y(\log p(x_o \mid y') + \log P(y')),
\end{equation}
where $P(y)$ indicates the prior distribution of the target variables. Similarly, we can also obtain $p(x_u \mid x_o)$ by marginalizing out $y$:
\begin{equation}
    p(x_u \mid x_o) = \frac{\sum_{y'} p(x_u,x_o \mid y') P(y')}{\sum_{y'} p(x_o \mid y')P(y')},
\end{equation}

Training the ACFlow model can be performed by maximizing the conditional log-likelihood $\log p(x_u \mid x_o)$. We formulate the training procedure in a multi-task fashion,
\begin{equation}
    \mathbb{E}_{u,o \sim p(\mathcal{T})}\left[\mathbb{E}_{x \sim p(\mathcal{D})} \left[\log p(x_u \mid x_o) \right] \right],
\end{equation}
where the different configurations of $u$ and $o$ are considered as different tasks from the task distribution $p(\mathcal{T})$. In practice, we select the observed dimensions at random and regard all the remaining dimensions as unobserved. $p(\mathcal{D})$ indicates the data distribution.
For supervised tasks, we additionally maximize the log-likelihood of the target variable, i.e.,
\begin{equation}
    \mathbb{E}_{u,o \sim p(\mathcal{T})}\left[\mathbb{E}_{x \sim p(\mathcal{D})} \left[\log p(x_u \mid x_o) + \beta \log p(y \mid x_o) \right] \right],
\end{equation}
where $\beta$ balances the two loss terms.

\section{AFA with Greedy Acquisition Policy}\label{sec:afa_greedy}
As we have discussed above, the arbitrary conditionals $p(x_u \mid x_o)$ explicitly capture the dependencies among features, which are informative of useful features to acquire. In this section, we propose to assess the informativeness of each candidate feature based on a trained ACFlow model and directly determine the next feature to be acquired in a greedy manner.

As more features are acquired, we expect to increase the confidence of our prediction.
Therefore, we desire that the acquired features contain as much information about the target variable as possible for some given budget.
It is natural, then, to use conditional mutual information to measure the amount of information of features.
Here, we propose a straightforward, yet surprisingly effective strategy that greedily acquires a new unobserved feature with the highest expected mutual information. 
Specifically, we estimate the mutual information between each candidate feature $x_i$ and $y$ conditioned on current observed set $x_o$, i.e., $I(x_i ; y \mid x_o)$, where $o \subseteq \{1,\ldots,d\}$ and $i \in \{1,\ldots,d\} \setminus o$. At each acquisition step, the feature with the highest mutual information will be our choice. Note that before we actually acquire the feature, we do not know the exact value of $x_i$, and we never observe the target variable $y$. Therefore, we need the arbitrary conditional distributions aforementioned (Sec. \ref{sec:acflow}) to infer these quantities. After the acquisition, the newly acquired feature, as well as its value, are added to the observed subset and the model proceeds to acquire the next feature.
Note that since this approach is greedy, the resulting set of acquired features may be sub-optimal (e.g.~in the case where features are jointly informative, but not informative on their own). Notwithstanding, empirical results (Sec. \ref{sec:exp_afa}) show that this greedy approach is effective when using a performant conditional estimator such as ACFlow.
See Algorithm \ref{alg:afa_greedy} for pseudo code of the greedy acquisition process.

\begin{algorithm}[H]
\small
\caption{Active Feature Acquisition with Greedy Acquisition Policy}
\label{alg:afa_greedy}
\begin{algorithmic}[1]
\REQUIRE{pretrained ACFlow model; total feature dimension $d$; stop criterion;}
\STATE{$o \leftarrow \emptyset; u \leftarrow \{1,\ldots,d\}; x_o \leftarrow \emptyset$} \tcp*[f]{initialize the observed set as empty}
\REPEAT
    \STATE{estimate $\hat{I}(x_i ; y \mid x_o)$ for $\forall i \in u$} \tcp*[f]{estimate info-gain for each candidate}
    \STATE{$i^* \leftarrow \argmax_i \hat{I}(x_i ; y \mid x_o)$} \tcp*[f]{select the one with the highest info-gain}
    \STATE{acquire the feature value $x_{i^*}$} \tcp*[f]{acquire the actual value of selected feature}
    \STATE{$o \leftarrow o \cup i^*; u \leftarrow u \setminus i^*; x_o \leftarrow x_o \cup x_{i^*}$} \tcp*[f]{update observed and unobserved set}
    \STATE{predict $y = \argmax p(y \mid x_o)$} \tcp*[f]{predict using current acquired subset}
\UNTIL{stop criterion} \tcp*[f]{good enough prediction or out of acquisition budget}
\end{algorithmic}
\end{algorithm}

The conditional mutual information can be factorized as follows:
\begin{align}
    I(x_i ; y \mid x_o) &= I(x_i ; y, x_o) - I(x_i ; x_o) \label{cmi:0}\\
    &= H(x_i \mid x_o) - \E_{y \sim p(y \mid x_o)}\left[ H(x_i \mid y, x_o)\right] \label{cmi:1}\\
    &= H(y \mid x_o) - \E_{x_i \sim p(x_i \mid x_o)}\left[ H(y \mid x_i, x_o)\right] \label{cmi:2},
\end{align}
where $H(\cdot)$ represents the entropy of the corresponding distribution.
From \eqref{cmi:0}, we see our acquisition metric prefers features that contain more information about $y$ without redundant information already in $x_o$. 
According to \eqref{cmi:1}, when $x_o$ is observed, we prefer $x_i$ with higher entropy, that is, if we are already certain about the value of $x_i$, we do not need to acquire it anymore. Since $H(y \mid x_o)$ is fixed at each acquisition step, from \eqref{cmi:2}, we prefer feature that can reduce the most of our uncertainty about the target $y$.
Next, we will describe how to estimate the conditional mutual information in different settings.

\paragraph{Regression}
For regression problems, the real-valued target variables $y$ can be concatenated with $x$ as inputs to the ACFlow model so that $p(y \mid x_o)$ can be predicted by the same ACFlow as well. The conditional mutual information, by definition, is
\begin{equation}\label{cmi:reg}
    I(x_i ; y \mid x_o) = \E_{p(x_i, y \mid x_o)}\left[ \log \frac{p(x_i, y \mid x_o)}{p(x_i \mid x_o)p(y \mid x_o)}\right] = \E_{p(x_i, y \mid x_o)}\left[ \log \frac{p(y \mid x_i,x_o)}{p(y \mid x_o)}\right].
\end{equation}
We then perform a Monte Carlo estimation of the expectation by sampling from $p(x_i, y \mid x_o)$. Note that $p(y \mid x_i, x_o)$ is evaluated on sampled $x_i$ rather than the exact value, since we have not acquired its value yet.

\paragraph{Classification}
For a discrete target variable $y$, we predict the target following the Bayes's rule \eqref{eq:afa_pred_cls}. To estimate conditional mutual information for classification problems, we can further simplify \eqref{cmi:reg} as
\begin{equation}\label{cmi:cls}
    I(x_i ; y \mid x_o) = \E_{p(x_i \mid x_o)P(y \mid x_i, x_o)}\left[ \log \frac{P(y \mid x_i,x_o)}{P(y \mid x_o)}\right]
    = \E_{p(x_i \mid x_o)}\left[ \KL\left[P(y \mid x_i, x_o) \| P(y \mid x_o)\right]\right],
\end{equation}
where the KL divergence between two categorical distributions can be analytically computed. Note $x_i$ is sampled from $p(x_i \mid x_o)$ as before. We again use Monte Carlo estimation for the expectation.

\paragraph{Unsupervised}
For unsupervised tasks, the conditional mutual information can be similarly computed as in \eqref{cmi:reg} by simply changing $y$ to all the remaining unobserved features $x_r$, where $r = u \setminus {i}$ and $u = \{1,\ldots,d\} \setminus o$.
\begin{equation}\label{cmi:unsup}
    I(x_i ; x_r \mid x_o) = \E_{p(x_u \mid x_o)}\left[ \log \frac{p(x_r \mid x_i,x_o)}{p(x_r \mid x_o)}\right].
\end{equation}
Since $x_r$ typically has higher dimensions than the target $y$, we would need to draw more samples to accurately estimate the expectation in \eqref{cmi:unsup}.

\section{AFA with Reinforcement Learning}\label{sec:afa_rl}
The aforementioned greedy formulations are myopic and unaware of the long-term goal of obtaining multiple features that are \emph{jointly} informative;
instead, the AFA problem can be formulated as a Markov Decision Process (MDP) \citep{zubek2002pruning,ruckstiess2011sequential,shim2018joint}. Therefore, reinforcement learning based approaches can be utilized, where a long-term discounted reward is optimized. In this section we improve the reinforcement learning training of an AFA agent via guidance from a surrogate model.

\subsection{AFA as Markov Decision Process}
We formulate the AFA problem as a Markov Decision Process (MDP) following \citep{zubek2002pruning,shim2018joint}:
\begin{equation}
    s = [o,x_o], \quad a \in u \cup \{\phi\}, \quad
    r(s,a) = -\mathcal{L}(\hat{y}(x_o), y) \mathbb{I}(a=\phi) - \alpha \mathcal{C}(a) \mathbb{I}(a \neq \phi).
\end{equation}
The state $s$ is the current acquired feature subset $o \subseteq \{1,\ldots,d\}$ and their values $x_o$. The action space contains the remaining candidate features $u = \{1,\ldots,d\} \setminus o$ and a special action $\phi$ that indicates the termination of the acquisition process (and to use the prediction $\hat{y}(x_o)$). To optimize the MDP, a reinforcement learning agent acts according to the observed state and receives rewards from the environment. When the agent acquires a new feature $i$, the current state transits to a new state following
\begin{equation}
o \xrightarrow{i} o \cup \{i\}, \quad x_o \xrightarrow{i} x_o \cup \{x_i\},
\end{equation} 
and the reward is the negative acquisition cost of this feature. The value of $x_i$ is obtained from the environment (i.e., we observe the true $i^\mathrm{th}$ feature value for the instance). Once the agent decides to terminate the acquisition ($a=\phi$), it makes a final prediction based on the features acquired thus far and obtains a final reward $-\mathcal{L}(\hat{y}(x_o), y)$. For supervised task, $\mathcal{L}(\hat{y}(x_o), y)$ represents the loss function between the prediction $\hat{y}(x_o)$ and the target $y$; for unsupervised task, the target variable $y$ equals to $x$, and the loss function is $\mathcal{L}(\hat{x}(x_o), x)$. For notational simplicity, we use $\mathcal{L}(\hat{y}(x_o), y)$ to indicate the loss for both supervised and unsupervised tasks. The hyperparameter $\alpha$ controls the trade-off between the prediction loss and the acquisition cost.

\subsection{Generative Surrogate Models for RL}\label{sec:afa_gsmrl}
Although MDPs are broad enough to encapsulate the active acquisition of features, there are several challenges that limit the success of a naive reinforcement learning approach. In the aforementioned MDP, the agent pays the acquisition cost at each acquisition step but only receives a reward about the prediction after completing the acquisition process. This results in sparse rewards leading to credit assignment problems for potentially long episodes \citep{minsky1961steps,sutton1988learning}, which may make training difficult. In addition, an agent that is making feature acquisitions must also navigate a complicated high-dimensional action space, as the action space scales with the number of features, making for a challenging RL problem \citep{dulac2015deep}. Finally, the agent must manage multiple complicated roles as it has to: implicitly model dependencies (to avoid selecting redundant features); perform a cost/benefit analysis (to judge what unobserved informative feature is worth the cost); and act as a classifier (to make a final prediction). Attaining these roles individually is a challenge, doing so jointly and without gradients on the reward (as with the MDP) is an even more daunting problem. To lessen the burden on the agent, and to assuage these challenges, we propose a model-based alternative. Our key observation is that the dynamics of the above MDP can be modeled by the conditional dependencies among features. Thus, we leverage a generative model of \emph{all} conditional dependencies as a surrogate. Equipped with the surrogate model, our method, \emph{Generative Surrogate Models for RL} (GSMRL), essentially combines model-free and model-based RL into a holistic framework. Please refer to Algorithm~\ref{alg:RL} for the pseudo code of the acquisition process.

\paragraph{Feature Dependencies as Dynamics Model} 
A surprisingly unexplored property of the AFA MDP is that the dynamics of the problem are dictated by \emph{conditional dependencies among the data's features}. That is, the state transitions are based on the conditionals: $p(x_j \mid x_o)$, where $x_j$ is an unobserved feature. These conditionals, $p(x_j \mid x_o)$, can be estimated by our ACFlow model (Sec.~\ref{sec:acflow}). Therefore, we frame our approach according to the estimation of conditionals among features with the generative surrogate model. We find that the most efficacious use of our generative surrogate model is a hybrid model-based approach that utilizes intermediate rewards
and side information 
stemming from dependencies.

\paragraph{Intermediate Rewards} 
When the agent terminates the acquisition and makes a prediction, the reward equals to the negative prediction loss using current acquired features. Since the prediction is made at the end of acquisitions, the reward of the prediction is received only when the agent decides to terminate the acquisition process. This is a typical temporal credit assignment problem, which may affect the learning of the agent \citep{minsky1961steps,sutton1988learning}.
Given the surrogate model, we propose to remedy the credit assignment problem \emph{by providing intermediate rewards for each acquisition}. Inspired by the information gain, the surrogate model assesses the intermediate reward for a newly acquired feature $i$ with
\begin{equation}\label{eq:interm_reward}
    r_m(s,i) := H(y \mid x_o) - \gamma H(y \mid x_o, x_i),
\end{equation}
where $\gamma$ is a discount factor for the MDP. 
For discrete target variable, the entropy terms in \eqref{eq:interm_reward} can be analytically computed. For continuous target variable, we approximate the distributions $p(y \mid x_o)$ and $p(y \mid x_o, x_i)$ with Gaussian distributions, and thus the entropy terms can be estimated as a function of the empirical variances.
For unsupervised task where $y:= x$, \eqref{eq:interm_reward} involves estimating the entropy for potentially high dimensional distributions, which itself is an open problem \citep{kybic2007high}. Therefore, we modify the intermediate reward as the reduction of negative log likelihood per dimension, i.e.,
\begin{equation}\label{eq:interm_reward_unsup}
    r_m(s,i) := \frac{-\log p(x_u \mid x_o)}{|u|} - \gamma \frac{-\log p(x_{u \setminus i} \mid x_o, x_i)}{|u|-1}.
\end{equation}
In appendix \ref{sec:invariance}, we show that our intermediate rewards will not change the optimal policy.

\paragraph{Side Information} 
In addition to intermediate rewards, we propose using the surrogate model to also provide auxiliary information to augment the current state and assist the agent. The side information shall explicitly inform the agent of 1) uncertainty and imputations for unobserved features; 2) an estimate of the expected information gain of future acquisitions; 3) uncertainty of the target output for supervised tasks.
This allows the agent to easily assess its current uncertainty and helps the agent `look ahead' to expected outcomes from future acquisitions.

The imputed values and uncertainties of the unobserved features give the agent the ability to look ahead into the future and guide its exploration. For example, if the surrogate model is very confident about the value of a currently unobserved feature, then acquiring it would be redundant. We draw multiple imputations from $p(x_u \mid x_o)$ and provide the agent with the empirical mean and variance of each unobserved feature.

The \emph{expected} information gain assesses the informativeness of each candidate feature, which essentially quantifies the conditional mutual information $I(x_i; y \mid x_o)$ between each candidate feature and the target variable, and can be estimated following \eqref{cmi:reg}, \eqref{cmi:cls} and \eqref{cmi:unsup}. However, it involves some overhead especially for long episodes, since we need to calculate them for each candidate feature at each acquisition step. Moreover, the Monte Carlo estimation of the expectation may require multiple samples. To reduce the computation overhead, we turn to \eqref{cmi:1} and estimate the entropy terms with Gaussian approximations. That is, we approximate $p(x_i \mid x_o)$ and $p(x_i \mid y,x_o)$ as Gaussian distributions and the entropy terms reduce to a function of the corresponding variances. We use sample variance as an approximation. We found that this Gaussian entropy approximation performs comparably while being much faster.

For supervised tasks, we also provide the agent with the current prediction and its uncertainty, which informs the agent about its current confidence and can help the agent determine whether to stop the acquisition process. For the classification task, the prediction probabilities $P(y \mid x_o)$ are used as the side information; for the regression task, we utilize the mean and variance of samples from $p(y \mid x_o)$.

\begin{algorithm}[H]
\caption{Active Feature Acquisition with GSMRL}
\label{alg:RL}
\begin{algorithmic}[1]
\REQUIRE{pretrained surrogate model $M$; agent $\textit{agent}$; test dataset $D$ to be acquired;}
\STATE{instantiate an environment: $\textit{env}=\text{Env}(D)$} \tcp*[f]{define the environment}
\STATE{$x_o$, o = \textit{env}.reset()} \tcp*[f]{initialize the observed set as empty}
\STATE{done = False; reward = 0} \tcp*[f]{initialize a flag variable and episode reward}
\WHILE{not done}
  \STATE{aux = $M$.query($x_o$, o)} \tcp*[f]{query the surrogate model for auxiliary info}
  \STATE{action = \textit{agent}.act($x_o$, o, aux)} \tcp*[f]{evaluate the policy to get an action}
  \STATE{$r_m =$ $M$.reward($x_o$, $o$, action)} \tcp*[f]{intermediate rewards as in \eqref{eq:interm_reward} and \eqref{eq:interm_reward_unsup}}
  \STATE{$x_o$, o, done, r = \textit{env}.step(action)} \tcp*[f]{acquire the feature value or terminate}
  \STATE{reward += r + $r_m$} \tcp*[f]{accumulate rewards}
\ENDWHILE
\STATE{aux = $\textit{M}$.query($x_o$, o)} \tcp*[f]{auxiliary info based on the acquired features}
\STATE{prediction = $\textit{agent}$.predict($x_o$, o, aux)} \tcp*[f]{final prediction}
\STATE{$r_p = $ $\textit{env}$.reward(prediction)} \tcp*[f]{reward from the prediction}
\STATE{reward += $r_p$} \tcp*[f]{accumulate episode reward}
\end{algorithmic}
\end{algorithm}

\subsection{Hierarchical Acquisition Policy}\label{sec:afa_hier}
The model-based approach described above rectifies several shortcomings of a model-free scheme such as JAFA \citep{shim2018joint}. However, there are still some obstacles in extending it to practical use. One of them is the potentially large number of candidate features. Reinforcement learning algorithms are known to have difficulties with a high dimensional action space \citep{dulac2015deep}. In this section, we propose to cluster the candidate features into groups and use a hierarchical reinforcement learning agent to select the next feature to be acquired.

Dealing with large action space for RL is generally challenging, since the agent may not be able to effectively explore the entire action space. Several approaches have been proposed to train RL agents with a large discrete action space. For instance, \citet{dulac2015deep} proposes a Wolpertinger policy that maps a state to a continuous proto-action embedding. The proto-action embedding is then used to look up $k$-nearest valid actions using the given action embeddings. Finally, the action with the highest Q value is selected and executed in the environment. Wolpertinger policy assumes the apriori availability of action representations (embeddings) for k-nearest neighbor searching. However, there is no such representation for general AFA problems. That is, in general datasets, features are enumerated and proximity in indices (i.e., $|i-j|, i,j \in \{1,\ldots,d\}$) is typically \emph{not} informative of feature similarity.
\citet{majeed2020exact} propose a sequentialization scheme, where the action space is transformed into a sequence of $\mathcal{B}$-ary decision code words. A pair of bijective encoder-decoder is defined to perform this transformation. Running the agent will produce a sequence of decisions, which are subsequently decoded as a valid action that can be executed in the environment.

\begin{figure}
    \centering
    \includegraphics[width=0.5\linewidth]{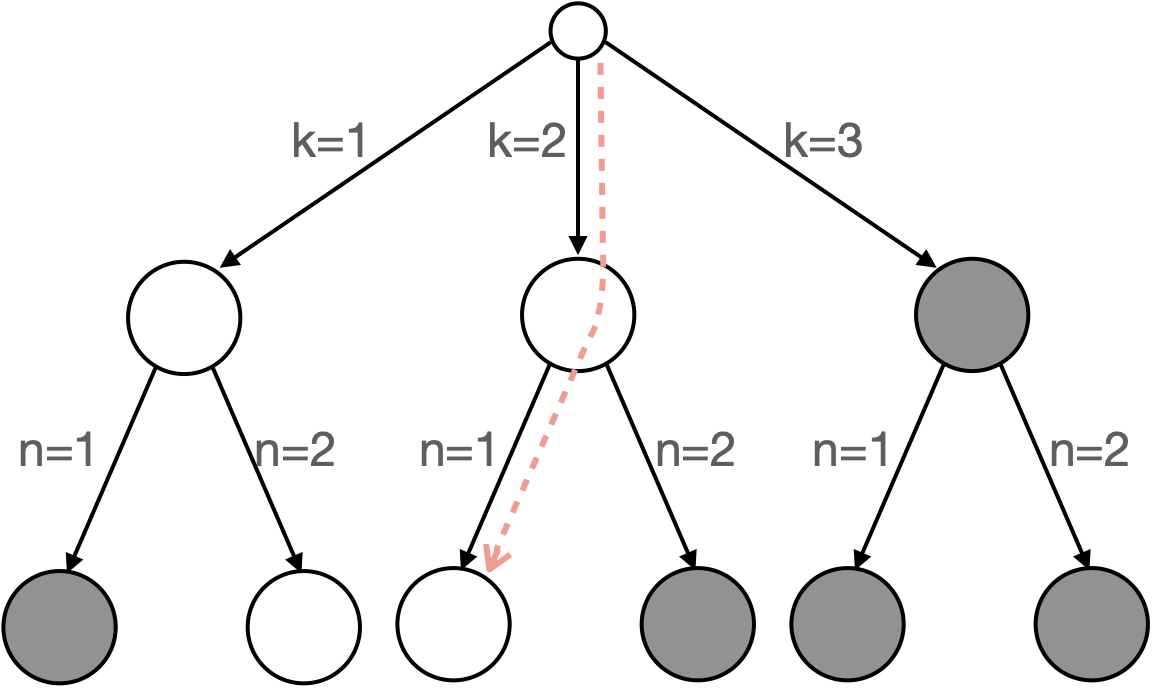}
    \caption{An illustrative example of the grouped action space, where 6 features are grouped into 3 clusters. The grayed circles represent the current observed features (or fully observed groups) and are not considered as candidates anymore. The dashed line shows one acquisition at the current step, which acquires the feature $g_2^{(1)}$. The corresponding circles will be grayed after this acquisition step.}
    \label{fig:action_space}
\end{figure}

Similar to \citep{majeed2020exact}, we also formulate the action space as a sequence of decisions. Here, we propose to cluster the candidate features into hierarchical groups. Given a set of features, $\{x_1,\ldots,x_d\}$, we cluster the features based on their marginal informativeness to the target variable $y$. That is, we partition features based on how they (in isolation) inform one about $y$. 
We assess the informativeness of the candidate features using their mutual information to the target variable, $y$, i.e., $I(x_i ; y)$, where $i \in \{1,\ldots,d\}$. The mutual information can be estimated using the learned arbitrary conditionals of the surrogate model
\begin{equation}
    I(x_i; y) := \E_{p(x_i,y)}\left[ \log \frac{p(x_i, y)}{p(x_i)p(y)}\right] = \E_{p(x_i,y)}\left[ \log \frac{p(y \mid x_i)}{p(y \mid \emptyset)}\right],
\end{equation}
where the expectation is estimated using a held-out validation set. Given the estimated mutual information, we can simply sort and divide the candidate features into different groups. For the sake of implementation simplicity, we use clusters with the same number of features. We may further group features inside each cluster into smaller clusters and develop a tree-structured action space as in \citep{majeed2020exact}, which we leave for future works. Note that the clustering is not performed actively for each instance; instead, we cluster 
features 
once for each dataset and keep the cluster structure fixed throughout the acquisition process. 
Our grouping scheme partitions features based on how informative individual features are (marginally, i.e. in isolation) to the target. This acts as an additional form of auxiliary information (see Fig.~\ref{fig:acquisition}), which guides the agent in earlier acquisitions (as marginally informative features will be more useful). The grouping shall also help guide the agent in later acquisitions, where it may seek less marginally informative features (that may be \emph{jointly} informative with the current observations) to obtain more nuanced discrimination.

It is worth noting that the mutual information $I(x_i ; y)$ is not the only choice for clustering features. Alternative quantities, such as the pairwise mutual information $I(x_i ; x_j)$ or a metric $d(x_i, x_j) := H(x_i, x_j) - I(x_i ; x_j)$, can be used together with a hierarchical clustering procedure to group candidate features. However, these alternatives need to be estimated for each pair of candidate features, which incurs a $O(d^2)$ computational complexity, while the mutual information, $I(x_i ; y)$, only has $O(d)$ complexity. During early experiments, we found the marginal mutual information obtains comparable performance while being much faster to estimate.

Given the grouped action space, $\mathcal{A} = \{g_k\}_{k=1}^{K}$, with $K$ distinct clusters, we develop a hierarchical policy to select one candidate feature at each acquisition step. $g_k = \{g_k^{(1)}, \ldots, g_k^{(N)}\} \subseteq \{1,\ldots,d\}$ represents the $k_{th}$ group of features of size $N$, where $\forall k \neq k',\ g_k \cap g_{k'} = \emptyset$ and $\cup_{k=1}^K g_k = \{1, \ldots, d\}$. The policy factorizes autoregressively by first selecting the group index, $k$, and then selecting the feature index, $n$, inside the selected group, i.e.,
\begin{equation}\label{eq:autoreg_action}
    p(a \mid s) = p(k \mid s)p(n \mid k, s), \quad k \in \{1, \ldots, K\}, \quad n \in \{1, \ldots, N\}.
\end{equation}
The actual feature index being acquired is then decoded as $g_k^{(n)}$. As the agent acquires features, the already acquired features are removed from the candidate set. We simply set the probabilities of those features to zeros and renormalize the distribution. Similarly, if all features of a group have been acquired, the probability of this group is set to zero. With the proposed action space grouping, the original $d$-dimensional action space is reduced to $K+N$ decisions. Please refer to Fig.~\ref{fig:action_space} for an illustration.

\section{Goal-based Explainable AFA}\label{sec:explain_afa}
The AFA models described above provide a way to acquire informative features while minimizing the acquisition cost. 
It is interesting to note that by virtue of dynamically selecting features that are informative, the AFA framework allows one to peer into the estimator for additional insight. That is, AFA allows one to learn
\emph{what features the estimator considers to be important} and \emph{the order (or when) these features were considered to be important}. 
In this section, we propose to enhance the explainability of acquisitions and answer \emph{why a feature was important} by introducing explicit sub-goals that acquisitions try to achieve. That is, we propose to include additional sub-goals for acquisition steps that explicitly provide disambiguation targets. For example, as a result of our extension, it shall be possible to infer that the agent acquired feature $j$ not only because it was informative for our prediction, but specifically because it was informative (disambiguated) between a particular class $c_k$ and class $c_l$. 

\begin{figure}
    \centering
    \includegraphics[width=0.65\linewidth]{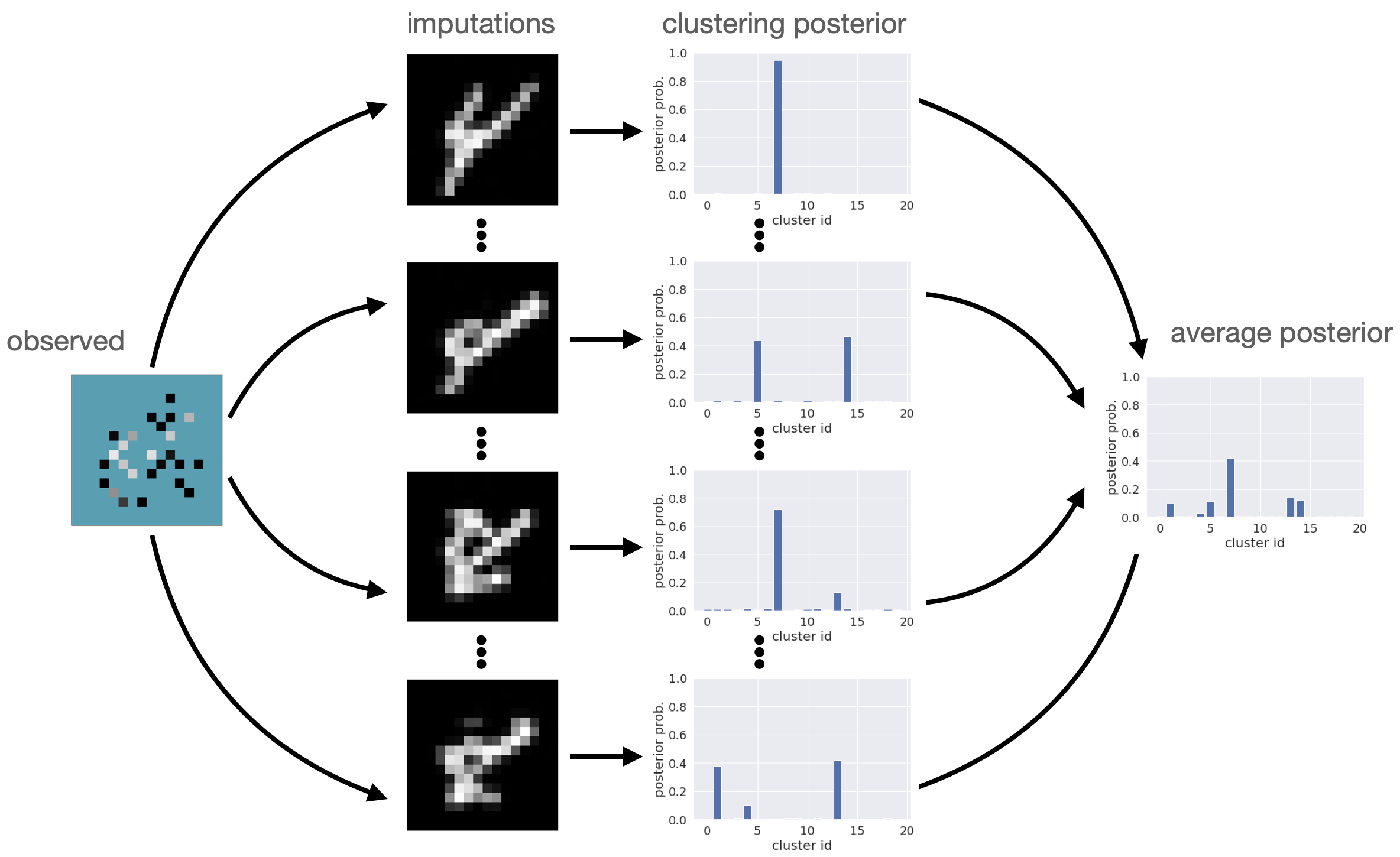}
    \caption{Estimating the clustering posterior for a partially observed instance. We first impute the unobserved features based on the surrogate model, and then get the cluster assignment probabilities for each imputed sample based on a fully observed clustering model. The estimated posterior is the average of those cluster assignment probabilities.}
    \label{fig:posterior}
\end{figure}

\paragraph{Multiclass Sub-goals} We begin by designing sub-goals for increased explainability in the case of multiclass classification. Here we acquire features to predict a set of classes $\mathcal{Y} = \{c_1, c_2,\ldots,c_{|\mathcal{Y}|}\}$. We propose to assign a sub-goal for the next acquisition as follows. Before performing the acquisition step, the agent has a prediction based on the current acquired features, i.e., $p(y \mid x_o)$. Given the current prediction probabilities, we choose two classes, $c_i$ and $c_j$, that are most ambiguous for the agent as the target sub-goal for next acquisition. This sub-goal will encourage the next acquisition to distinguish between these two selected classes. Thus, we can explain \emph{why} the subsequent acquisition was performed: to disambiguate between the selected classes. In practice, we select the two classes with the highest probabilities as the sub-goal:
\begin{equation}
    \text{sub-goal} := \text{topK}(\{p(y=c_k \mid x_o)\}_{k=1}^{|\mathcal{Y}|}, 2),
\end{equation}
where the function $\text{topK}(\cdot,2)$ returns a tuple of class indices with the highest probabilities.

Given the selected sub-goal, $(c_i, c_j)$, the agent learns a conditional policy conditioned on the sub-goal to acquire features. In order to encourage interpretability of the acquisition, we utilize the information gain over the selected classes as an additional reward. Let $p_i = p(y=c_i \mid s)$ and $p_j = p(y=c_j \mid s)$, where $s$ represents the current state (i.e., the observed features), the probability over all remaining classes are $p_r = 1-p_i-p_j$. When the agent acquires a feature $x_i$, it receives a reward as
\begin{equation}\label{eq:xafa_reward_classes}
    r_e(s, i) := H([p_i,p_j,p_r] \mid x_o) - \gamma H([p_i,p_j,p_r] \mid x_o,x_i).
\end{equation}
The reward essentially describes the amount of unique information $x_i$ provides to the selected classes. Therefore, the acquisition is encouraged to achieve the sub-goal.

\paragraph{Cluster Posterior Sub-goals} 
The multiclass sub-goals formulated above will not be informative in problems when the number of classes is small, nor is it applicable when there are no classes (e.g.~in regression or unsupervised AFA tasks).
Next, we propose alternative sub-goals that distinguish among a set of clusters. This sub-goal formulation shall provide fine-grained sub-goals that shall generally be informative. We propose to utilize an \emph{a priori} set of clusters $\mathcal{Z} = \{c_1,c_2,\ldots,c_{|\mathcal{Z}|}\}$, for which one may determine cluster posterior probabilities $p_{\mathcal{Z}}(z \mid x)$.
E.g.~$p_{\mathcal{Z}}$ stemming from a probabilistic clustering model (such as a GMM) over the fully observed instances. 
We now prescribe sub-goals based on disambiguating among a subset of \emph{clusters}, based on \emph{partially observed posteriors}, $p(z \mid x_o)$.

\begin{algorithm}[H]
\caption{Goal-based Explainable Active Feature Acquisition}
\label{alg:xafa}
\begin{algorithmic}[1]
\REQUIRE{pretrained surrogate model $M$; agent $\textit{agent}$; test dataset $D$ to be acquired; the number of clusters for a goal $C$; the number of acquisitions $T$ for each goal;}
\STATE{instantiate an environment: $\textit{env}=\text{Env}(D)$} \tcp*[f]{define the environment}
\STATE{$x_o$, o = \textit{env}.reset()} \tcp*[f]{initialize the observed set as empty}
\STATE{done = False; reward = 0} \tcp*[f]{initialize a flag variable and episode reward}
\WHILE{not done}
  \STATE{goal = \textit{agent}.goal($x_o$, o)} \tcp*[f]{select a goal based on the observed subset}
  \FOR{acquire $T$ features for the goal}
    \STATE{aux = $M$.query($x_o$, o)} \tcp*[f]{query the surrogate model for auxiliary info}
    \STATE{action = \textit{agent}.act($x_o$, o, aux, goal)} \tcp*[f]{ policy conditioned on goal}
    \STATE{$r_m =$ $M$.reward($x_o$, $o$, action)} \tcp*[f]{intermediate rewards as in \eqref{eq:interm_reward} and \eqref{eq:interm_reward_unsup}}
    \STATE{$r_e =$ $M$.reward($x_o$, $o$, action, goal)} \tcp*[f]{reward for goals as in \eqref{eq:xafa_reward_classes} and \eqref{eq:xafa_reward_clusters}}
    \STATE{$x_o$, o, done, r = \textit{env}.step(action)} \tcp*[f]{acquire a feature or terminate}
    \STATE{reward += r + $r_m$ + $r_e$} \tcp*[f]{accumulate rewards}
  \ENDFOR \tcp*[f]{acquire $T$ features to achieve the goal}
\ENDWHILE
\STATE{aux = $\textit{M}$.query($x_o$, o)} \tcp*[f]{auxiliary info based on the acquired features}
\STATE{prediction = $\textit{agent}$.predict($x_o$, o, aux)} \tcp*[f]{final prediction}
\STATE{$r_p = $ $\textit{env}$.reward(prediction)} \tcp*[f]{reward from the prediction}
\STATE{reward += $r_p$} \tcp*[f]{accumulate episode reward}
\end{algorithmic}
\end{algorithm}

We propose to utilize a set of clusters with the highest probabilities as the targets for the cluster-based sub-goal:
\begin{equation}
    \text{sub-goal} := \text{topK}(\{p(z=c_k \mid x_o)\}_{k=1}^{|\mathcal{Z}|}, C),
\end{equation}
the exact number of selected clusters, $C$, is an adjustable hyperparameter. 
The agent similarly learns a sub-goal conditioned acquisition policy and receives a reward based on the information gain over the selected clusters. Let the posterior probability over sub-goal for a given state $s$ as $p_{\text{sub-goal}} = \{p(z=c \mid s); c \in \text{goal}\} \cup \{1-\sum_{c \in \text{sub-goal}} p(z=c \mid s)\}$, the reward can be represented as 
\begin{equation}\label{eq:xafa_reward_clusters}
    r_e(s, i) := H(p_{\text{sub-goal}} \mid x_o) - \gamma H(p_{\text{sub-goal}} \mid x_o, x_i).
\end{equation}

In order to calculate the cluster assignment posterior for a partially observed instance, $p(z \mid x_o)$, we marginalize out the unobserved dimensions
\begin{equation}\label{eq:xafa_posterior}
    p(z \mid x_o) = \mathbb{E}_{p(x_u \mid x_o)} \left[ p_{\mathcal{Z}}(z \mid x_o, x_u) \right],
\end{equation}
where $p_{\mathcal{Z}}$ gives the cluster assignment probabilities for a fully observed instance.
The above expectation is estimated via Monte Carlo samples from our ACFlow surrogate generative model. That is, we perform multiple imputations according to $\hat{x}_u \sim p(x_u \mid x_o)$, and then predict the posterior over clusters using the imputed data. See Fig.~\ref{fig:posterior} for an illustration.

\paragraph{Sub-goals as Hierarchical RL} Our goal based AFA formulation is essentially an example of hierarchical RL, except that the high-level policy for sub-goal selection is a fixed policy based on the prediction (posterior) probabilities.
In practice, we select a sub-goal for every $T$ acquisition step, and all the following $T$ acquisitions are conditioned on the same sub-goal.
Please refer to Algorithm~\ref{alg:xafa} for additional details.

\paragraph{Alternative Sub-goals} Above, we define the sub-goal as the classes (clusters) with the highest probabilities and collapse the remaining ones into a single class (cluster). We note that this is not the only possible sub-goal definition. During early experiments, we explored several alternative sub-goals. For example, selecting the classes (clusters) with the highest probabilities as targets for disambiguation and ignoring the rest. To make it a proper distribution, we re-normalize the probabilities on those selected classes (clusters) so that they sum to one. Another sub-goal explicitly enumerates all combinations of classes (clusters), re-normalizes the distribution for each combination, and then selects the one with the highest entropy. The third sub-goal definition is based on the variance of the cluster assignment probabilities over a large number of imputations. The intuition is that the model is uncertain about a cluster if the cluster has a high probability for some imputations and a low probability for other imputations. Therefore, the ones with the highest variances are selected as sub-goal. During early experiments, we found all those sub-goal definitions give similar results, with our definition being computationally and conceptually simpler and numerically more stable.

\section{Robust AFA}\label{sec:robust_afa}
Next, we extend our AFA framework for additional robustness. Namely, we consider the ability to detect out-of-distribution (OOD) instances whilst actively acquiring features.
In many real-world scenarios, it is possible for an AFA model to encounter inputs that are different from its training distribution. For example, in medical domains, an AFA model may be asked to acquire features for patients with a disease not seen during training time.
Rather than silently acquiring potential irrelevant features, and making an incorrect prediction in these OOD instances, here we develop an approach that will select features that also yield good OOD detection and allow the model to alert end-users when it is being deployed on instances that differ from what was seen at training time.
Note that dealing with out-of-distribution inputs is difficult in general \citep{nalisnick2018do}, and is even more challenging for AFA models since the model only has access to a subset of features at any acquisition step. In this section, we propose a novel algorithm for detecting out-of-distribution inputs with partially observed features and utilize it to improve the robustness of the AFA models.

\subsection{Partially Observed Out-of-distribution Detection}\label{sec:po3d}
Current state-of-the-art approaches to out-of-distribution (OOD) detection assume data are fully observed. In an AFA framework, however, data are partially observed at any acquisition step, which renders those approaches inappropriate. In this section, we develop a novel OOD detection algorithm specifically tailored for partially observed data. 

Inspired by MSMA \citep{mahmood2021multiscale}, we propose to use the norm of scores from an arbitrary marginal distribution $p(x_o)$ as summary statistics and further detect partially observed OOD inputs with a DoSE \citep{morningstar2021density} approach.
MSMA for fully observed data is built by the following steps:
\begin{enumerate}[label=(\roman*)]
    \item Train a noise conditioned score matching network $s_\theta$ \citep{song2019generative} with $L$ noise levels by optimizing
    \begin{equation}\label{eq:score_matching}
        \frac{1}{L}\sum_{i=1}^{L} \frac{\sigma_i^2}{2} \E_{p_{data}(x)}\E_{\tilde{x} \sim \calN(x, \sigma_i^2 I)} \left[ \left\Vert s_\theta(\tilde{x}, \sigma_i) + \frac{\tilde{x}-x}{\sigma_i^2} \right\Vert_2^2 \right].
    \end{equation}
    The score network essentially approximates the score of a series of smoothed data distributions $\nabla_{\tilde{x}} \log q_{\sigma_i}(\tilde{x})$, where $q_{\sigma_i}(\tilde{x}) = \int p_{data}(x)q_{\sigma_i}(\tilde{x} \mid x) dx$, and $q_{\sigma_i}(\tilde{x} \mid x)$ transforms $x$ by adding some Gaussian noise form $\calN(0, \sigma_i^2I)$.
    
    \item For a given input $x$, compute the L2 norm of scores at each noise level, i.e., $\texttt{s}_i(x) = \Vert s_\theta(x, \sigma_i) \Vert$.
    
    \item Fit a low dimensional likelihood model for the norm of scores using in-distribution data, i.e., $p(\texttt{s}_1(x), \ldots, \texttt{s}_L(x))$, which is called density of states in \citep{morningstar2021density} following the concept in statistical mechanics.
    
    \item Threshold the likelihood $p(\texttt{s}_1(x), \ldots, \texttt{s}_L(x))$ to determine whether the input $x$ is OOD or not.
\end{enumerate}
In order to deal with partially observed data, we propose to modify the score network to output scores of arbitrary marginal distributions, i.e., $\nabla_{\tilde{x}_o}\log q_{\sigma_i}(\tilde{x}_o)$, where $o \subseteq \{1, \ldots, d\}$ represents an arbitrary subset of features. We do so by extending \eqref{eq:score_matching} to
\begin{equation}
    \frac{1}{L}\sum_{i=1}^{L} \frac{\sigma_i^2}{2} \E_{p_{data}(x)}\E_{\tilde{x} \sim \calN(x, \sigma_i^2 I)}\E_{o \sim p(o)} \left[ \left\Vert s_\theta(\tilde{x} \odot \I_o, \I_o, \sigma_i) \odot \I_o + \frac{\tilde{x} \odot \I_o-x \odot \I_o}{\sigma_i^2} \right\Vert_2^2 \right],
\end{equation}
where $\I_o$ represents a $d$-dimensional binary mask indicating the partially observed features, $\odot$ represents the element-wise product operation, and $p(o)$ is the distribution for generating observed dimensions. Similarly to the fully observed case, we compute the L2 norm of scores at each noise level, i.e., $\texttt{s}_i(x_o) = \Vert s_\theta(x \odot \I_o, \I_o, \sigma_i) \odot \I_o \Vert$, and fit a likelihood model in this transformed low-dimensional space. The likelihood model is also conditioned on the binary mask $\I_o$ to indicate the observed dimensions, i.e., $p(\texttt{s}_1(x_o),\ldots,\texttt{s}_L(x_o) \mid \I_o)$. Given an input $x$ with observed dimensions $o$, we define the OOD score as 
\begin{equation}\label{eq:ood_score}
-\log p(\texttt{s}_i(x_o), \ldots, \texttt{s}_L(x_o) \mid \I_o).
\end{equation}
A larger OOD score means the instance is more likely to be out-of-distribution.
We then threshold the OOD score to determine whether the partially observed instance $x_o$ is OOD or not. In practice, we report the AUROC scores instead of choosing a fixed threshold.
To train the partially observed MSMA (PO-MSMA), we generate a mask for each input data $x$ at random. The conditional likelihood over norm of scores is estimated by a conditional autoregressive model, for which we utilize the efficient masked autoregressive implementation \citep{papamakarios2017masked}.

One benefit of our proposed PO-MSMA approach is that a single model can be used to detect OOD inputs with arbitrary observed features, which is convenient for detecting OOD inputs along the acquisition trajectories. Furthermore, sharing weights across different tasks (i.e., different marginal distributions) could act as a regularization, thus the unified score matching network can potentially perform better than separately trained networks for each different conditional.

\subsection{Robust Active Feature Acquisition}\label{subsec:robust_afa}
In this section, we integrate partially observed OOD detection into the AFA framework. That is, we actively acquire features for a given problem and simultaneously detect OOD inputs using the acquired subset of features. 
In order to guide the AFA agent to acquire features that are informative for OOD detection, we propose to use the OOD score \eqref{eq:ood_score} as an auxiliary reward.
The reward is given at the end of the acquisition process, therefore, it depends on all the acquired features:
\begin{equation}\label{eq:ood_reward}
    r_d(x_o) = -\log p(\texttt{s}_i(x_o), \ldots, \texttt{s}_L(x_o) \mid \I_o).
\end{equation}
Intuitively, this auxiliary reward encourages the agent to acquire features that may detect OOD instances since it is high if the acquired features lead to a high OOD score. 

\begin{wrapfigure}{r}{0.4\linewidth}
\centering
\includegraphics[width=\linewidth]{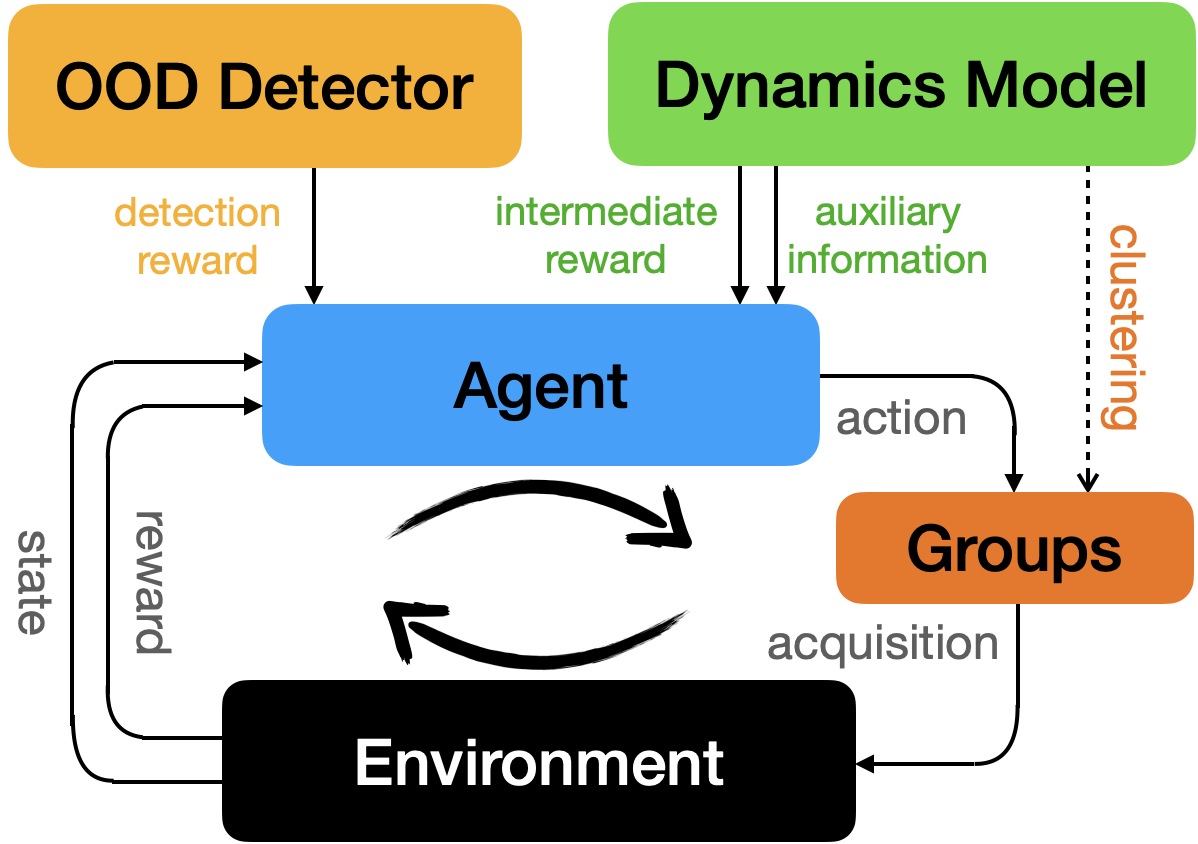}
\vspace{-10pt}
\caption{Schematic illustration of our robust AFA framework.}
\label{fig:robust_afa}
\end{wrapfigure}

In summary, our robust AFA framework contains a dynamics model, a grouping of actions (features), an OOD detector and an RL agent. The dynamics model captures the arbitrary conditionals, $p(x_u, y \mid x_o)$, and is utilized to provide auxiliary information and intermediate rewards (see section~\ref{sec:afa_gsmrl}). It also enables a simple and efficient action space grouping technique and thus scales AFA up to applications with large action spaces (see section~\ref{sec:afa_hier}). The partially observed OOD detector is used to distinguish OOD inputs alongside the acquisition procedure and is also used to provide an auxiliary reward so that the agent is encouraged to acquire informative features for OOD detection. The RL agent takes in the current acquired features and auxiliary information from the dynamics model and predicts what next feature to acquire. When the feature is actually acquired, the agent pays the acquisition cost of the feature and receives an intermediate reward from the dynamics model. When the acquisition process is terminated, the agent makes a final prediction about the target, $y$, using all its acquired features, and receives a reward based on its prediction, $-\mathcal{L}(\hat{y}, y)$. It also receives a reward from the OOD detector based on the negative log-likelihood of the score norms of the acquired feature subset(see \eqref{eq:ood_reward}). Please refer to Algorithm~\ref{alg:robust_afa} for additional details and to Fig.~\ref{fig:robust_afa} for an illustration.

\begin{algorithm}[H]
\caption{Robust Active Feature Acquisition}
\label{alg:robust_afa}
\begin{algorithmic}[1]
\REQUIRE{acquisition environment $\textit{env}$; dynamics model $\textit{M}$; partially observed OOD detector $\textit{D}$; AFA agent $\textit{agent}$; acquisition budget $\textit{B}$}
\STATE{$x_o$, o = \textit{env}.reset()} \tcp*[f]{initialize the observed set as empty}
\STATE{done = False; reward = 0} \tcp*[f]{initialize a flag variable and episode reward}
\WHILE{not done}{
  \STATE{aux = $\textit{M}$.query($x_o$, o)} \tcp*[f]{query the surrogate model for auxiliary info}
  \STATE{action = \textit{agent}.act($x_o$, o, aux)} \tcp*[f]{evaluate the policy to get an action}
  \STATE{$r_m =$ $\textit{M}$.reward($x_o$, $o$, action)} \tcp*[f]{intermediate rewards as in \eqref{eq:interm_reward} and \ref{eq:interm_reward_unsup}}
  \STATE{$x_o$, o, done, $r$ = \textit{env}.step(action)} \tcp*[f]{acquire a feature or terminate}
  \STATE{reward += $r$ + $r_m$}} \tcp*[f]{accumulate rewards}
\ENDWHILE
\STATE{aux = $\textit{M}$.query($x_o$, o)} \tcp*[f]{auxiliary info based on the acquired features}
\STATE{prediction = $\textit{agent}$.predict($x_o$, o, aux)} \tcp*[f]{final prediction}
\STATE{$r_p = $ $\textit{env}$.reward(prediction)} \tcp*[f]{reward from prediction}
\STATE{$r_d =$ $\textit{D}$.reward($x_o$, $o$)} \tcp*[f]{reward from OOD detector, see \eqref{eq:ood_reward}}
\STATE{reward += $r_p$ + $r_d$} \tcp*[f]{accumulate episode reward}
\STATE{ood\_score = $\textit{D}$.score($x_o$, o)} \tcp*[f]{OOD detection scores, see \eqref{eq:ood_score}}
\end{algorithmic}
\end{algorithm}

\section{Related Works}\label{sec:related_works}
\subsection{Arbitrary Conditionals}
Previous attempts to learn probability distributions conditioned on arbitrary subsets of known covariates include the Universal Marginalizer  \citep{douglas2017marginalizer}, which is trained as a feed-forward network to approximate the marginal posterior distribution of each unobserved dimension conditioned on the observed ones. 
VAEAC \citep{ivanov2018variational} utilizes a conditional variational autoencoder and extends it to deal with arbitrary conditioning. The decoder network outputs likelihoods that are over all possible dimensions, although, since they are conditionally independent given the latent code, it is possible to use only the likelihoods corresponding to the unobserved dimensions.
NeuralConditioner (NC) \citep{belghazi2019learning} is a GAN-based approach that leverages a discriminator to distinguish real data and imputed samples.
ACE \citep{strauss2021arbitrary} utilizes an energy based approach to model the arbitrary conditional distributions.
\citet{strauss2022any} propose to learn a posterior network for partially observed instances so that any VAE model can be turned to model arbitrary conditionals.
Unlike VAEAC and NC, ACFlow is capable of producing an analytical (normalized) likelihood and avoids blurry samples and mode collapse problems ingrained in these approaches. Furthermore, in contrast to the Universal Marginalizer, ACFlow captures dependencies between unobserved covariates at training time via the change of variables formula.

\subsection{Active Feature Acquisition}
Previous works have explored actively acquiring features in the cost-sensitive setting. Active perception is a relevant sub-field where a robot with a mounted camera is planning by selecting the best next view \citep{bajcsy1988active,aloimonos1988active, cheng2018reinforcement,jayaraman2018learning}. In this work, we consider general features and take images as one of many data sources. For general data, \citet{ling2004decision}, \citet{chai2004test} and \citet{nan2014fast} propose decision tree, naive Bayes, and maximum margin based classifiers, respectively, to jointly minimize the misclassification cost and feature acquisition cost. \citet{ma2018eddi} and \citet{gong2019icebreaker} acquire features greedily using mutual information as the estimated utility. \citet{zubek2002pruning} formulate the AFA problem as an MDP and fit a transition model using complete data, then they use the AO* heuristic search algorithm to find an optimal policy. \citet{ruckstiess2011sequential} formulate the problem as a partially observable MDP and solve it using Fitted Q-Iteration. \citet{he2012imitation} and \citet{he2016active} instead employ a imitation learning approach guided by a greedy reference policy. \citet{shim2018joint} utilize Deep Q-Learning and jointly learn a policy and a classifier. The classifier is treated as an environment that calculates the classification loss as the reward. ODIN \citep{zannone2019odin} presents an approach to learn a policy and a prediction model using augmented data with a Partial VAE \citep{ma2018eddi}. In contrast, GSMRL uses a surrogate model, which estimates both the state transitions and the prediction in a unified model, to directly provide intermediate rewards and auxiliary information to an agent.

\subsection{Model-based and Model-free RL}
Reinforcement learning can be roughly grouped into model-based methods and model-free methods depending on whether they use a transition model \citep{li2017deep}. Model-based methods are more data efficient but could suffer from significant bias if the dynamics are misspecified. On the contrary, model-free methods can handle arbitrary dynamic system but typically requires substantially more data samples. There have been works that combine model-free and model-based methods to compensate with each other. The usage of the model includes generating synthetic samples to learn a policy \citep{gu2016continuous}, back-propagating the reward to the policy along a trajectory \citep{heess2015learning}, and planning \citep{chebotar2017combining,pong2018temporal}. In this work, we rely on the model to provide intermediate rewards and side information.

\subsection{Explainable Machine Learning}
Explainable Artificial Intelligence (XAI), i.e., the development of more transparent and interpretable AI models, has gained increased interest over the last few years. In order to explain the model behavior, researchers have explored methods that find an explanation for a model in a post-hoc fashion. The techniques explored in this direction include feature importance \citep{breiman2001random,casalicchio2018visualizing}, local surrogate model \citep{ribeiro2016should}, Shapley values \citep{shapley1953value}, etc. Models that are intrinsically explainable have also been built, such as linear models, tree based models, rule based models \citep{friedman2008predictive}, etc. Please refer to the comprehensive surveys \citet{tjoa2020survey,islam2021explainable} for details. Explainable RL, as a subarea of XAI, focuses on developing methods to reveal the underlying reasoning for their decisions. Programmatically Interpretable Reinforcement Learning framework (PIRL) proposes to represent the policy as a high-level, human readable programming language \citep{verma2018programmatically}. Linear Model U-Trees (LMUTs) utilize decision trees to represent the policy. \citet{madumal2020explainable} propose to use the structural causal model to explain how the agent's action influence the environment. Please see \citet{puiutta2020explainable} for a comprehensive survey for Explainable RL. In this work, we explain the actions by the specific sub-goals they are trying to achieve.

\subsection{Out-of-distribution Detection}\label{sec:ood}
Detecting OOD inputs is an active research direction nowadays. One approach relies on the uncertainty of prediction \citep{ovadia2019can,kumar2019verified}. The predictions for OOD inputs are expected to have higher uncertainty. Bayesian neural networks (BNNs) \citep{blundell2015weight}, deep ensemble \citep{lakshminarayanan2017simple} and MC dropout \citep{gal2016dropout} are all explored to provide reliable uncertainty estimations. Recently, Gaussian processes have also been combined with deep neural networks to quantify the prediction uncertainty, such as SNGP \citep{liu2020simple} and DUE \citep{van2021improving}.
In addition to discriminative models, generative models are expected to be capable of detecting OOD inputs using the likelihood \citep{bishop1994novelty}, that is, OOD inputs should have a lower likelihood than the in-distribution ones. However, recent works \citep{nalisnick2018do, hendrycks2018deep} show that it is not the case for high dimensional distributions. This oddity has been verified for many modern deep generative models, such as PixelCNN, VAE and normalizing flows. Ever since, people have developed various techniques to rectify this pathology \citep{choi2018waic,ren2019likelihood,nalisnick2019detecting,morningstar2021density, mahmood2021multiscale,bergamin2022model}. Our method builds on top of several recently proposed techniques and extends them to detect OOD with partially observed instances.

\section{Implementation Details}\label{sec:implement}
We implement our AFA framework in Tensorflow \citep{abadi2016tensorflow}. Our code is available at \url{https://github.com/lupalab/DistAFA}. 
To implement the surrogate model, we adapt the original ACFlow models \citep{li2019flow} to include the target variables.
The MDP is optimized using Proximal Policy Optimization (PPO) algorithm \citep{schulman2017proximal}. 
The state (i.e., partially observed features) is imputed with zeros and coupled with a binary mask to indicate the observed dimensions.
The state, binary mask and the auxiliary information from the surrogate model are then concatenated as input to a feature extraction network.
The extracted embedding is shared for both the policy and the critic network.

The AFA model acquires features sequentially and adds the newly acquired feature to the observed subset thus giving an updated state. We also manually remove the acquired features from the candidate set since acquiring the same feature repeatedly is redundant. For certain applications, the features themselves might be internally ordered. For instance, when acquiring features from time-series data, the acquired features must follow the chronological order since we cannot go back in time to acquire another feature, therefore we need to remove all the features behind the acquired features from the candidate set. That is, after acquiring feature $t$, features $\{1,\ldots,t\}$ are no longer considered as candidates for the next acquisition. Similar spatial constraints can also be applied to spatial data. To satisfy those constraints, we manually set the probabilities of the invalid actions from the policy network to zeros and re-normalize the remaining probabilities.

For explainable AFA, we cluster the fully observed training set using GMM models. The cluster assignment posterior \eqref{eq:xafa_posterior} is estimated by making multiple imputations from the surrogate model and clustering the imputed instances. During experiments, we use 50 imputations. We observe no significant differences when using more imputations.

The partially observed OOD detector, PO-MSMA, consists of a score matching network and a likelihood model over the norm of scores. We modify the original NCSN model \citep{song2019generative} to produce the scores for arbitrary marginal distributions $\nabla_{\tilde{x}_o} \log q_{\sigma_i}(\tilde{x}_o)$. Specifically, the inputs contain the masked images $x_o$ and a binary mask indicating the observed pixels. The output is a tensor of the same size as the input image. We then mask out the unobserved dimensions for the output and compute the norm only for those observed pixels. Throughout the experiment, we use 10 noise scales, thus obtaining 10 summary statistics, $s_1, \ldots, s_{10}$, for each input image. Then, we train a conditional autoregressive model for the norms conditioned on the binary mask $\I_m$, i.e., $p(s_1, \ldots,s_{10} \mid \I_o)$. Given a test image $x$ with observed dimensions $o$, the OOD detection starts by calculating the norm of scores on different noise levels. Then, the conditional likelihood $p(s_1,\ldots,s_{10} \mid \I_o)$ is thresholded to determine whether the given partially observed input is OOD or not. Here, we report the AUROC scores to evaluate the OOD detection performance. 

\section{Experiments}\label{sec:experiments}
In this section, we conduct extensive experiments to validate the effectiveness of our AFA framework. We evaluate on both tabular data, such as those from UCI repository \citep{Dua:2019}, and images, where one pixel value is acquired at a time. We investigate our model for both the supervised and unsupervised AFA tasks below.
In Sec.~\ref{sec:exp_afa}, we compare our greedy and RL-based distribution guided AFA approaches with baselines. Previous works have typically considered smaller datasets (both in terms of the number of features and the number of instances). We instead consider a broad range of datasets with more instances and higher dimensionality. In terms of comparisons, previous works often compare to naively simple baselines, such as a random acquisition order. Here, we compare our approaches to the state-of-the-art models with both greedy policy and non-greedy RL policy. We also conduct thorough ablation studies to verify the effectiveness of our model. Section~\ref{sec:exp_hier} provides empirical evidence that our proposed action space grouping can help the agent navigate the high-dimensional action space when there is a large set of candidate features. In Sec.~\ref{sec:exp_xafa}, we empirically assess how informative the selected sub-goals are in explaining the agent's acquisitions. We explore both a set of classes and a set of clusters as the sub-goal. Section~\ref{sec:exp_rafa} evaluates our proposed partially observed OOD detection model and the robust AFA framework.

\begin{figure*}
\begin{minipage}{\linewidth}
    \centering
    \includegraphics[width=0.98\textwidth]{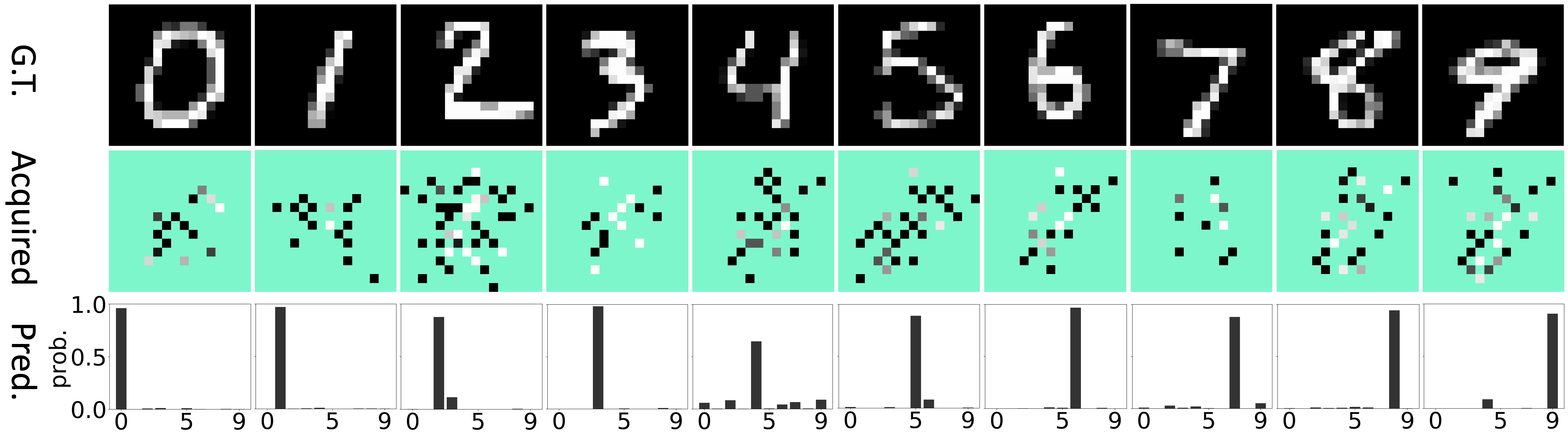}
    \caption{Example of acquired features and prediction probabilities. The green masks indicate the unobserved features.}
    \label{fig:mnist_cls}
\end{minipage}

\begin{minipage}{\linewidth}
    \vspace{15pt}
    \centering
    \subfigure[GSM+Greedy]{
    \includegraphics[width=0.48\linewidth]{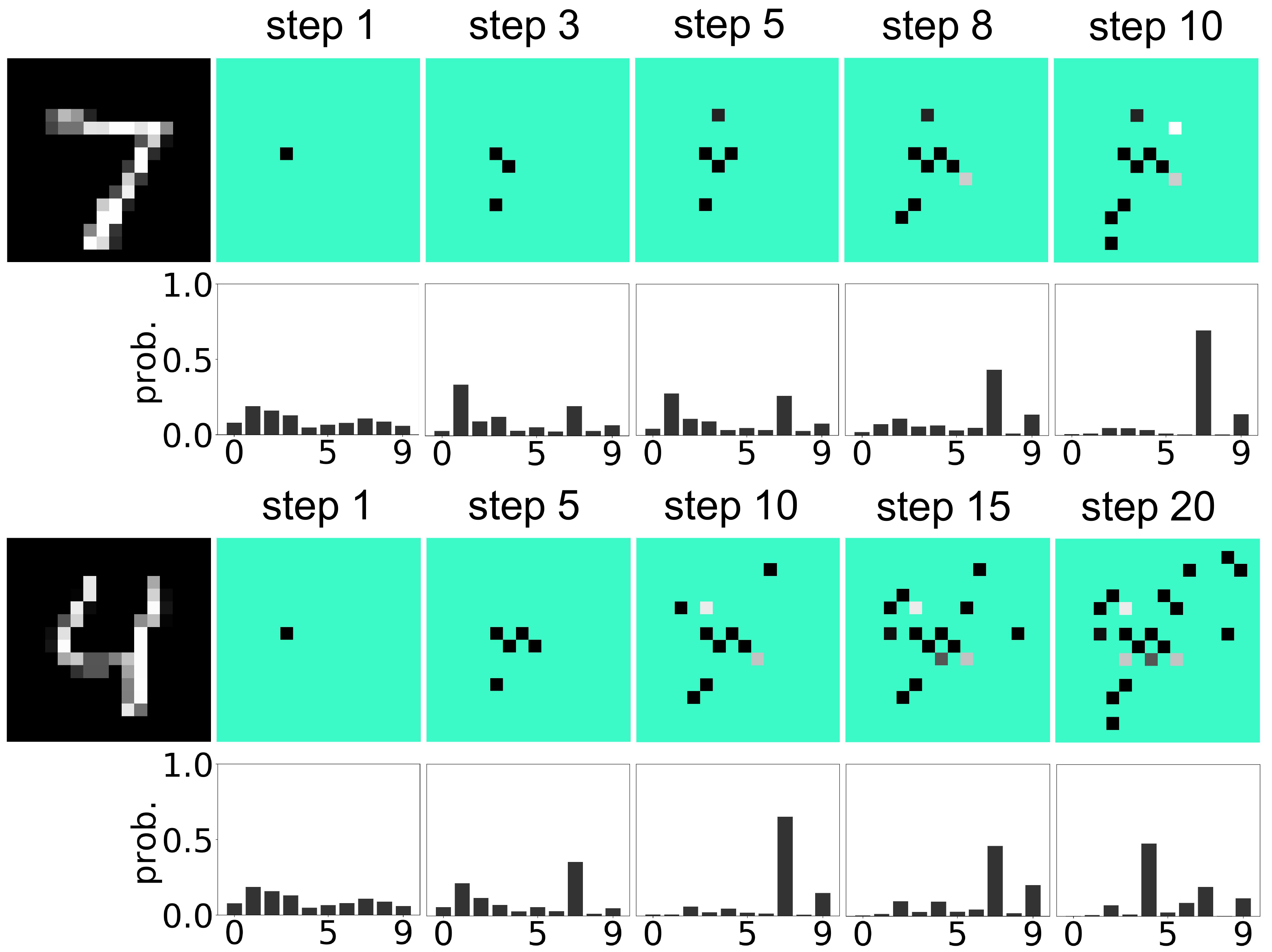}}
    \subfigure[GSMRL]{
    \includegraphics[width=0.485\linewidth]{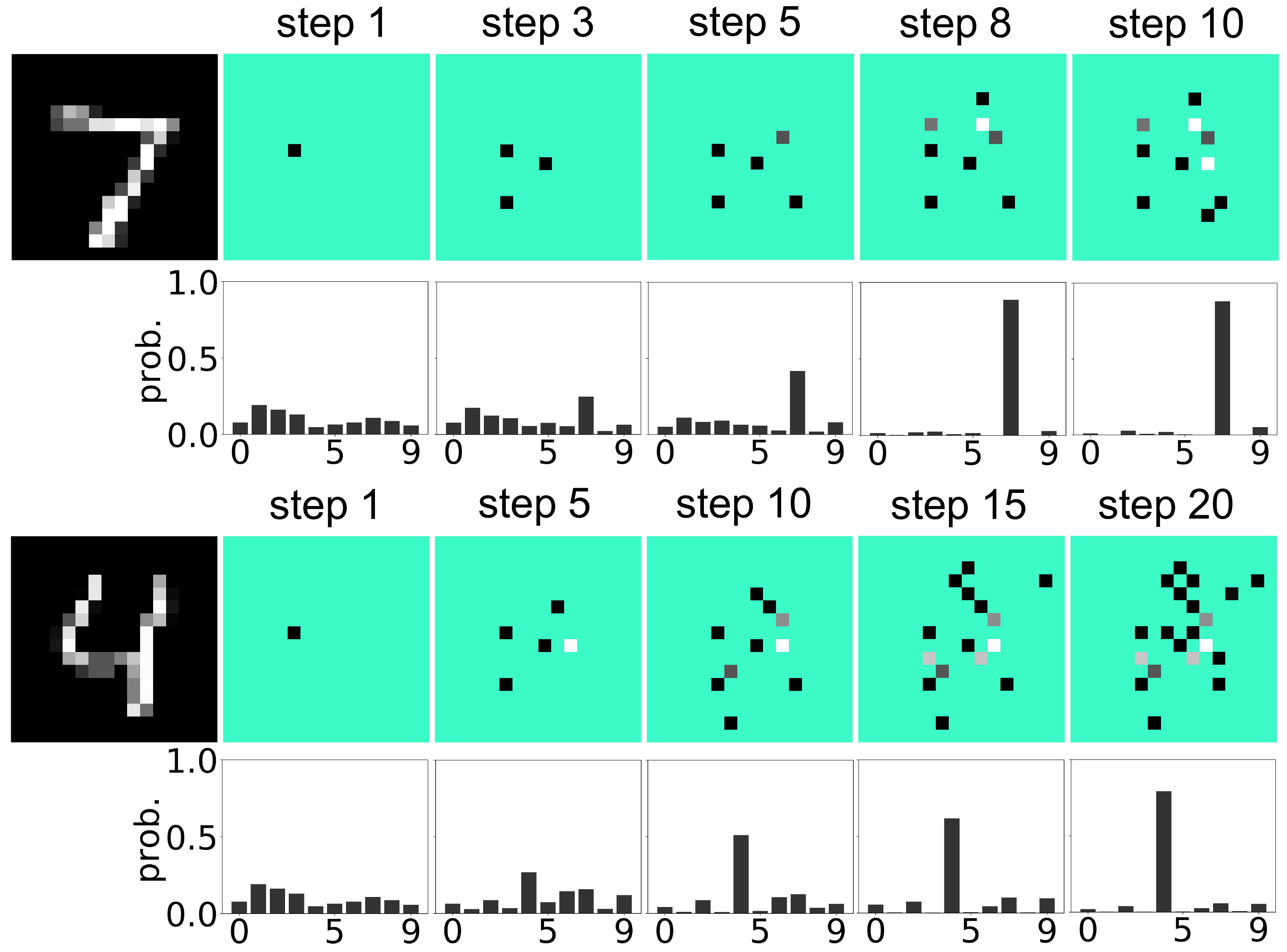}}
    \caption{Examples of the acquisition process for MNIST classification.}
    \label{fig:mnist_afa}
\end{minipage}
\end{figure*}

\begin{figure*}
    \centering
    \subfigure{\includegraphics[width=0.45\linewidth]{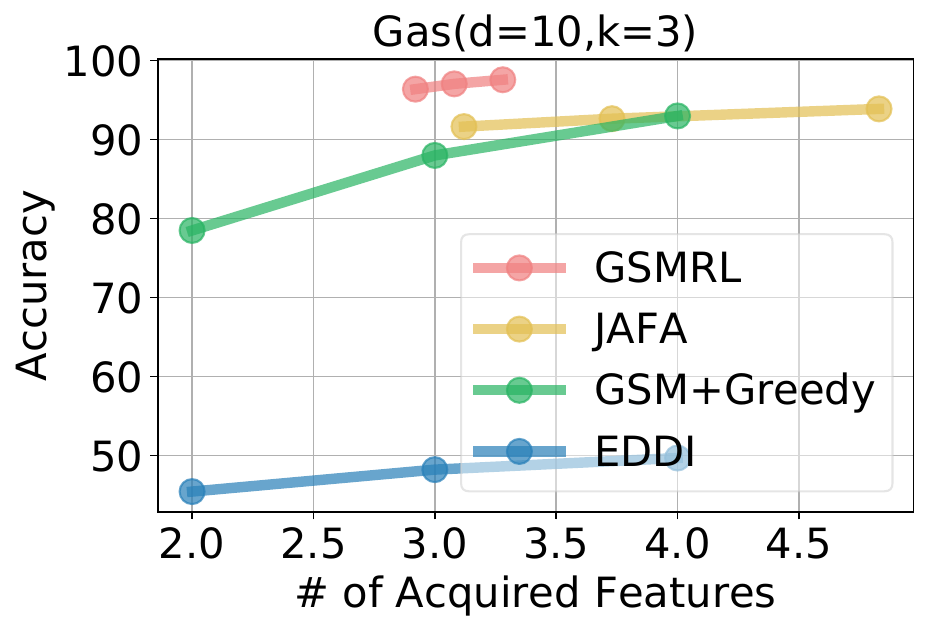}}
    \quad \quad
    \subfigure{\includegraphics[width=0.44\linewidth]{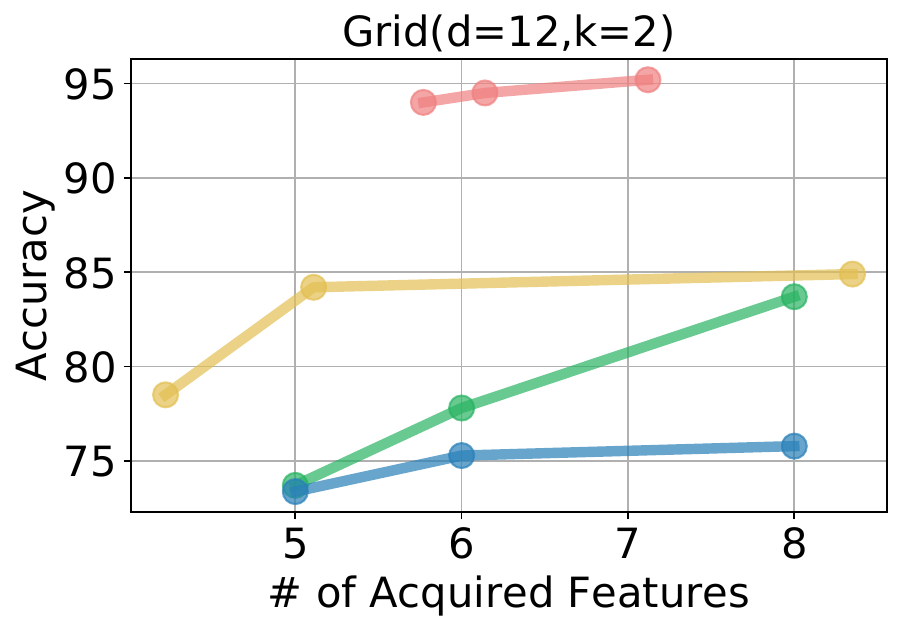}}
    \caption{Classification accuracy on UCI datasets for supervised AFA tasks.}
    \label{fig:uci_cls_acc}
\end{figure*}

\subsection{AFA with Greedy and RL Policy}\label{sec:exp_afa}
We assess our proposed greedy policy and the GSMRL policy on several benchmark environments built upon the UCI repository \citep{Dua:2019} and MNIST dataset \citep{lecun1998mnist}. We compare our methods to a model-free RL approach, JAFA \citep{shim2018joint}, which jointly trains an agent and a classifier. We also compare to a greedy policy EDDI \citep{ma2018eddi} that estimates the utility for each candidate feature using a VAE based model and selects one feature with the highest utility at each acquisition step. We use a fixed cost for each feature and report multiple results with different $\alpha$ (see equation \eqref{eq:goal_sup} and \eqref{eq:goal_unsup}) to control the trade-off between task performance and acquisition cost.

\subsubsection{Numerical Results}
\begin{wrapfigure}{r}{0.45\linewidth}
    \centering
    \vspace{-13pt}
    \includegraphics[width=\linewidth]{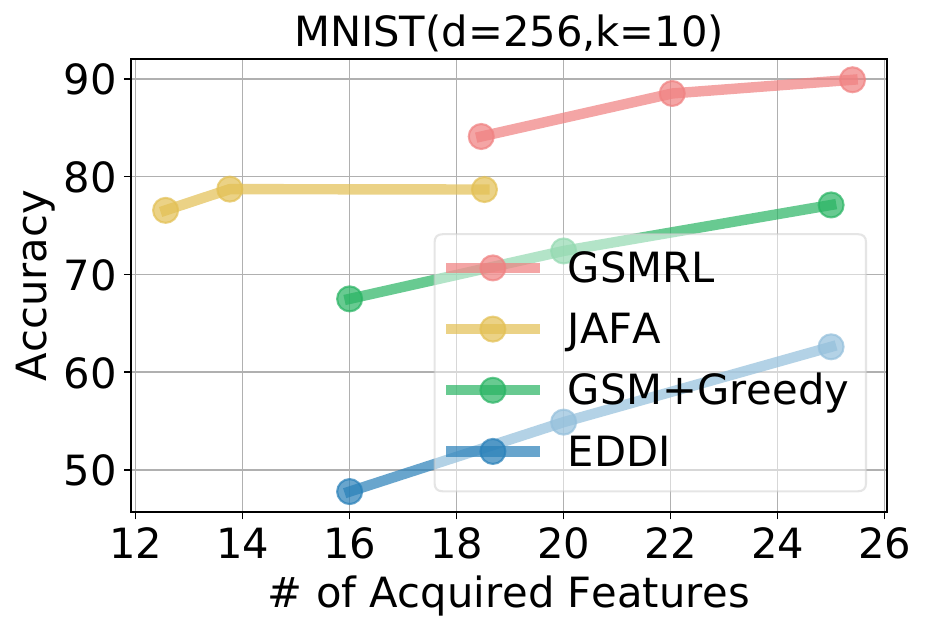}
    \vspace{-22pt}
    \caption{Supervised AFA for MNIST classification.}
    \label{fig:mnist_cls_acc}
\end{wrapfigure}
\paragraph{Classification}
We first perform classification on the MNIST dataset. We downsample the original images to $16\times16$ to reduce the action space since the greedy approaches have difficulty scaling.
Fig.~\ref{fig:mnist_cls} illustrates several examples of the acquired features using GSMRL and their final prediction probabilities after acquisition. We can see that our GSMRL model acquires a different subset of features for different instances. Notice the checkerboard patterns of the acquired features, which indicates our model is able to exploit the spatial correlation of the data. 
Fig.~\ref{fig:mnist_afa} presents the acquisition process for our greedy policy and GSMRL. We also present the prediction probabilities along the acquisition. We can see the prediction become certain after acquiring only a small subset of features. 
The test accuracy in Fig.~\ref{fig:mnist_cls_acc} demonstrates the superiority of GSMRL over other baselines. It typically achieves higher accuracy with a lower acquisition cost. It is worth noting that our surrogate model with a greedy acquisition policy outperforms EDDI. We believe the improvement is due to the better distribution modeling ability of our surrogate model so that the utility and the prediction are more accurately estimated. 
Similarly, we also perform classification using several UCI datasets. The test accuracy is presented in Fig.~\ref{fig:uci_cls_acc}. Again, our GSMRL model outperforms baselines under the same acquisition budget.

\begin{figure}

\begin{minipage}{\linewidth}
    \centering
    \subfigure{\includegraphics[width=0.45\textwidth]{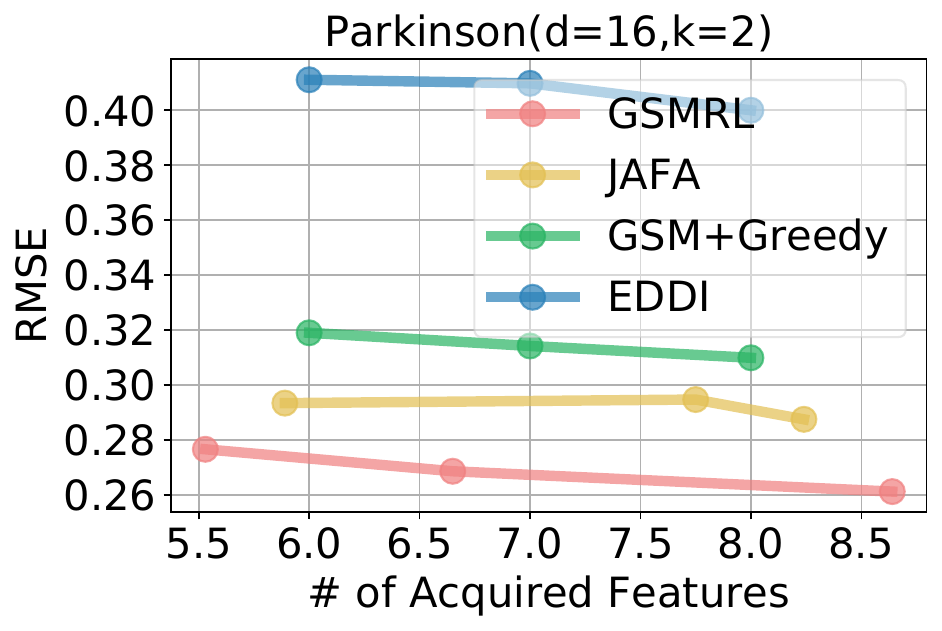}}
    \quad \quad
    \subfigure{\includegraphics[width=0.45\textwidth]{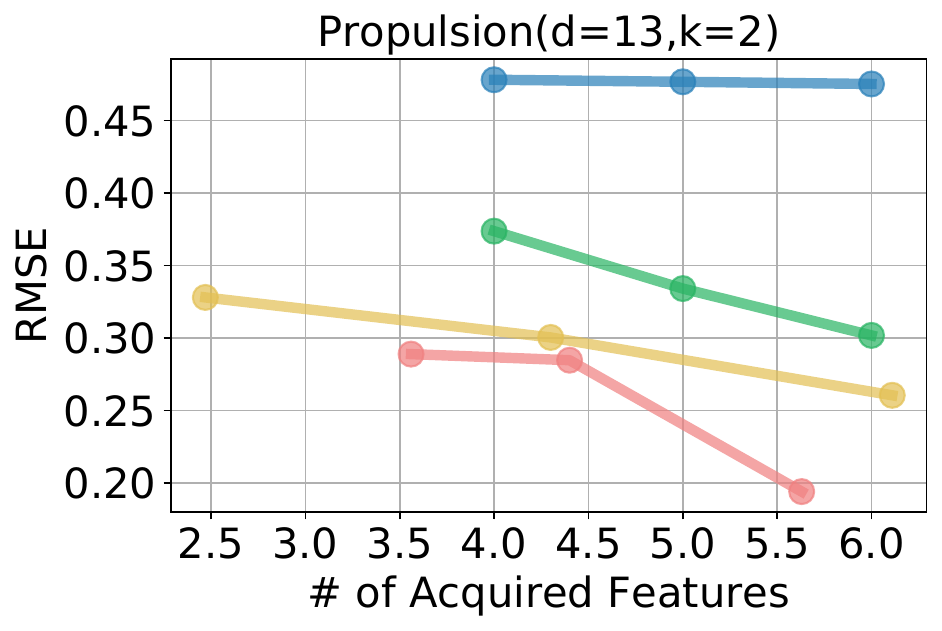}}
    \caption{Regression RMSE on UCI datasets for supervised AFA tasks.}
    \label{fig:uci_reg_rmse}
\end{minipage}

\begin{minipage}{\linewidth}
    \vspace{15pt}
    \centering
    \subfigure{\includegraphics[width=0.45\textwidth]{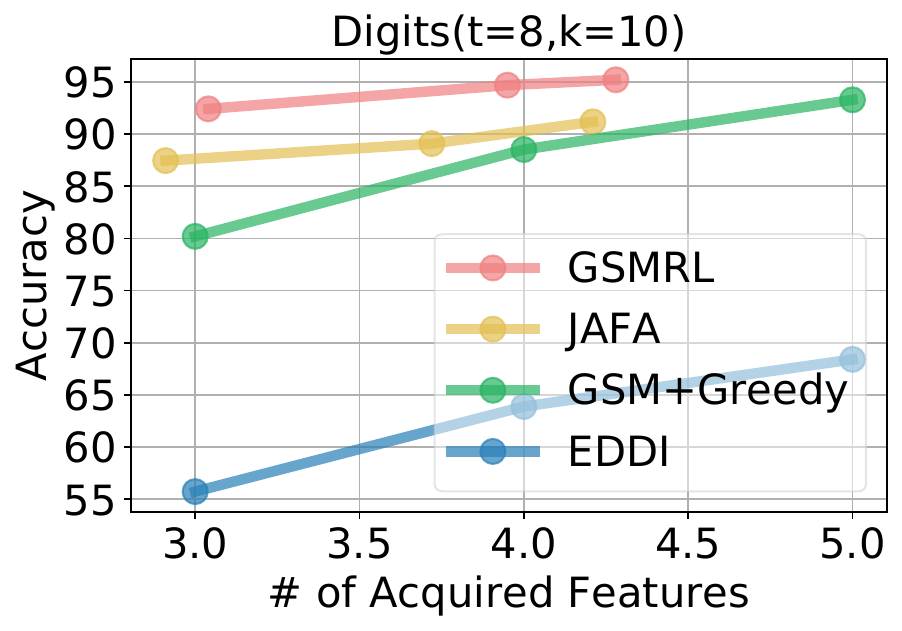}}
    \quad \quad
    \subfigure{\includegraphics[width=0.45\textwidth]{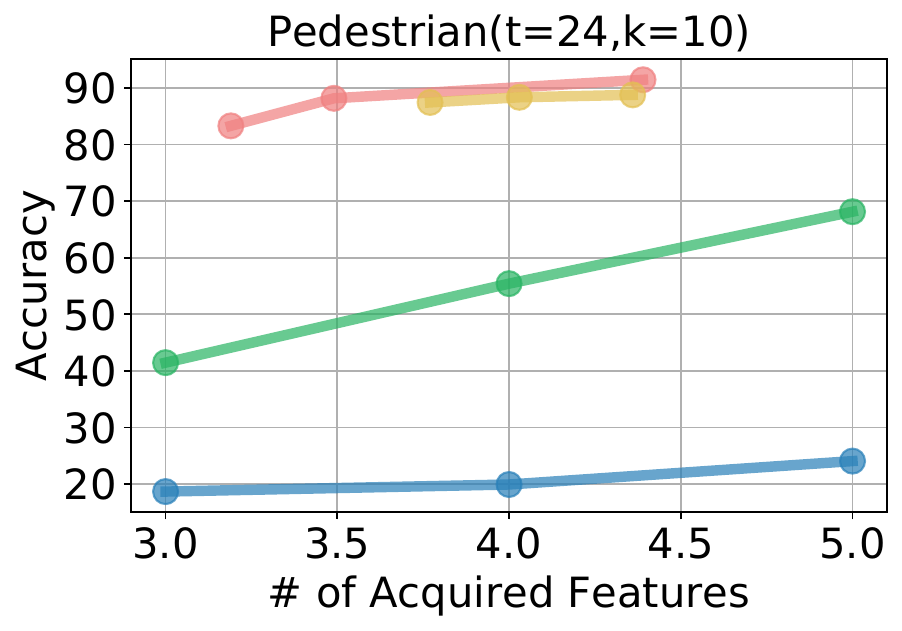}}
    \caption{Supervised AFA for time series classification.}
    \label{fig:uci_ts_acc}
\end{minipage}

\begin{minipage}{0.45\linewidth}
    \vspace{15pt}
    \centering
    \includegraphics[width=\linewidth]{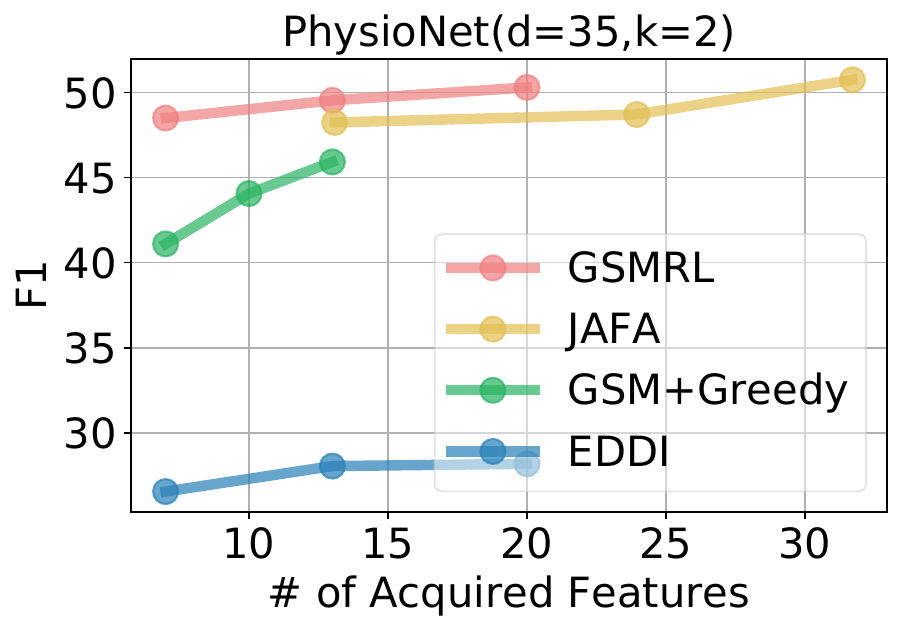}
    \vspace{-10pt}
    \caption{F1 scores for predicting mortality on Physionet.}
    \vspace{-10pt}
    \label{fig:physionet12}
\end{minipage}
\quad \quad
\begin{minipage}{0.45\linewidth}
    \vspace{15pt}
    \centering
    \includegraphics[width=\linewidth]{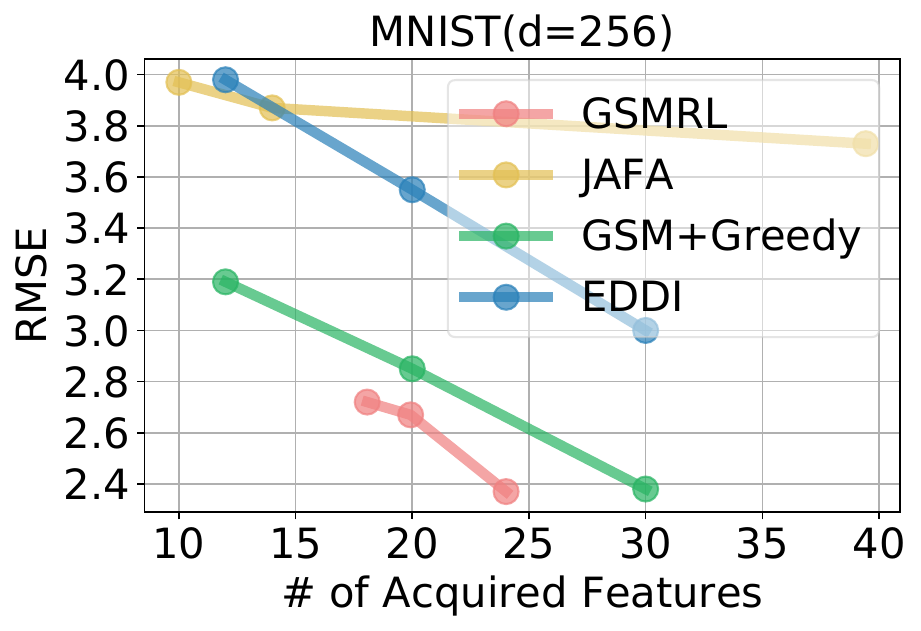}
    \vspace{-10pt}
    \caption{RMSE of imputing $x_u$ on MNIST. Lower is better.}
    \vspace{-10pt}
    \label{fig:mnist_air_rmse}
\end{minipage}

\end{figure}

\paragraph{Regression}
In addition to classification tasks, we also conduct experiments for regression tasks using several UCI datasets. We report the root mean squared error (RMSE) of the target variable in Fig.~\ref{fig:uci_reg_rmse}. Similar to the classification task, our model outperforms baselines with a lower acquisition cost.

\paragraph{Time Series}
To evaluate the performance with constraints in action space, we classify over time series data where the acquired features must follow chronological ordering (i.e., once we acquire a feature at time $t$, no additional features at time $t' < t$ may be acquired). These results illustrate our models' ability to acquire features in a temporal domain, which is of use in many real world applications. The datasets are from the UEA \& UCR time series classification repository \citep{bagnall2017great}. We manually clip the probability of invalid actions to zero. Fig.~\ref{fig:uci_ts_acc} shows the accuracy with different numbers of acquired features. Our method achieves high accuracy by collecting a small subset of the features. 

\paragraph{Medical Diagnosis}
As a real-world application, we evaluate the feature acquisition performance for medical diagnosis. We use the Physionet challenge 2012 dataset \citep{goldberger2000physiobank} to predict in-hospital mortality. We first preprocess the dataset by removing some non-relevant features (such as patient ID) and eliminating the instances with a very high missing rate (larger than 80\%). Since the classes are heavily imbalanced, we use weighted cross entropy as training loss and the final rewards. For evaluation, we report the F1 scores in Fig.~\ref{fig:physionet12}. Compared to baselines, our model achieves higher F1 with lower acquisition costs.

\begin{figure*}
\begin{minipage}{\linewidth}
    \centering
    \includegraphics[width=0.98\textwidth]{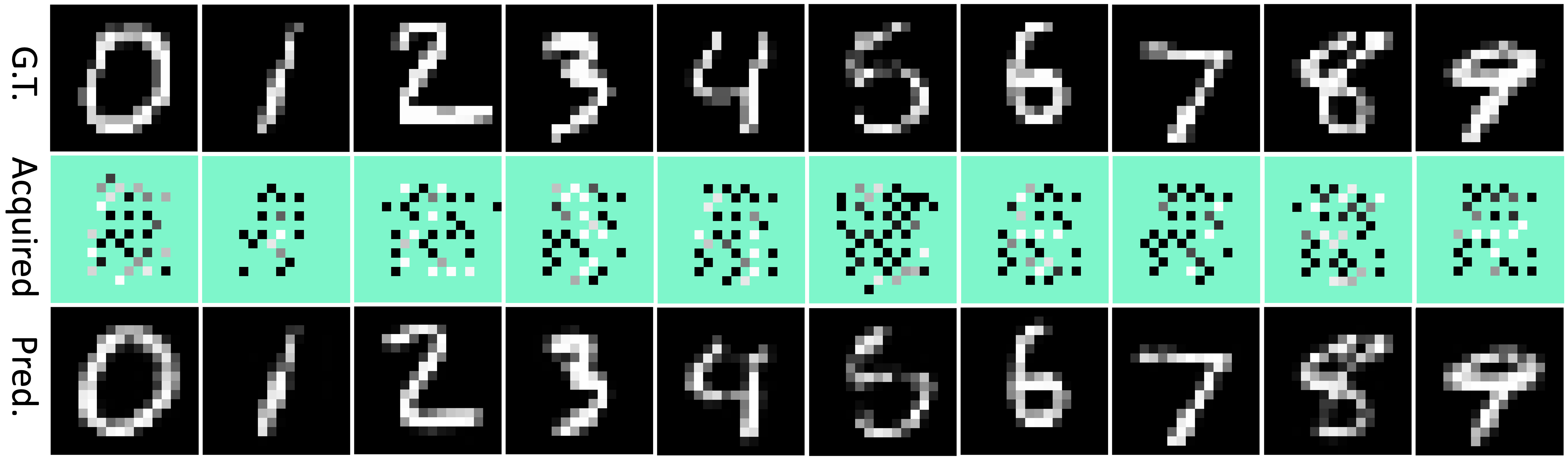}
    \vspace{-5pt}
    \caption{Example of acquired features and inpaintings. The green masks indicate the unobserved features.}
    \label{fig:mnist_rec}
\end{minipage}

\begin{minipage}{\linewidth}
    \centering
    \subfigure[GSM+Greedy]{
    \includegraphics[width=0.48\textwidth]{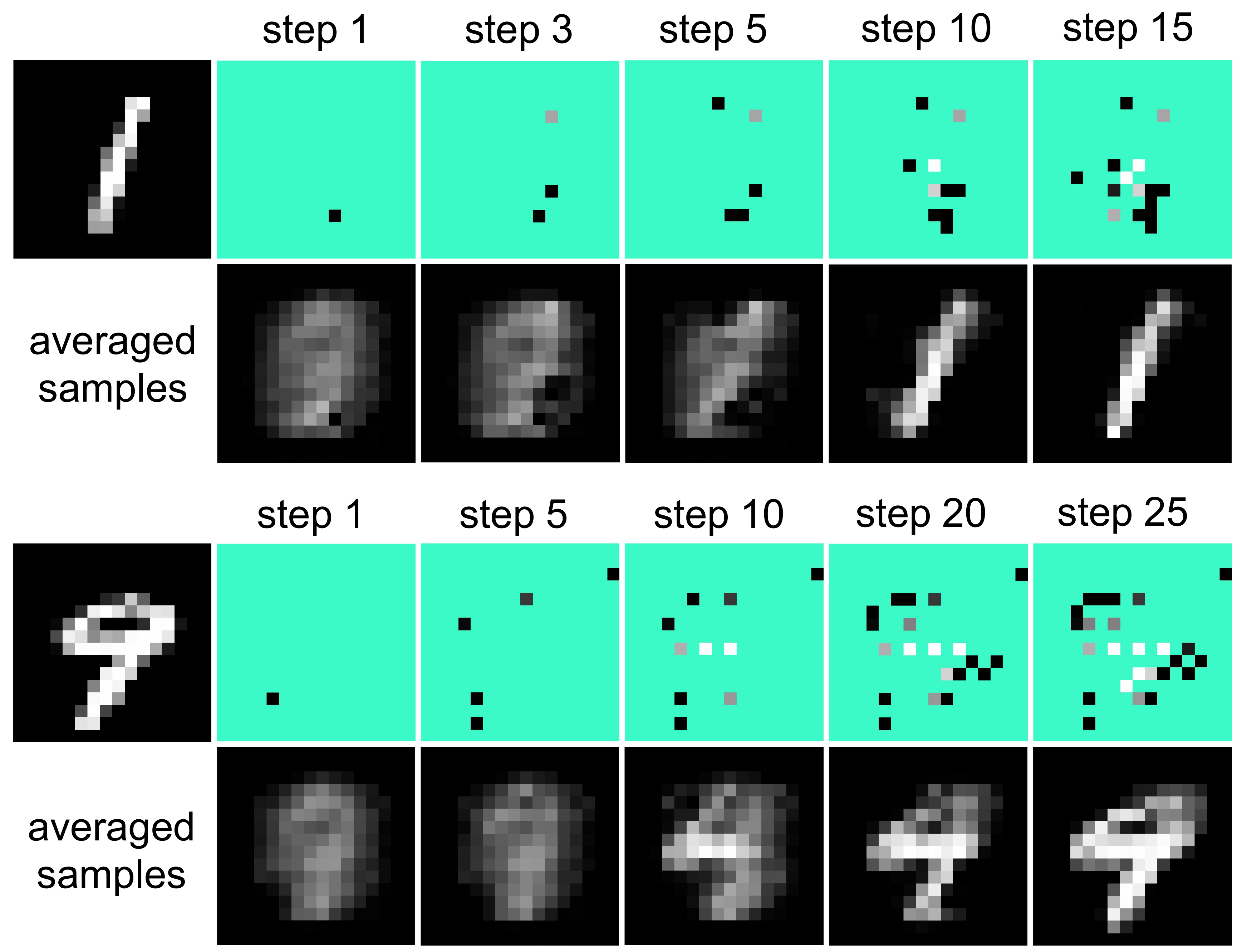}}
    \subfigure[GSMRL]{
    \includegraphics[width=0.48\textwidth]{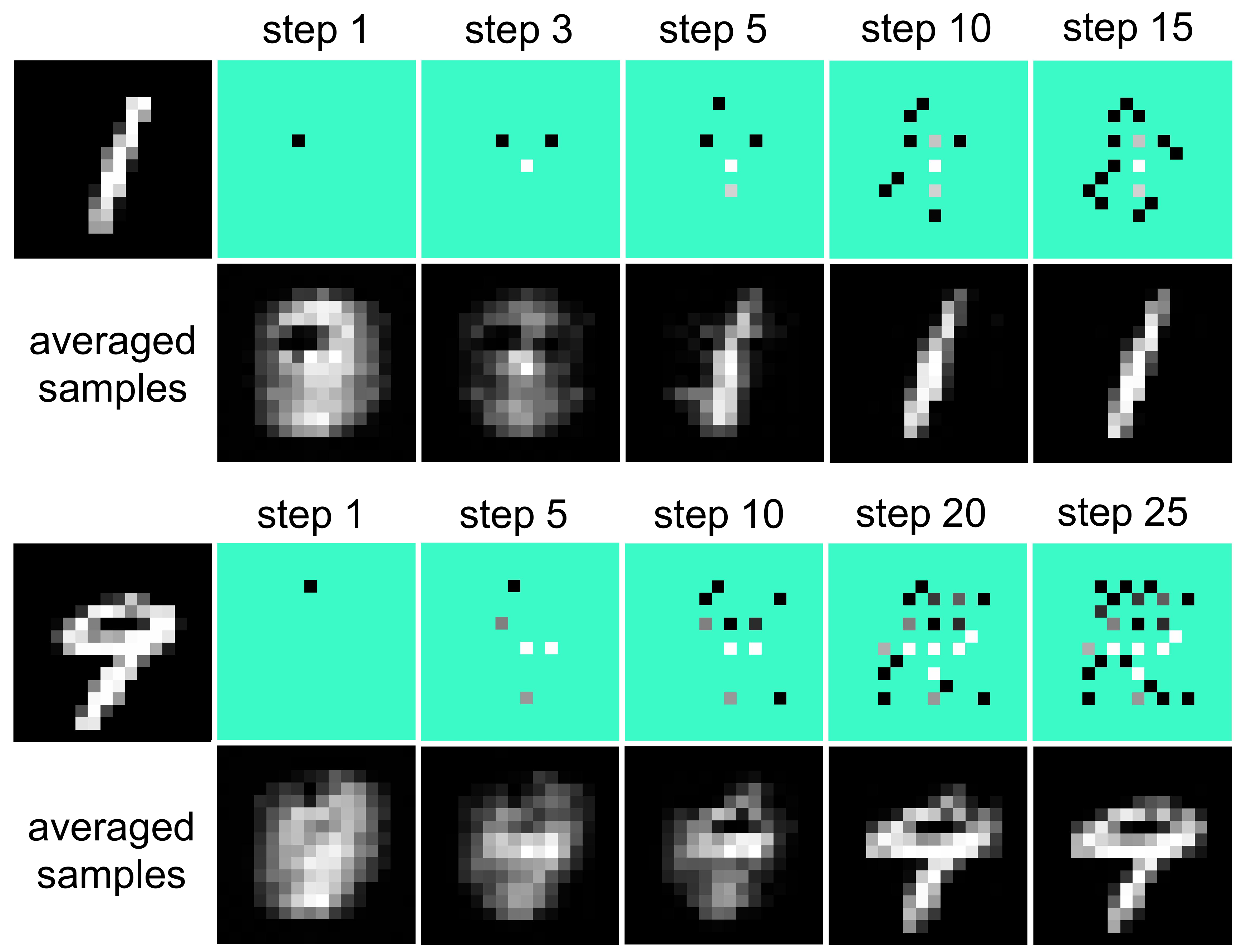}}
    \vspace{-5pt}
    \caption{Examples of the acquisition process for unsupervised AFA on MNIST.}
    \label{fig:mnist_air}
\end{minipage}
\vspace{-5pt}
\end{figure*}

\begin{figure*}
    \subfigure{\includegraphics[width=0.45\textwidth]{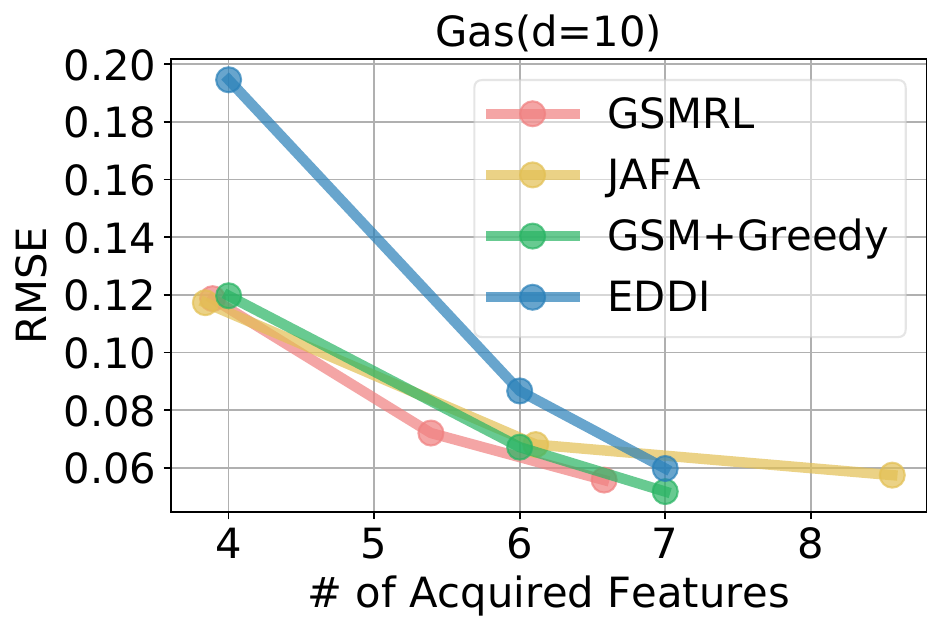}}
    \quad \quad
    \subfigure{\includegraphics[width=0.45\textwidth]{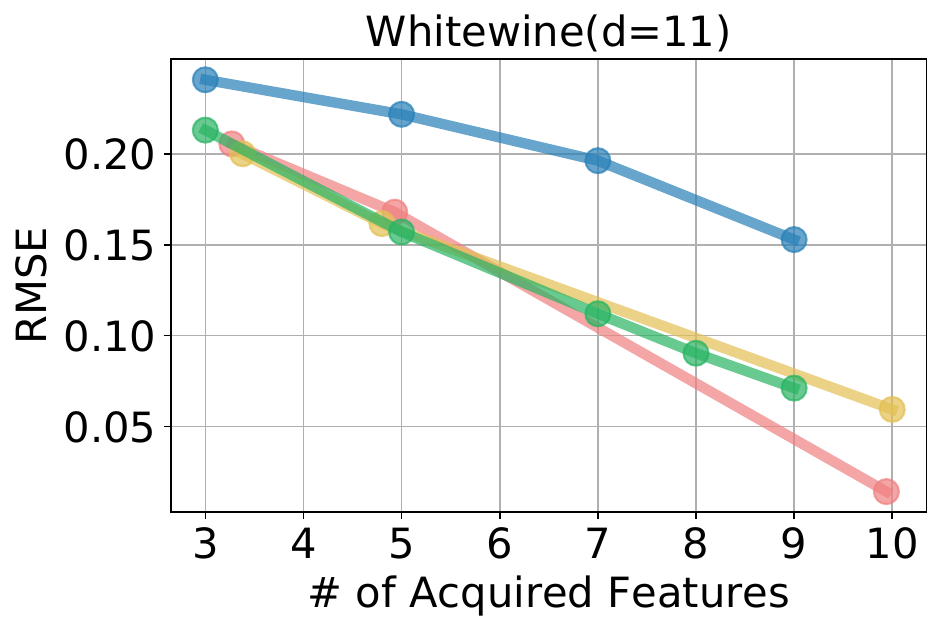}}
    \vspace{-5pt}
    \caption{RMSE for unsupervised AFA tasks on UCI datasets.}
    \label{fig:uci_air_rmse}
    \vspace{-5pt}
\end{figure*}

\paragraph{Unsupervised AFA}
Next, we evaluate our method on unsupervised AFA tasks where features are actively acquired to impute the remaining unobserved features. 
We use negative MSE as the reward for GSMRL and JAFA. The greedy policy calculates the utility following \eqref{cmi:unsup}. For low dimensional UCI datasets, our method is comparable to baselines as shown in Fig.~\ref{fig:uci_air_rmse}; however, for the high dimensional case, as shown in Fig.~\ref{fig:mnist_air_rmse}, our method performs considerably better. Note JAFA is worse than the greedy policy for MNIST. We found it hard to train the policy and the reconstruction model jointly without the help of the surrogate model in this case. See Fig.~\ref{fig:mnist_air} for examples of the acquisition process.

\subsubsection{Ablations}
We now conduct a series of ablation studies to explore the capabilities of our AFA framework.

\begin{wrapfigure}{r}{0.45\linewidth}
    \centering
    \vspace{-2pt}
    \includegraphics[width=\linewidth]{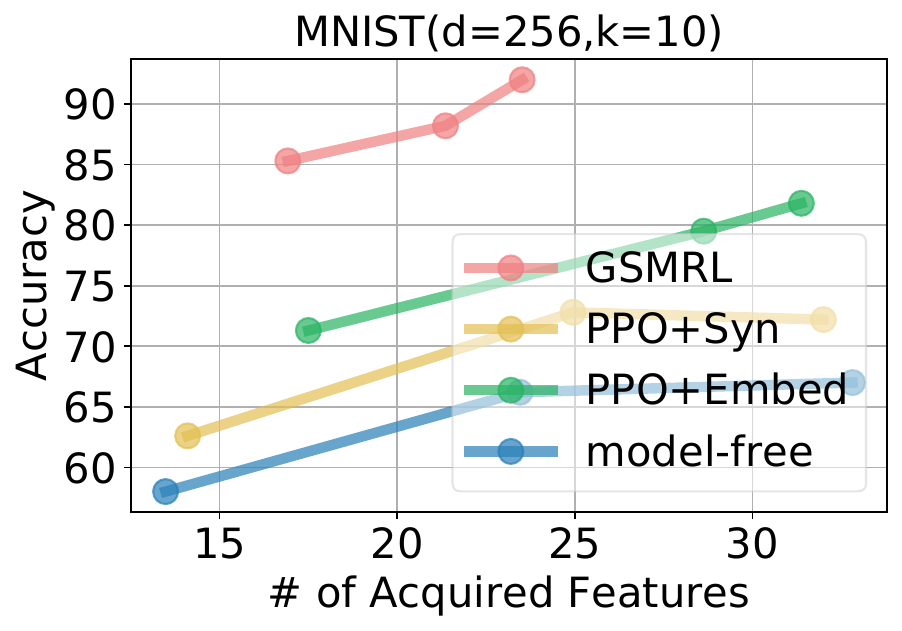}
    \vspace{-20pt}
    \caption{Other model-based approaches.}
    \label{fig:model-based}
    \vspace{-10pt}
\end{wrapfigure}
\paragraph{Model-based Alternatives}
Our GSMRL model combines model-based and model-free approaches into a holistic framework by providing the agent with auxiliary information and intermediate rewards. Here, we study different ways of utilizing the dynamics model. As in ODIN \citep{zannone2019odin}, we utilize class conditioned generative models to generate synthetic trajectories. The agent is then trained with both real and synthetic data (PPO+Syn). Another way of using the model is to extract a semantic embedding from the observations \citep{kumar2018consistent}. We use a pretrained EDDI to embed the current observed features into a 100-dimensional feature vector. 
An agent then takes the embedding as input and predicts the next acquisition (PPO+Embed). Figure \ref{fig:model-based} compares our method with these alternatives. We also present the results from a model-free approach as a baseline. We see our GSMRL outperforms other model-based approaches by a large margin.

\begin{figure}
    \vspace{-5pt}
    \centering
    \begin{minipage}{0.45\linewidth}
    \includegraphics[width=\linewidth]{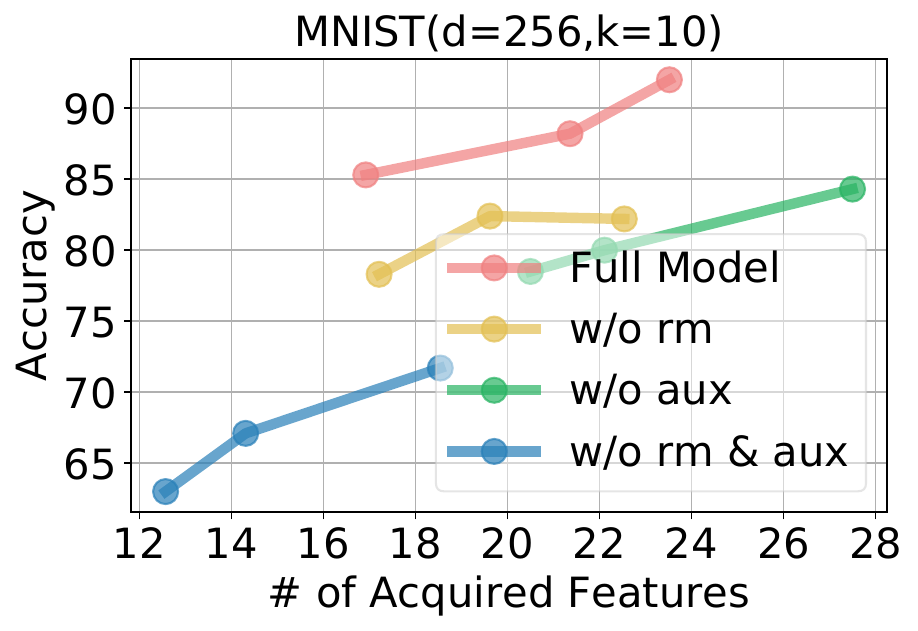}
    \vspace{-22pt}
    \caption{Ablations}
    \label{fig:ablation}
    \vspace{7pt}
    \end{minipage}
    \quad \quad \quad \quad
    \begin{minipage}{0.42\linewidth}
    \includegraphics[width=\linewidth]{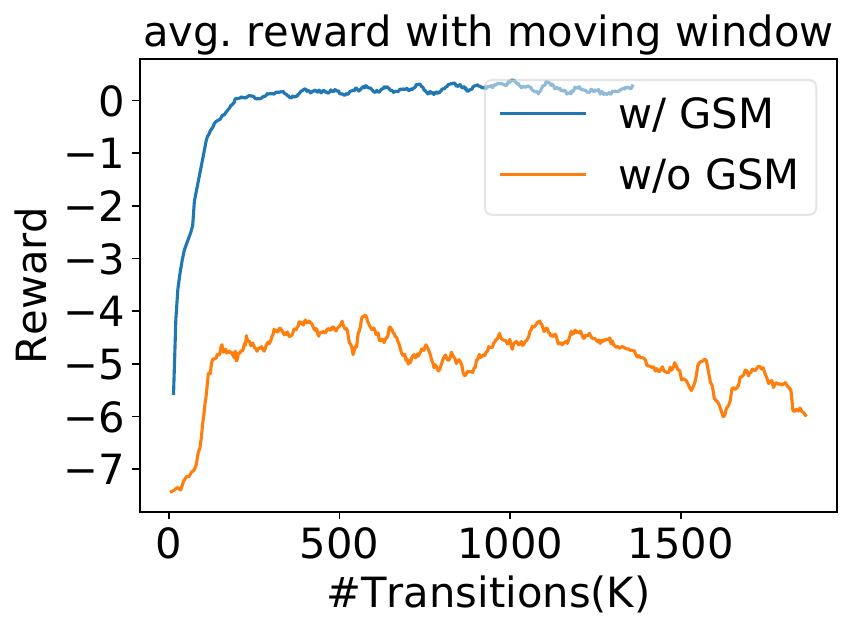}
    \vspace{-22pt}
    \caption{Rewards}
    \label{fig:rewards}
    \end{minipage}
    \vspace{-12pt}
\end{figure}

\paragraph{Surrogate Models}
Our GSMRL method relies on the surrogate model to provide intermediate rewards and auxiliary information. To better understand the contributions each component does to the overall framework, we conduct ablation studies using the MNIST dataset for the classification task. We gradually drop one component from the full model and report the results in Fig.~\ref{fig:ablation}. The `Full Model' uses both intermediate rewards and auxiliary information. 
We then drop the intermediate rewards and denote it as `w/o rm'. The model without auxiliary information is denoted as `w/o aux'. We further drop both components and denote it as `w/o rm \& aux'. From Fig.~\ref{fig:ablation}, we see these two components contribute significantly to the final results. We also compare models with and without the surrogate model. For models without a surrogate model, we train a classifier jointly with the agent as in JAFA. We plot the smoothed rewards using the moving window average during training in Fig.~\ref{fig:rewards}. The agent with a surrogate model not only produces higher and smoother rewards but also converges faster.

\begin{wrapfigure}{r}{0.45\linewidth}
\vspace{-5pt}
\includegraphics[width=\linewidth]{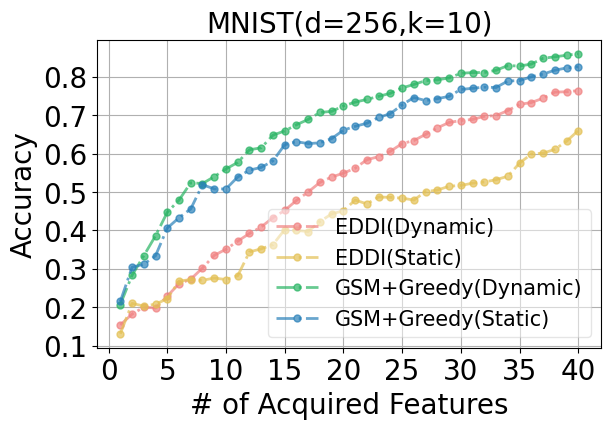}
\vspace{-20pt}
\caption{Compare dynamic and static acquisition strategy using greedy policies.}
\label{fig:mnist_dyn_stc}
\vspace{-15pt}
\end{wrapfigure}
\paragraph{Dynamic vs. Static Acquisition}
Our GSMRL acquires features following a dynamic order where it eventually acquires different features for different instances.
A dynamic acquisition policy should perform better than a static one (i.e., the same set of features are acquired for each instance) since the dynamic policy allows the acquisition to be specifically adapted to the corresponding instance. To verify this is actually the case, we compare the dynamic and static acquisition under a greedy policy for MNIST classification. 
Similar to the dynamic greedy policy, the static acquisition policy acquires the feature with maximum utility at each step, but the utility is averaged over the whole testing set, therefore the same acquisition order is adopted for the whole testing set. Figure \ref{fig:mnist_dyn_stc} shows the classification accuracy for both EDDI and GSM under a greedy acquisition policy. We can see the dynamic policy is always better than the corresponding static one. Furthermore, our GSM with a static acquisition can already outperform dynamic EDDI.

\begin{wrapfigure}{r}{0.45\linewidth}
\vspace{-13pt}
\includegraphics[width=\linewidth]{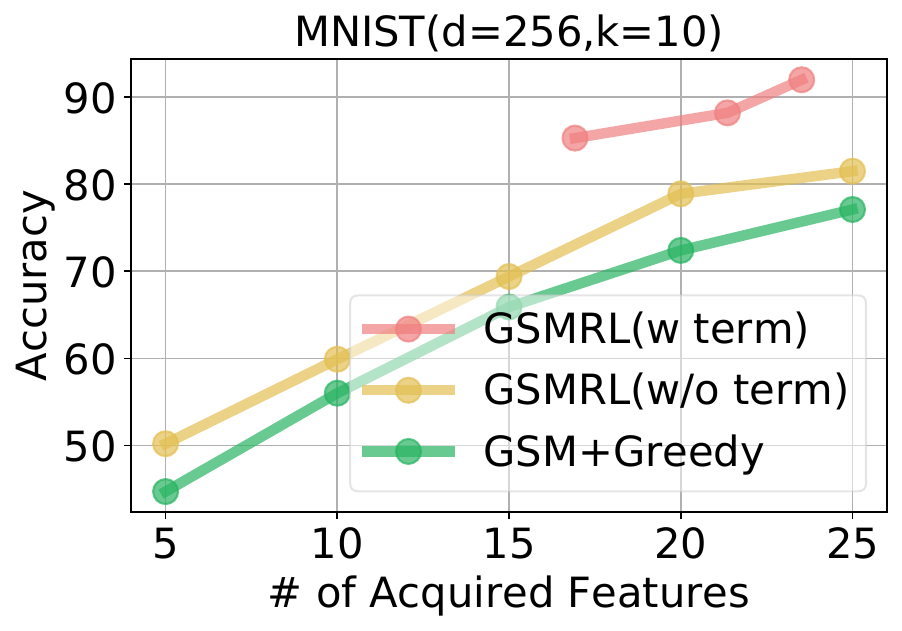}
\vspace{-20pt}
\caption{Acquisition with and without termination.}
\label{fig:mnist_fixed_budget}
\vspace{-15pt}
\end{wrapfigure}
\paragraph{Termination Action}
Our GSMRL will terminate the acquisition process if the agent deems the current acquisition achieves the optimal trade-off between the prediction performance and the acquisition cost.
To evaluate how much the termination action affects the performance and to directly compare with the greedy policies under the same acquisition budget, we conduct an ablation study that removes the termination action (`w/o term') and gives the agent a hard acquisition budget (i.e., forcing the agent to predict after some number of acquisitions).
We can see (Fig.~\ref{fig:mnist_fixed_budget}) GSMRL outperforms the greedy policy under all budgets.
Moreover, we see that the agent is able to correctly assess whether or not more acquisitions are useful since it obtains better performance when it dictates when to predict the termination action.

\subsection{AFA with a Hierarchical Acquisition Policy}\label{sec:exp_hier}
Above, we assessed our AFA framework on several benchmark datasets. 
In this section, we study the proposed hierarchical acquisition policy with a large number of candidate features, where the candidate features are grouped first based on their informativeness to the target. Specifically, we evaluate on MNIST dataset with 784 candidate features (i.e., without any downsampling). Greedy approaches, like EDDI \citep{ma2018eddi} and GSM-Greedy (see Sec.~\ref{sec:afa_greedy}), are computationally too expensive for this dataset; and the model-free approach, JAFA \citep{shim2018joint}, has difficulty converging based on our early experiments. Therefore, we only compare with GSMRL without the action space hierarchy.

\begin{figure}[H]
    \centering
    \subfigure{
    \includegraphics[width=0.45\linewidth]{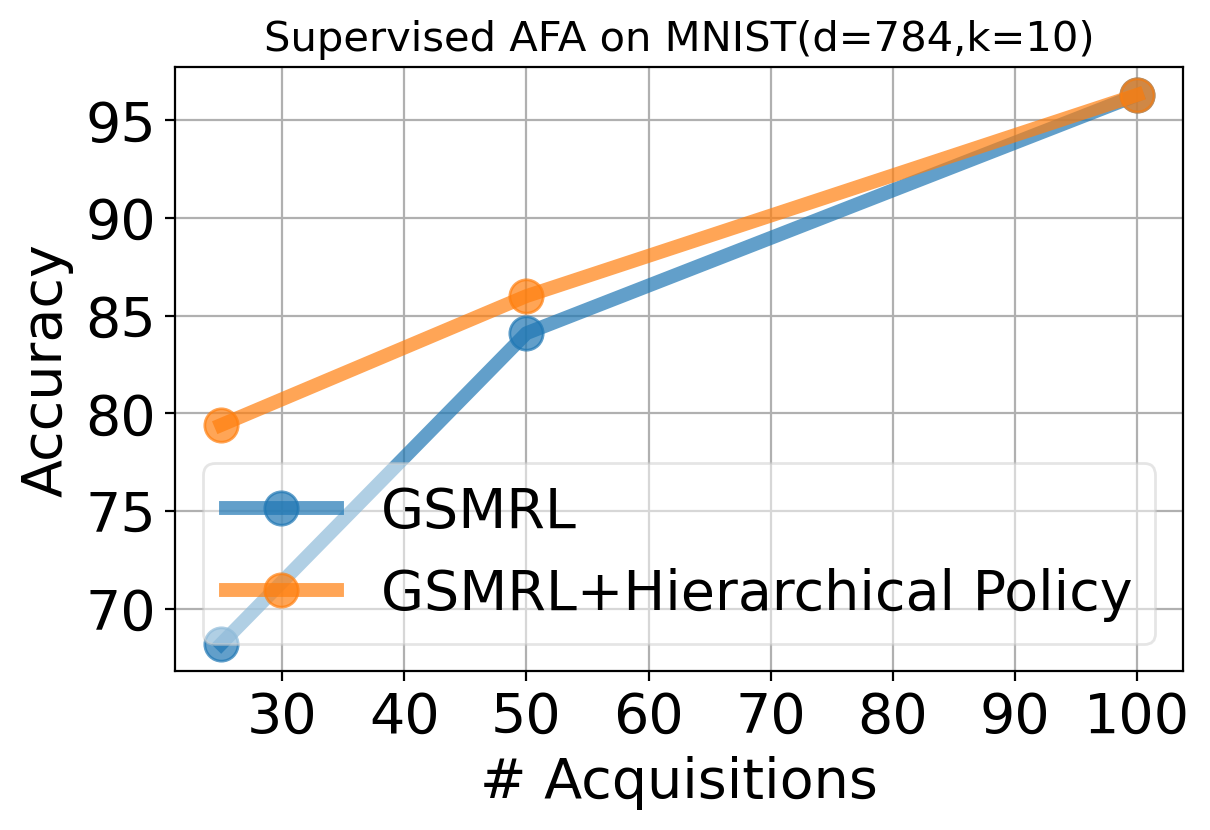}}
    \quad \quad
    \subfigure{
    \includegraphics[width=0.45\linewidth]{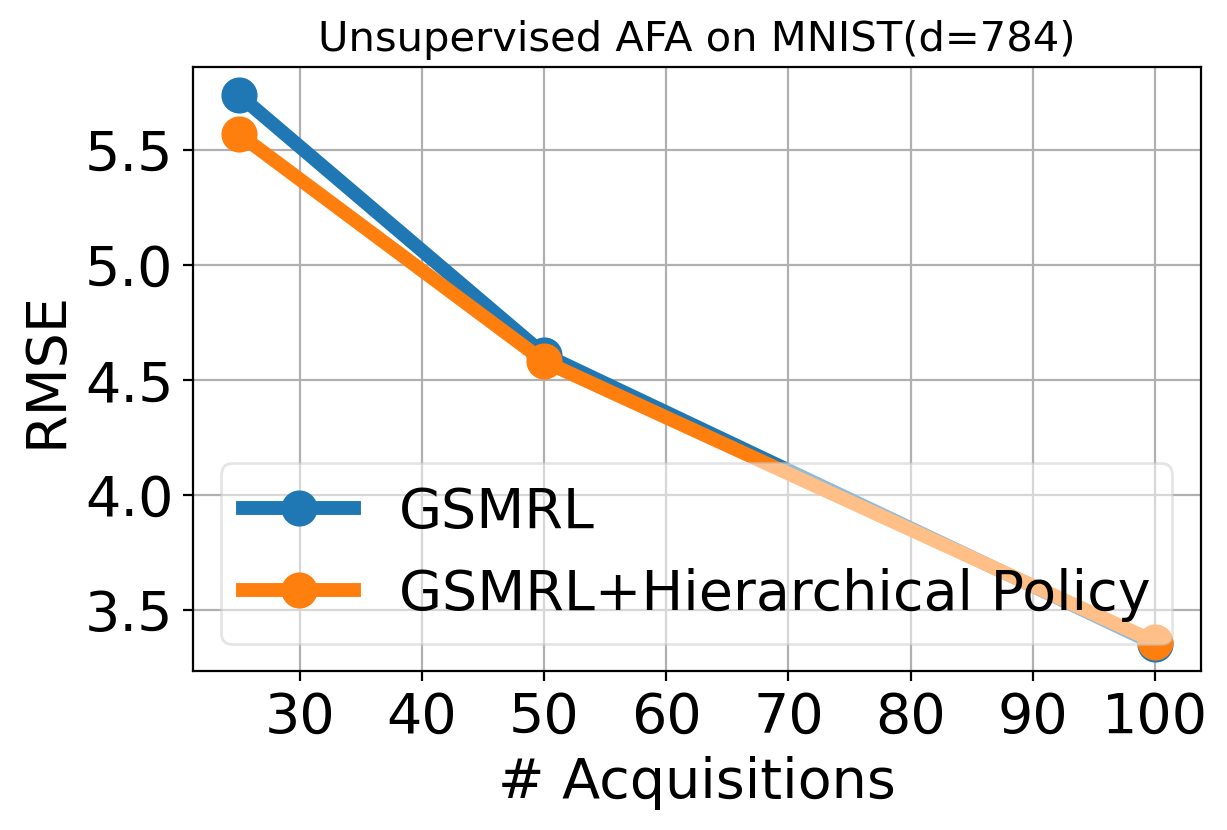}}
    \vspace{-5pt}
    \caption{Compare the hierarchical and non-hierarchical acquisition policies for both supervised and unsupervised tasks.}
    \label{fig:mnist_hier}
\end{figure}

\subsubsection{Numerical Results}
Figure~\ref{fig:mnist_hier} compares the proposed hierarchical acquisition policy with a non-hierarchical one. We evaluate both the supervised and unsupervised tasks. For a fair comparison, we give a hard budget to the agent (i.e., the same number of features are acquired for each instance). We can see the hierarchical policy achieves better performance especially when the acquisition budget is low. We believe our information based groups provide the agent with some additional information to encourage the selection of informative features.

\subsubsection{Ablations}

\begin{figure}[H]

\begin{minipage}{\linewidth}
    \centering
    \includegraphics[width=\linewidth]{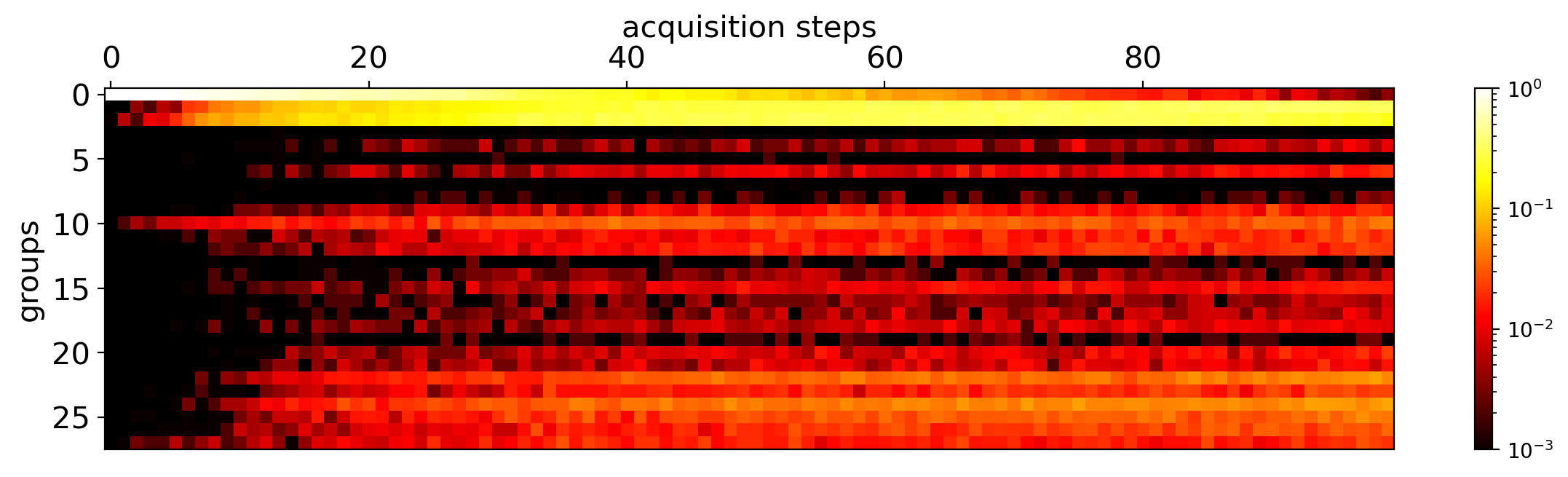}
    \vspace{-15pt}
    \caption{Acquired groups along the acquisition process for MNIST classification. Each column represents the frequency of each group being acquired at the corresponding acquisition step. Groups with smaller index have higher mutual information to the target variable. Frequencies are shown in log scale for better visualization.}
    \vspace{-10pt}
    \label{fig:acquisition}
\end{minipage}

\begin{minipage}{0.43\linewidth}
    \vspace{25pt}
    \centering
    \includegraphics[width=\linewidth]{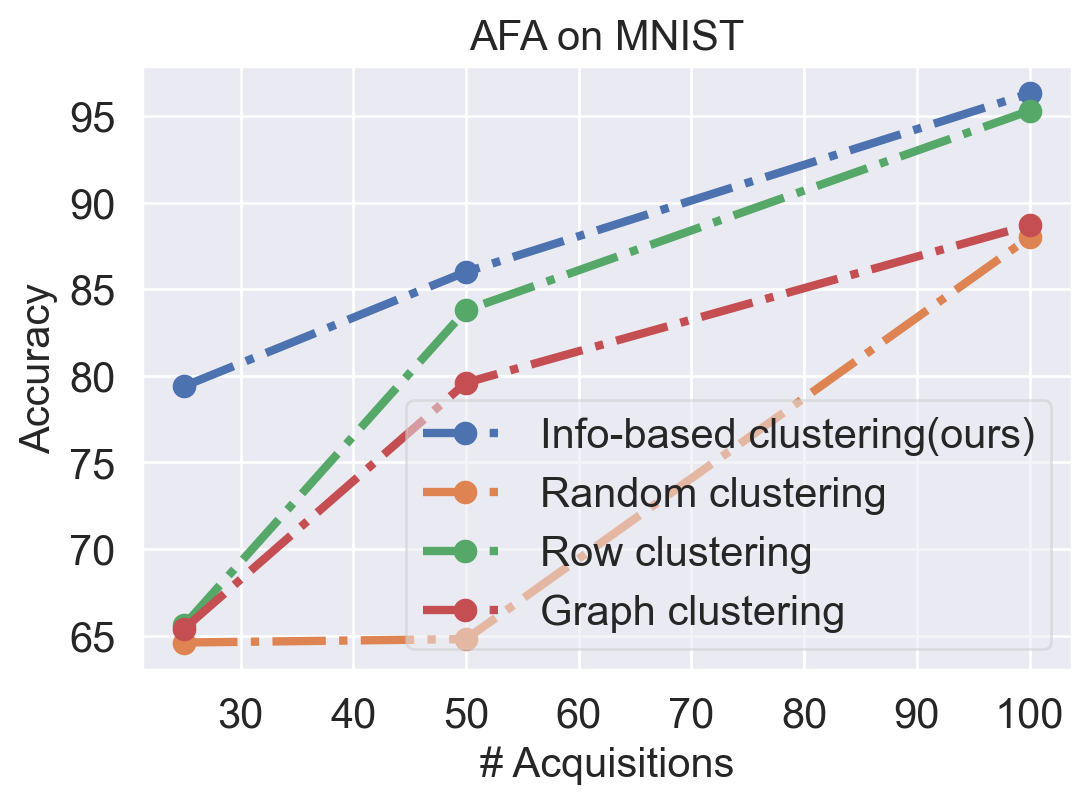}
    \vspace{-10pt}
    \caption{Hierarchical policy with alternative grouping schemes for MNIST classification.}
    \label{fig:clustering}
    \vspace{-25pt}
\end{minipage}
\hfill
\begin{minipage}{0.53\linewidth}
    \vspace{25pt}
    \centering
    \includegraphics[width=\linewidth]{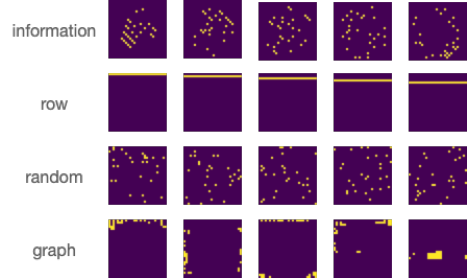}
    \vspace{-10pt}
    \caption{Examples of the clusters from different grouping scheme. For information based clusters, the mutual information decreases from left to right.}
    \label{fig:clusters}
    \vspace{-25pt}
\end{minipage}

\end{figure}

\paragraph{Information based Grouping}
The mutual information based clustering scheme could help the agent navigate the action space, thus simplifying exploration. Figure~\ref{fig:acquisition} presents the acquisition process for MNIST classification. We can see the acquired features concentrate on the marginally informative groups especially at the early stage, which verifies the effectiveness of our action space grouping scheme. From another perspective, our grouping scheme acts like a parametrization that enables the agent to make marginally informative acquisitions when the observed information is scarce.

\paragraph{Alternative Grouping Scheme}
The mutual information based clustering scheme could help the agent eliminate non-informative actions, thus simplifying exploration. Here, we compare with several alternative clustering schemes. Random clustering groups the candidate features into several equal-sized clusters at random. Row clustering groups pixels by their rows. Graph clustering first builds an undirected graph over candidate features and then groups the nodes using spectral clustering methods. We build the graph for MNIST pixels using the graphical lasso method based on the fully observed training set.
Figure~\ref{fig:clustering} presents the AFA performance using different clustering schemes. We can see our information based scheme outperforms other alternatives significantly, especially when the acquisition budget is low. Figure~\ref{fig:clusters} shows several clusters obtained from each grouping scheme. We can see the informative clusters contain pixels around the center part which aligns with our observation that the digits appear around the center of the image.

\subsection{Explainable AFA}\label{sec:exp_xafa}
In this section, we assess our goal-based acquisition policy to provide explanations for the acquisitions. We also employ the hierarchical policy described above (Sec.~\ref{sec:afa_hier}) due to its excellent performance on large action space (see Sec.~\ref{sec:exp_hier}). We further condition the acquisition policy on the selected sub-goals so that the policy is aware of the sub-goals it strives to achieve. We conduct experiments for both supervised and unsupervised tasks and explore explanations both in terms of pairs of classes and sets of clusters.

\subsubsection{Numerical Results}

\begin{figure}[H]
    \centering
    \includegraphics[width=0.7\linewidth]{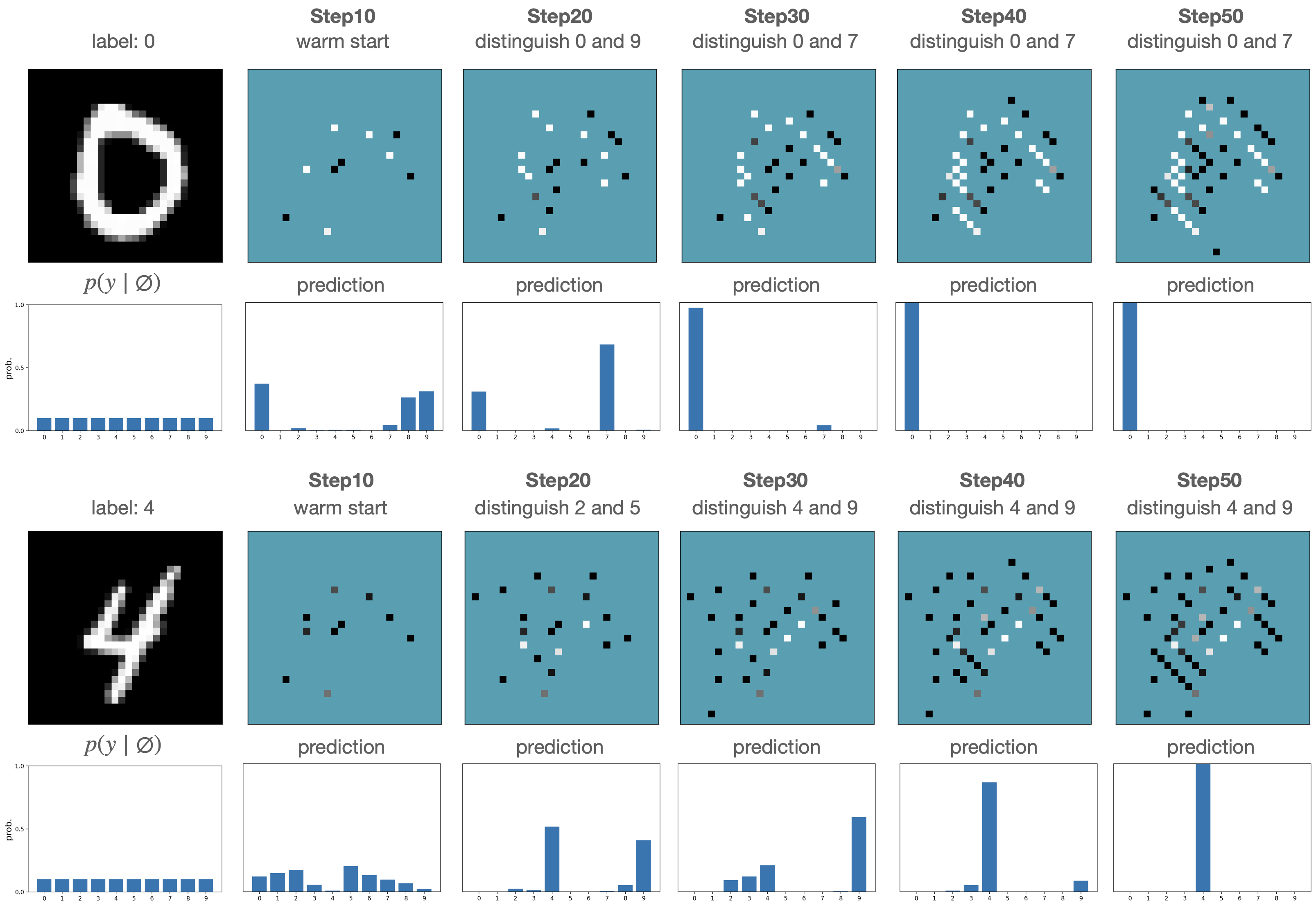}
    \caption{Goal-based acquisition policy for explainable AFA. The pixel values are acquired to achieve the corresponding sub-goal, which is the pair of classes to be distinguished. For each selected sub-goal, 10 pixels are acquired to achieve it. We acquire 10 pixels at the beginning without conditioning on any goal. $p(y \mid \emptyset)$ indicates the prediction probabilities before acquiring any features.}
    \label{fig:mnist_xafa_classes}
\end{figure}

\paragraph{Supervised Explainable AFA}
For the supervised AFA task, we evaluate on MNIST classification. The agent first selects a pair of classes with the highest probabilities as the sub-goal, and then acquires 10 pixels to distinguish these two classes. At the beginning of the acquisition process, the agent has no information about the possible classes. Therefore, we warm start the agent to acquire 10 pixels at the beginning without conditioning on any sub-goals. Figure~\ref{fig:mnist_xafa_classes} presents several examples of the acquisition process. We can see the sub-goal represents the two most uncertain classes for the agent based on its current observation. After the subsequent acquisitions, the uncertainty between those classes is reduced. Therefore, we can explain (attribute) those acquisitions as disambiguating the corresponding sub-goal classes. 

\begin{figure}[H]
    \centering
    \includegraphics[width=0.75\linewidth]{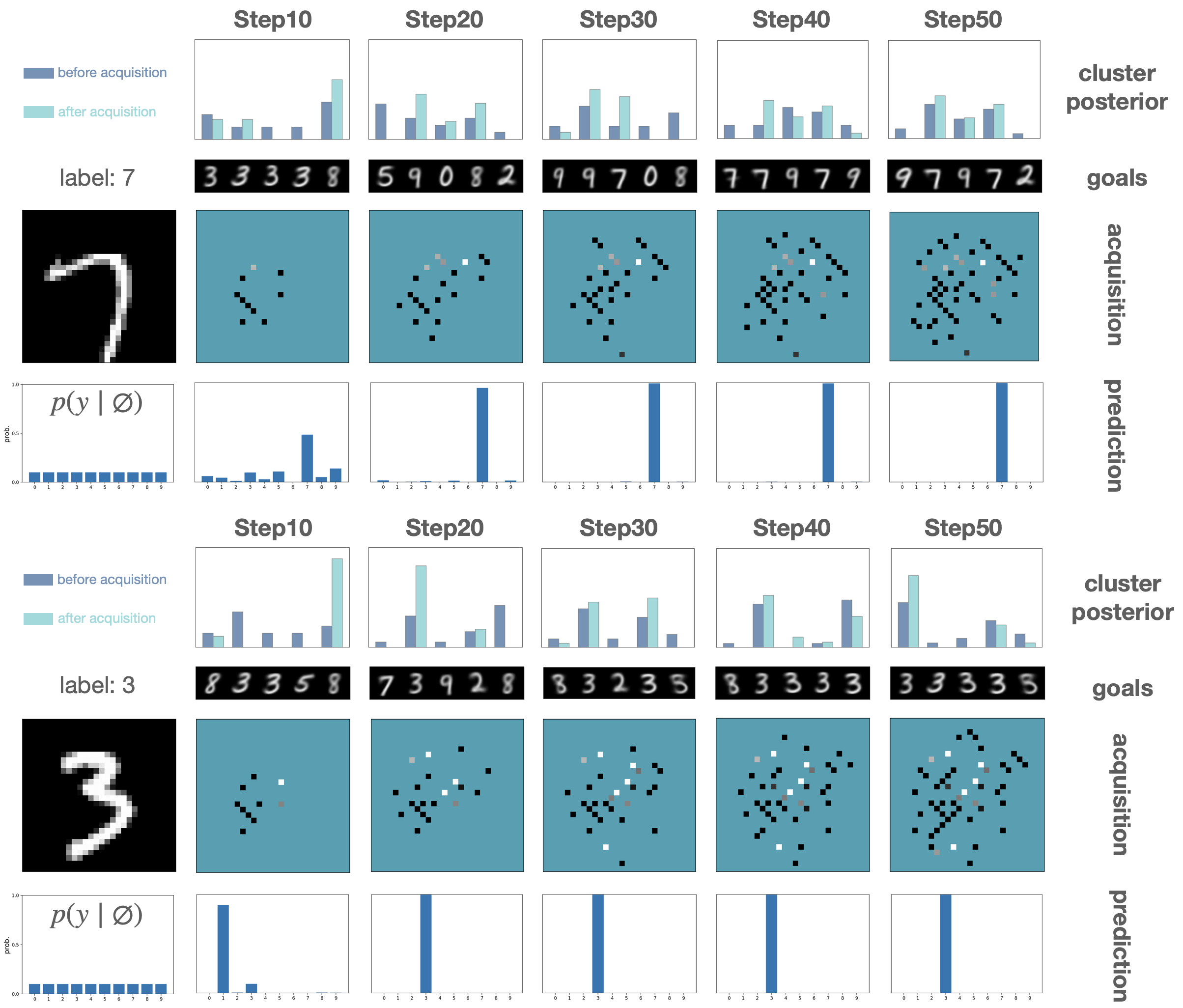}
    \caption{Goal-based acquisition policy for explainable AFA. The sub-goal is a set of clusters with the highest posteriors. 10 pixel values are acquired to achieve the corresponding sub-goal. Above the selected clusters, we show the cluster assignment posteriors before and after the acquisition. $p(y \mid \emptyset)$ indicates the prediction probabilities before acquiring any features.}
    \label{fig:mnist_xafa_clusters}
\end{figure}

Instead of using classes as the sub-goals, Figure~\ref{fig:mnist_xafa_clusters} presents the acquisition process when the goal is a set of clusters that the agent is uncertain about. We train a GMM model over the fully observed training set to obtain 100 clusters in total. The cluster assignment posterior for a partially observed instance is estimated by making 50 imputations and then taking the average of the posterior for imputed instances. Before acquisition, the current 5 clusters with the top partially observed posterior (equation \eqref{eq:xafa_posterior}) are selected as the sub-goal. After selecting a sub-goal, the agent acquires 10 pixels to distinguish the selected clusters using the goal conditioned acquisition policy. From Fig.~\ref{fig:mnist_xafa_clusters}, we can see the agent gradually rules out the incorrect clusters and selects the ones with similar appearances to the groundtruth image. The uncertainty of the posterior for selected clusters also decreases after acquisition for the corresponding goal.

\begin{figure}[H]
    \centering
    \includegraphics[width=0.98\linewidth]{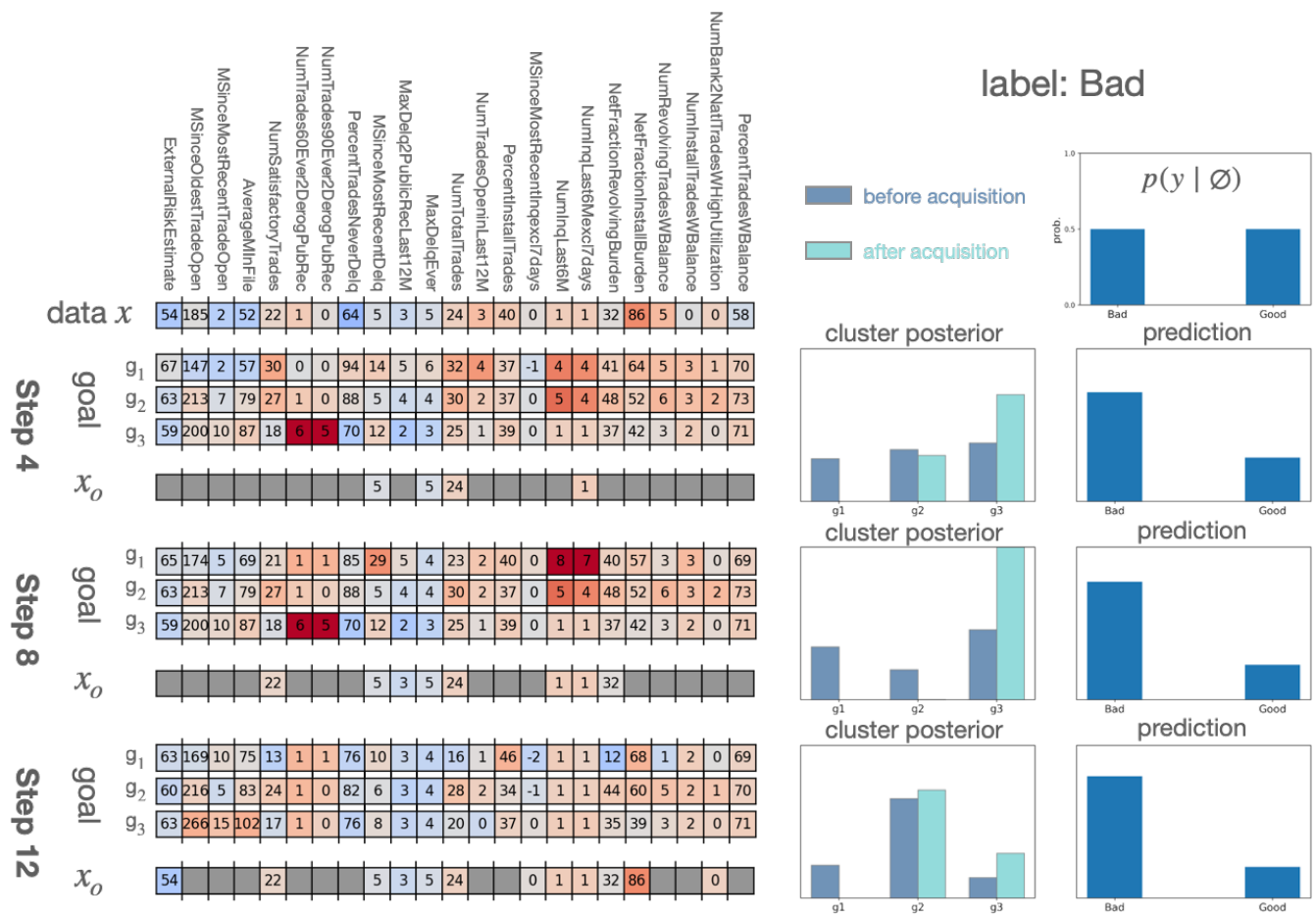}
    \caption{Explainable AFA for HELOC classification. The unobserved features are marked in grey. At each step, we show the three selected clusters ($g_1$, $g_2$ and $g_3$), the posteriors for the selected clusters before and after acquisitions, and the prediction probabilities. $p(y \mid \emptyset)$ indicates the prediction probabilities before acquiring any features.}
    \label{fig:heloc_xafa_clusters}
\end{figure}

In addition to images, we also evaluate on a tabular dataset called HELOC (abbreviated for Home Equity Line of Credit) \citep{fico}, where 23 integer features are given to predict whether the credit is good or bad. We train an ACFlow model over the integer features by adding some small Gaussian noise to them. We also fit a GMM model over the fully observed training set to obtain 50 clusters. The goal is then selected as 3 cluster centers the agent is uncertain about. After selecting a goal, the agent acquires 4 features to distinguish them. Figure~\ref{fig:heloc_xafa_clusters} presents the acquisition process. We can see the entropy of clustering posterior decreases after acquiring the features conditioned on that goal. The agent tends to select features that vary across the three selected cluster centers, which indicates our agent is trying to achieve the specified goal by its acquisitions.

\begin{figure}
    \centering
    \subfigure[MNIST Classification]{
    \includegraphics[width=0.3\linewidth]{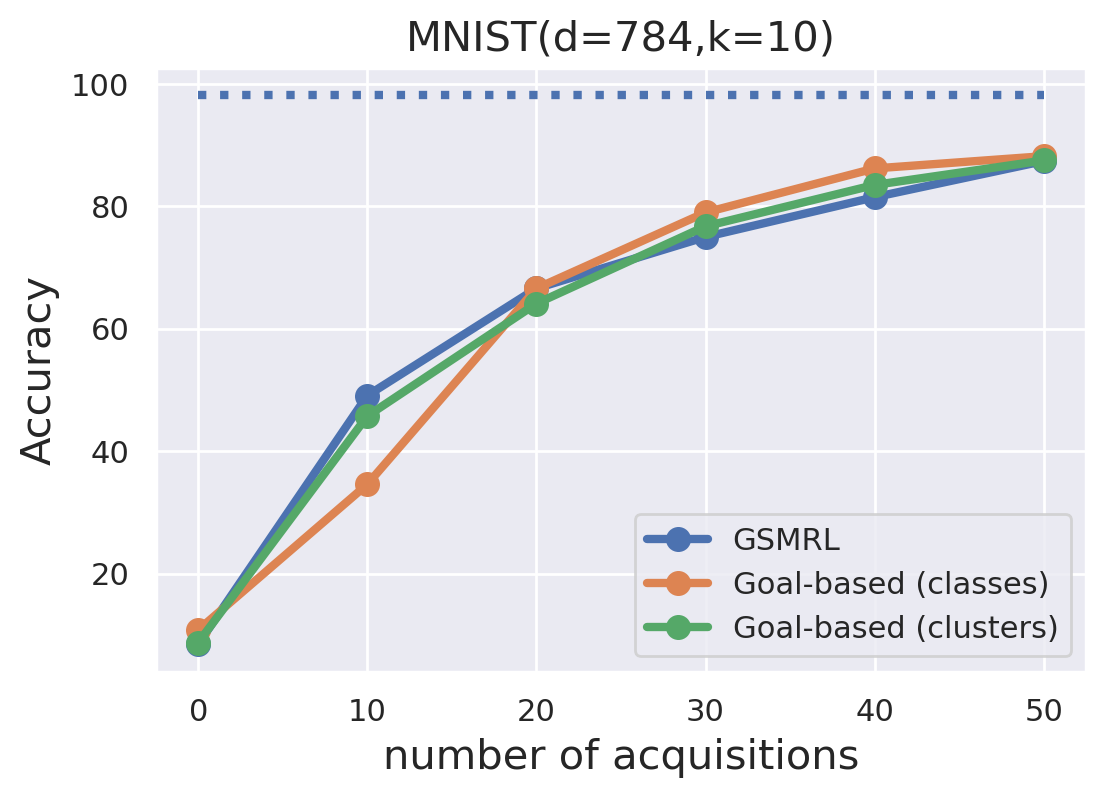}
    \label{fig:mnist_xafa_acc}}
    \subfigure[HELOC Classification]{
    \includegraphics[width=0.3\linewidth]{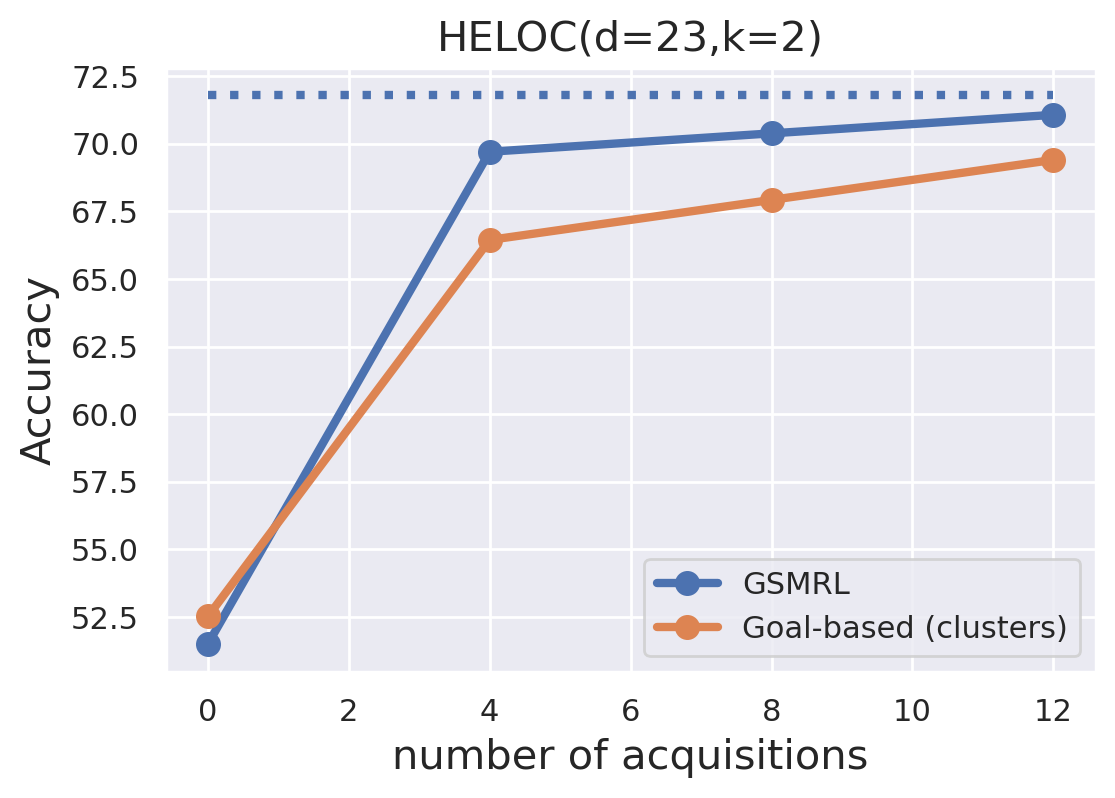}
    \label{fig:heloc_xafa_acc}}
    \subfigure[Unsupervised AFA]{
    \includegraphics[width=0.3\linewidth]{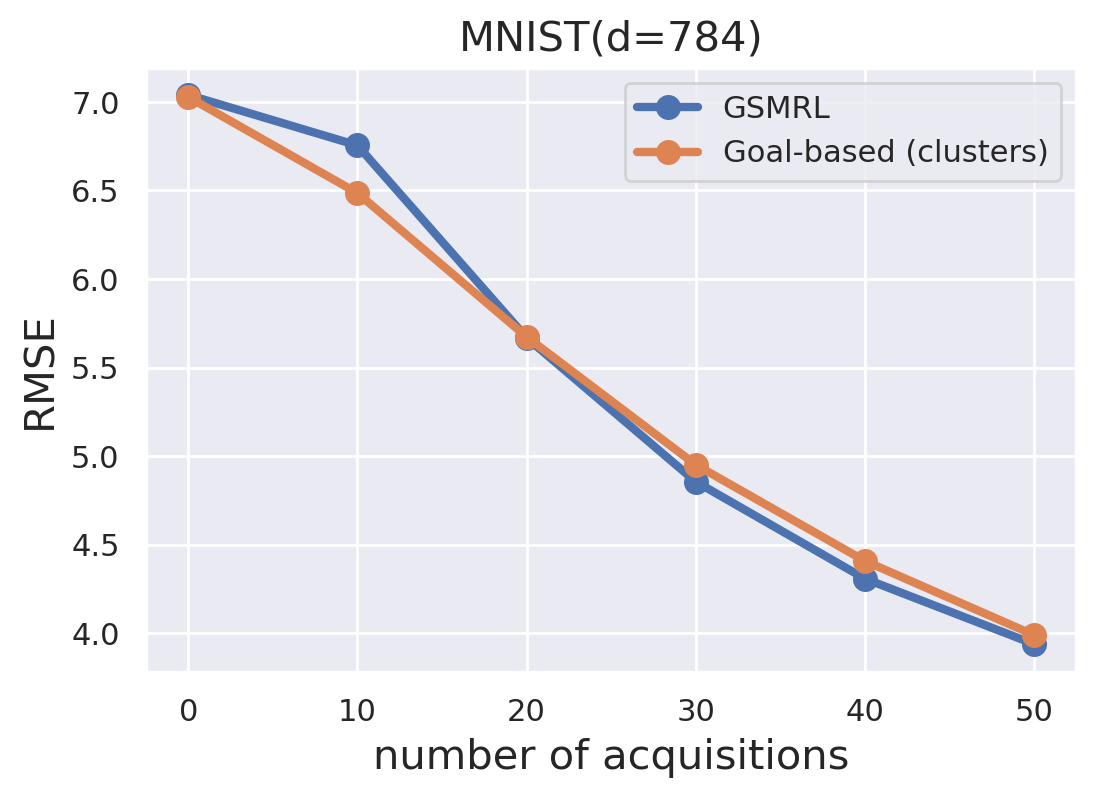}
    \label{fig:mnist_xair_rmse}}
    \caption{Comparing explainable AFA and GSMRL in terms of the prediction performance. The dashed line indicates the upper bound with all features acquired.}
    \label{fig:xafa_acc}
    \vspace{-15pt}
\end{figure}

Figure~\ref{fig:mnist_xafa_acc} and \ref{fig:heloc_xafa_acc} report the classification accuracy along the acquisition process for MNIST and HELOC respectively. We compare to the GSMRL model without a goal. We also report the upper bound when all features are observed. We can see that providing an explanation to acquisitions through sub-goals does not have a large negative impact on performance.

\begin{figure}[H]
    \centering
    \includegraphics[width=0.75\linewidth]{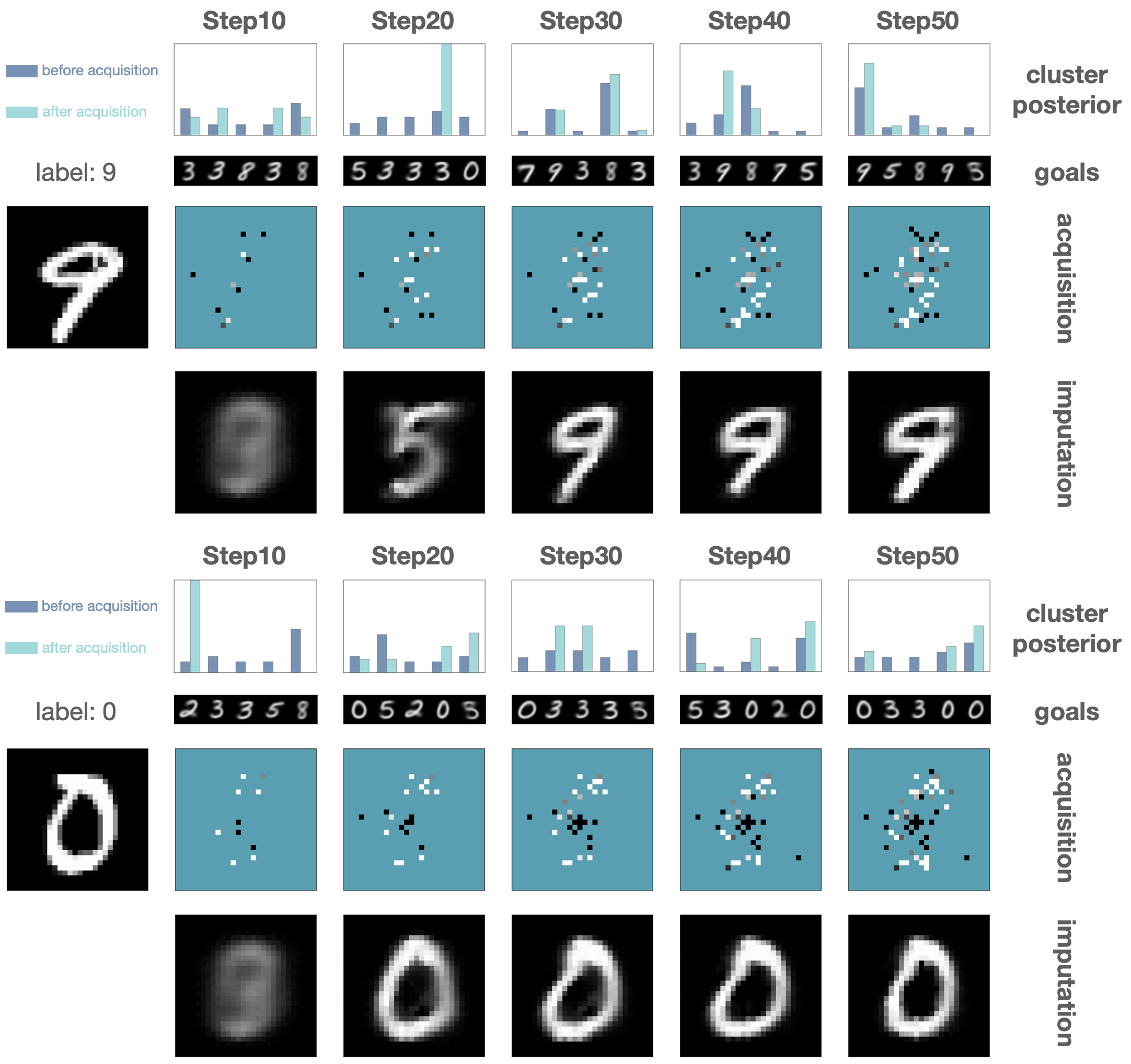}
    \caption{Goal-based acquisition policy for unsupervised AFA. The agent chooses a set of clusters as the sub-goal and acquires 10 pixels to distinguish the clusters.}
    \label{fig:mnist_xair_clusters}
    \vspace{-20pt}
\end{figure}

\paragraph{Unsupervised Explainable AFA}
For unsupervised AFA tasks, we do not have access to the underlying class labels. Therefore, we can only use clusters as the goal. We apply the same GMM model as used in the supervised tasks and select 5 cluster centers as the goal. Figure~\ref{fig:mnist_xair_clusters} shows several examples of the acquisition process. Similar to the supervised AFA, the uncertainty of the clustering posterior decreases after the agent acquires the features for the corresponding goal, and the agent gradually selects the clusters with similar appearances to the groundtruth image. Figure~\ref{fig:mnist_xair_rmse} compares the prediction to GSMRL along the acquisition process. Our goal-based formulation obtains similar performance to the GSMRL model without sub-goals, with additional interpretability.

\subsection{Robust AFA}\label{sec:exp_rafa}
In this section, we evaluate our robust AFA framework on several commonly used OOD detection benchmarks. Our model actively acquires features to predict the target and meanwhile determines whether the input is OOD using only the acquired features. Given that these benchmarks typically have a large number of candidate features, previous AFA approaches cannot be applied directly. We instead compare to a GSMRL algorithm, where candidate features are clustered with our proposed action space grouping technique. We also compare to a simple random acquisition baseline, where a random unobserved feature is acquired at each acquisition step. The random policy is repeated 5 times and the metrics are averaged from different runs. For the classification task, the performance is evaluated by the classification accuracy; for the reconstruction task, the performance is evaluated by the reconstruction MSE. We also detect OOD inputs using the acquired features and report the AUROC scores.

In Section~\ref{sec:exp_afa}, the acquisition procedure is terminated when the agent selects a special termination action, which means each instance could have a different number of features acquired. Although intriguing for practical use, it introduces additional complexity to assessing OOD detection performance.
To simplify the evaluation, we instead specify a fixed acquisition budget (i.e., the number of acquired features). The agent will terminate the acquisition process when it exceeds the prespecified acquisition budget.
Note that it is still feasible to incorporate a termination action into the framework.

\subsubsection{Numerical Results}

\begin{figure}
    \centering
    \begin{minipage}{\linewidth}
    \centering
    \subfigure[MNIST]{\includegraphics[width=0.4\linewidth]{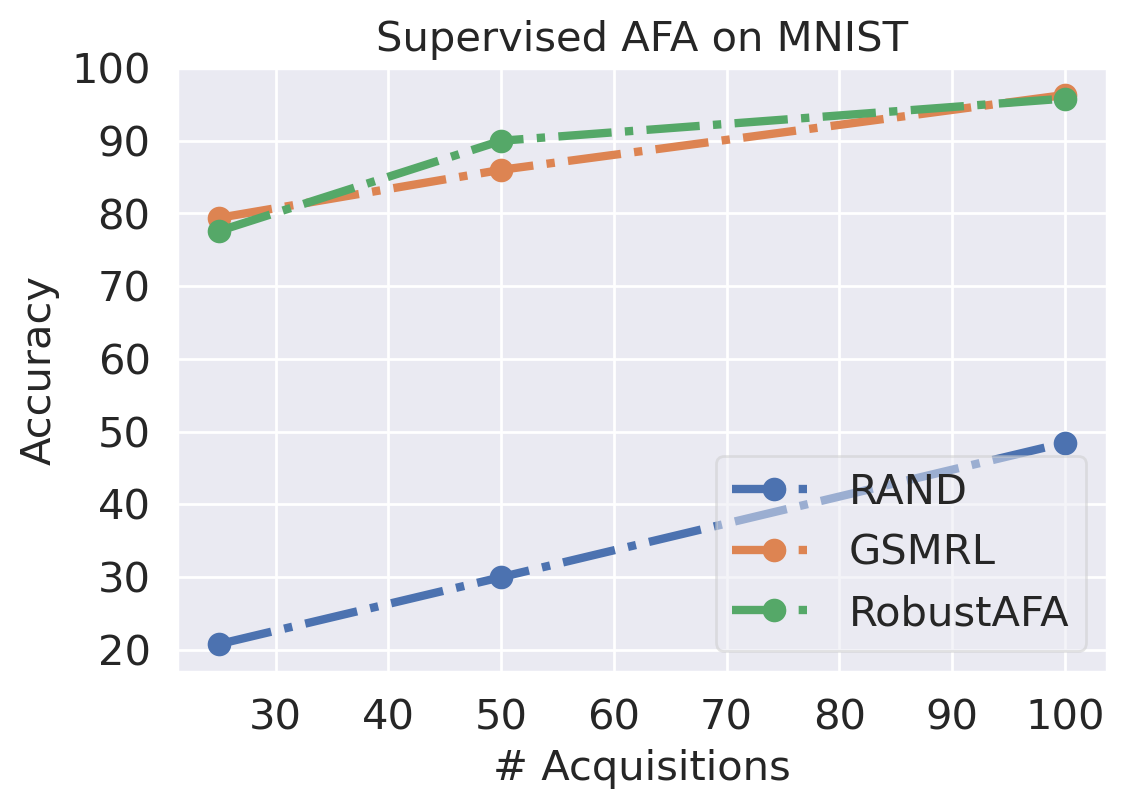}}
    \quad \quad
    \subfigure[FMNIST]{\includegraphics[width=0.4\linewidth]{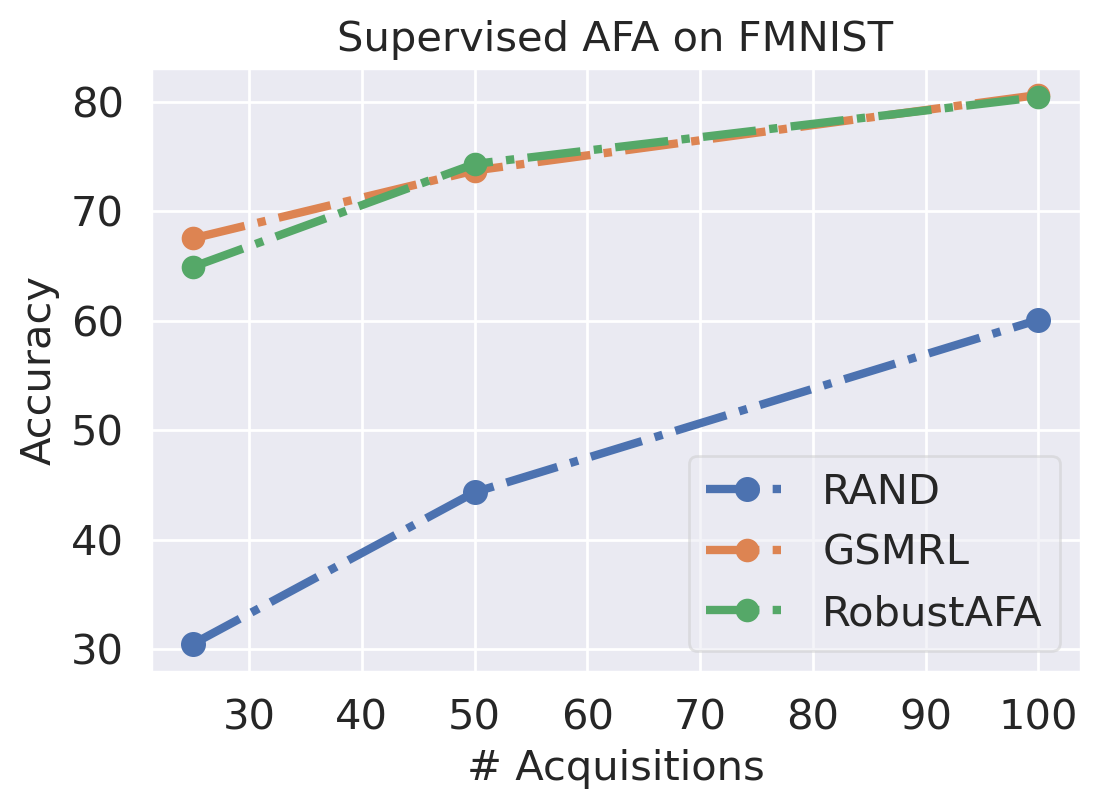}}
    \vspace{-5pt}
    \caption{Classification accuracy for acquiring different number of features.}
    \label{fig:acc}
    \end{minipage}
    \begin{minipage}{\linewidth}
    \vspace{10pt}
    \centering
    \subfigure[MNIST - Omniglot]{\includegraphics[width=0.4\linewidth]{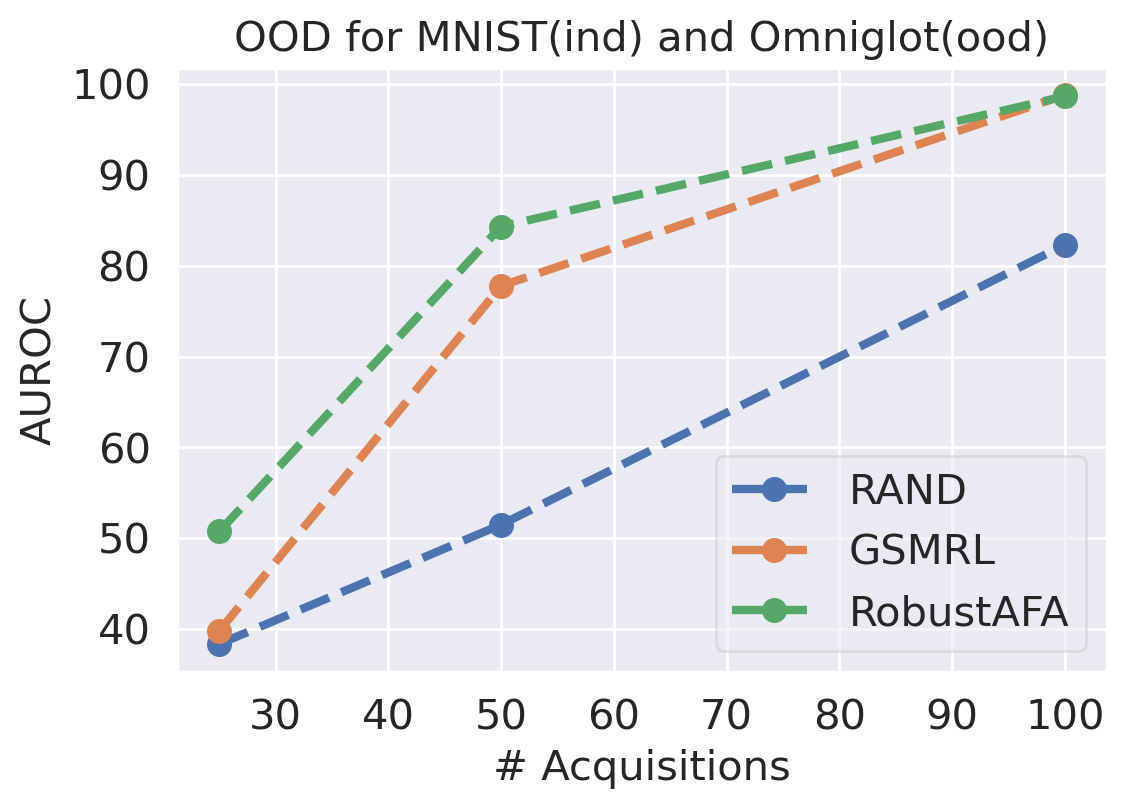}}
    \quad \quad
    \subfigure[FMNIST - MNIST]{\includegraphics[width=0.4\linewidth]{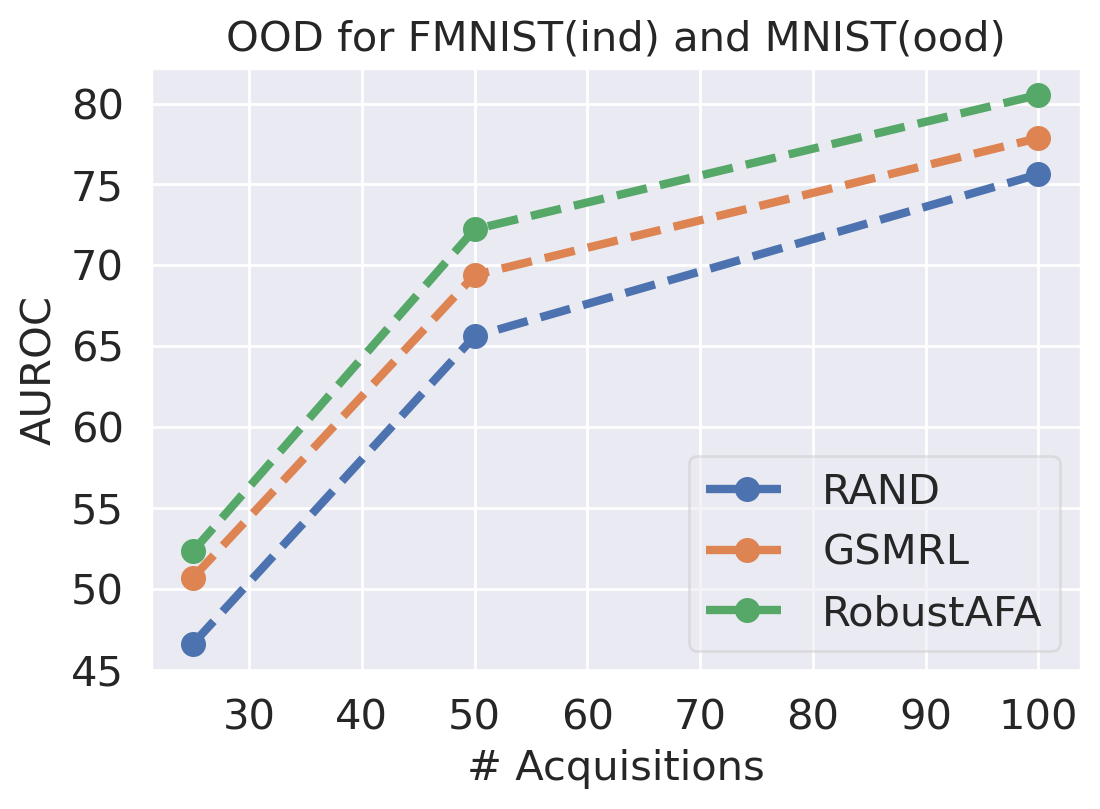}}
    \vspace{-5pt}
    \caption{AUROC for OOD detection with acquired features.}
    \label{fig:auc}
    \end{minipage}
    \begin{minipage}{\linewidth}
    \vspace{15pt}
    \centering
    \includegraphics[width=0.7\linewidth]{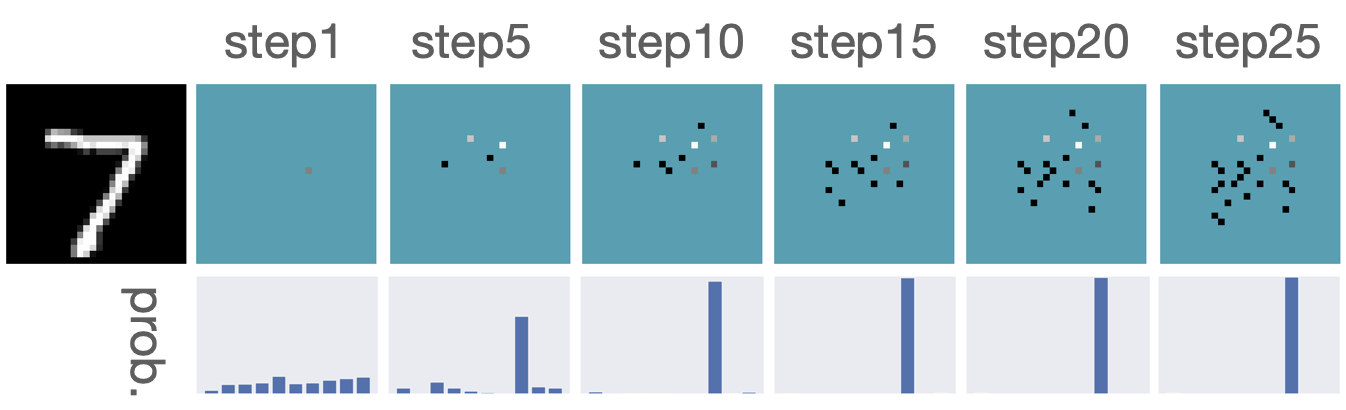}
    \includegraphics[width=0.7\linewidth]{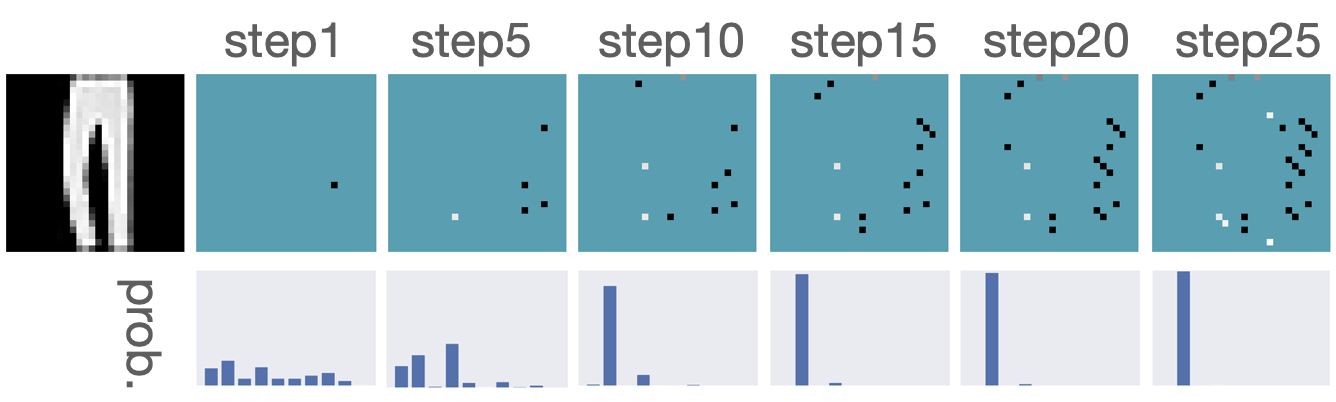}
    \vspace{-10pt}
    \caption{Examples of the acquisition process from our robust AFA framework. The bar charts demonstrate the class prediction probability at the corresponding acquisition step.}
    \label{fig:rafa}
\end{minipage}
\end{figure}

\paragraph{Supervised Robust AFA}
We first evaluate the supervised AFA tasks using several classification datasets, namely MNIST \citep{lecun1998mnist} and FashionMNIST \citep{xiao2017fashion}. The surrogate model is a class conditioned ACFlow \citep{li2019flow}, and the agent is trained to acquire the pixel values.
Figure~\ref{fig:acc} and \ref{fig:auc} report the classification accuracy and OOD detection AUROC respectively. The accuracy is significantly higher for RL approaches than for the random acquisition policy. Although we expect a trade-off between accuracy and OOD detection performance for our robust AFA framework, the accuracy is actually comparable to GSMRL and sometimes even better across the datasets. I.e., we can perform comparably well in the classification whist now being able to detect OOD samples. Meanwhile, the OOD detection performance for our robust AFA framework is significantly improved by enforcing the agent to acquire informative features for OOD identification. 
Figure~\ref{fig:rafa} presents several examples of the acquisition process from our robust AFA framework. We can see the prediction becomes certain after only a few acquisition steps.

\begin{figure}
\begin{minipage}{\linewidth}
    \centering
    \subfigure[MNIST]{\includegraphics[width=0.45\linewidth]{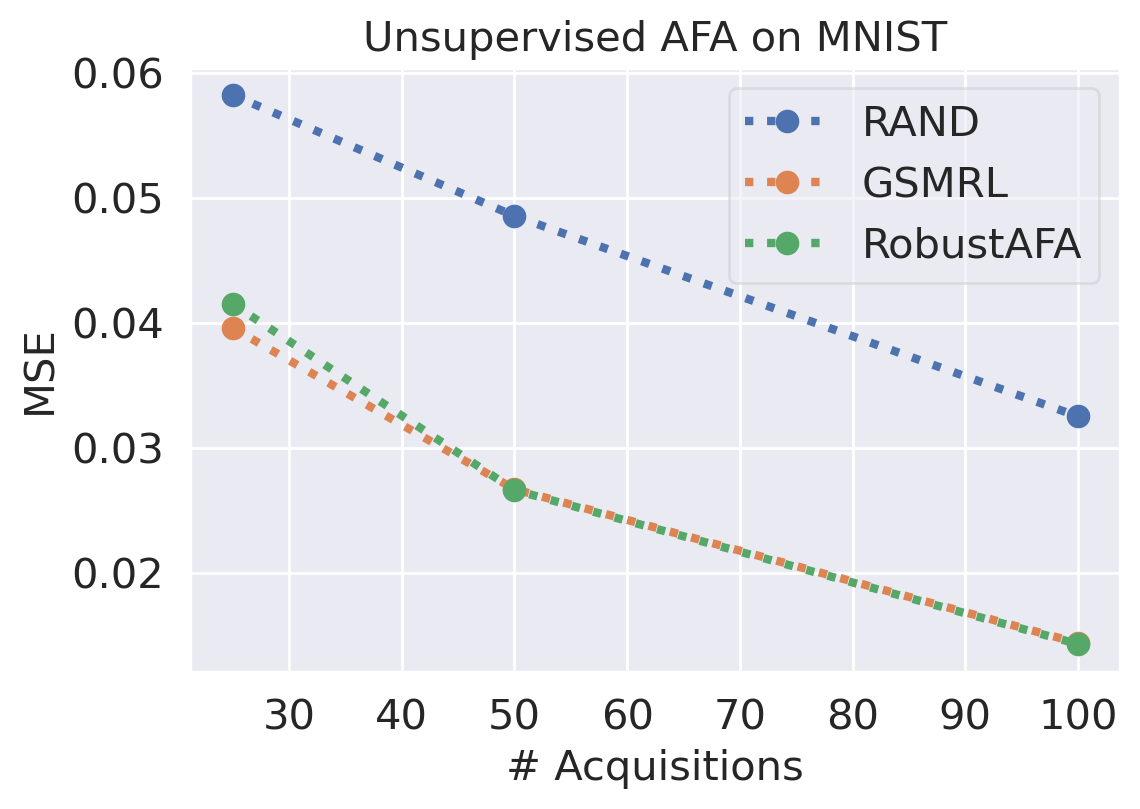}}
    \quad \quad
    \subfigure[FMNIST]{\includegraphics[width=0.45\linewidth]{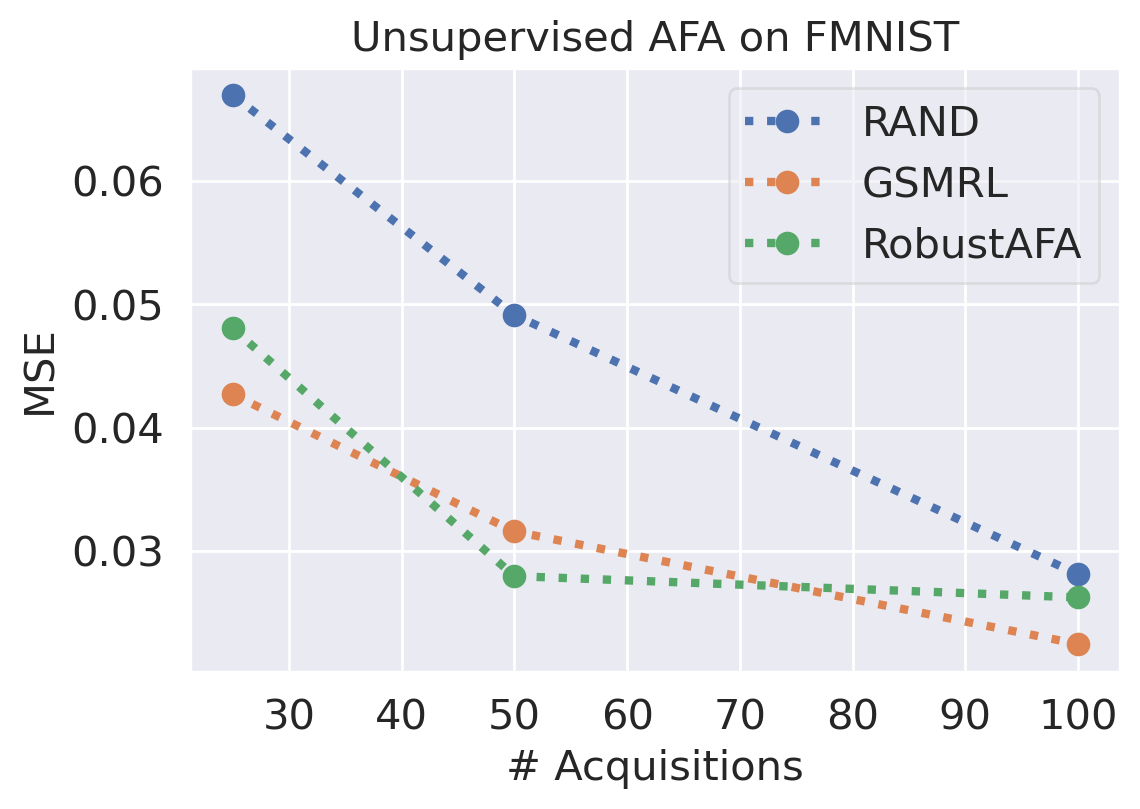}}
    \vspace{-5pt}
    \caption{Reconstruction MSE for unsupervised robust AFA.}
    \label{fig:mse}
\end{minipage}
\begin{minipage}{\linewidth}
    \vspace{15pt}
    \centering
    \subfigure[MNIST - Omniglot]{\includegraphics[width=0.45\linewidth]{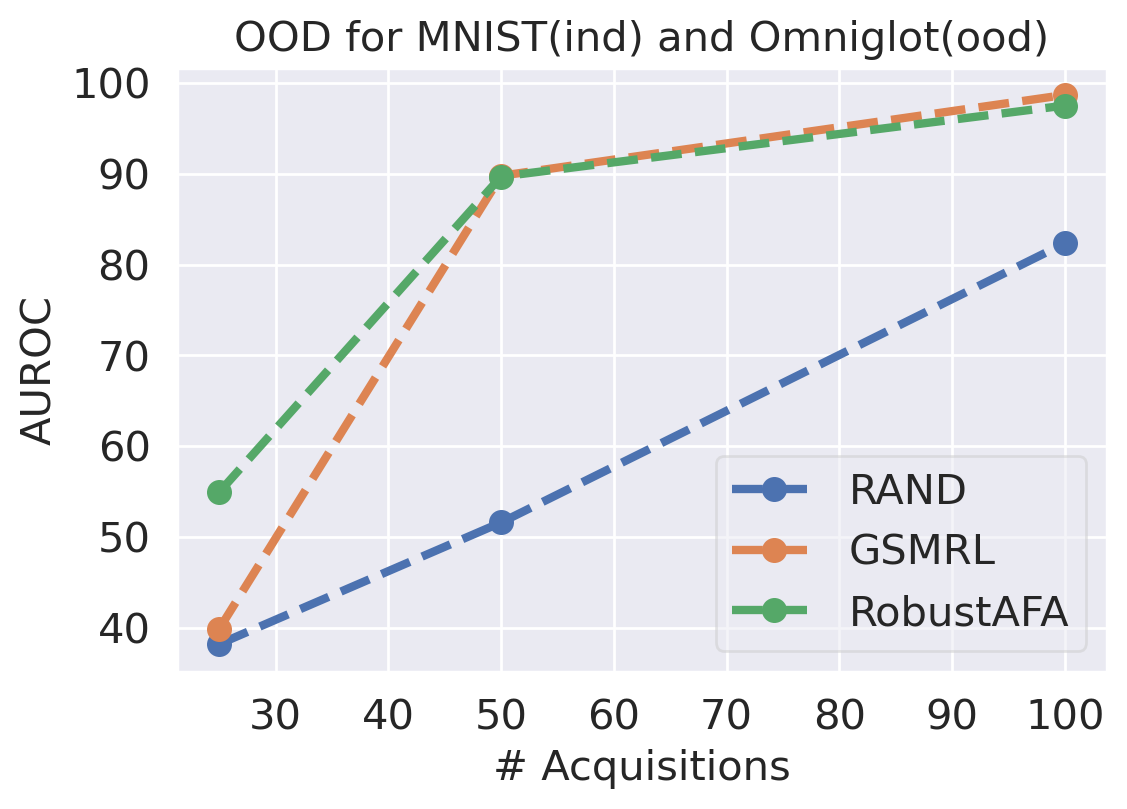}}
    \quad \quad
    \subfigure[FMNIST - MNIST]{\includegraphics[width=0.45\linewidth]{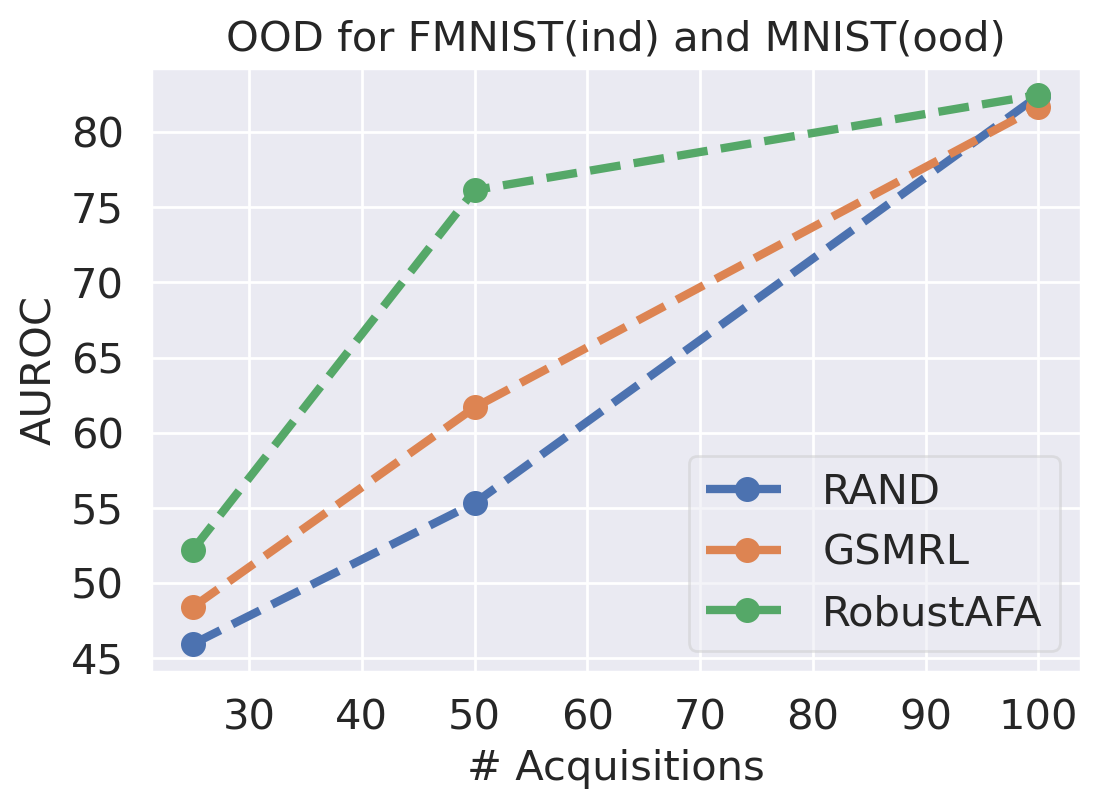}}
    \vspace{-5pt}
    \caption{OOD detection for unsupervised robust AFA.}
    \label{fig:rec}
\end{minipage}
\begin{minipage}{\linewidth}
    \vspace{15pt}
    \centering
    \includegraphics[width=0.7\linewidth]{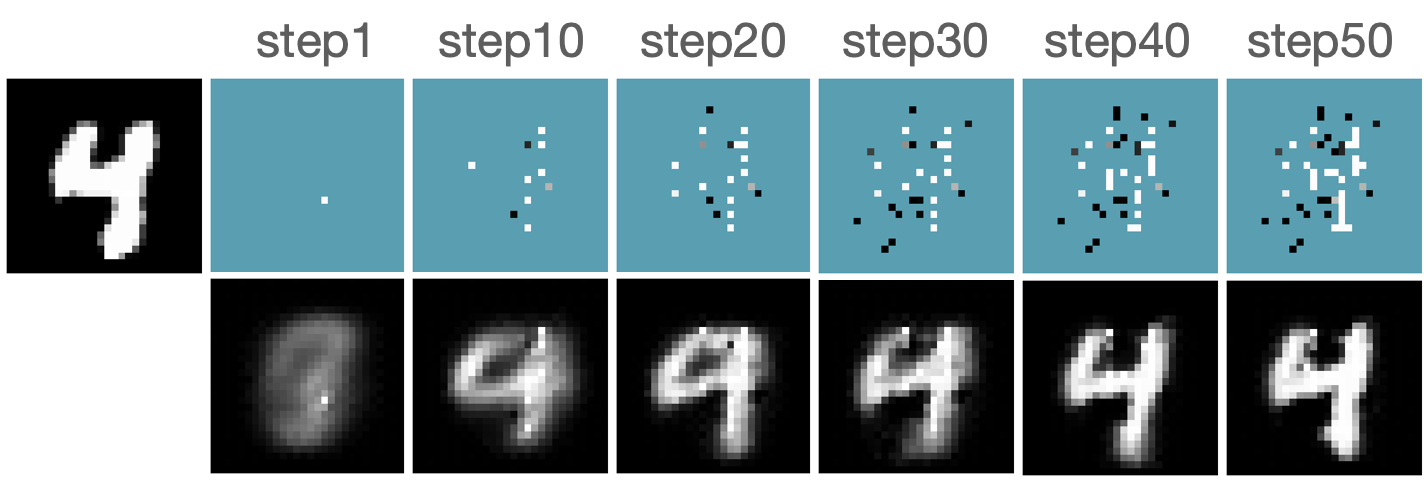}
    \includegraphics[width=0.7\linewidth]{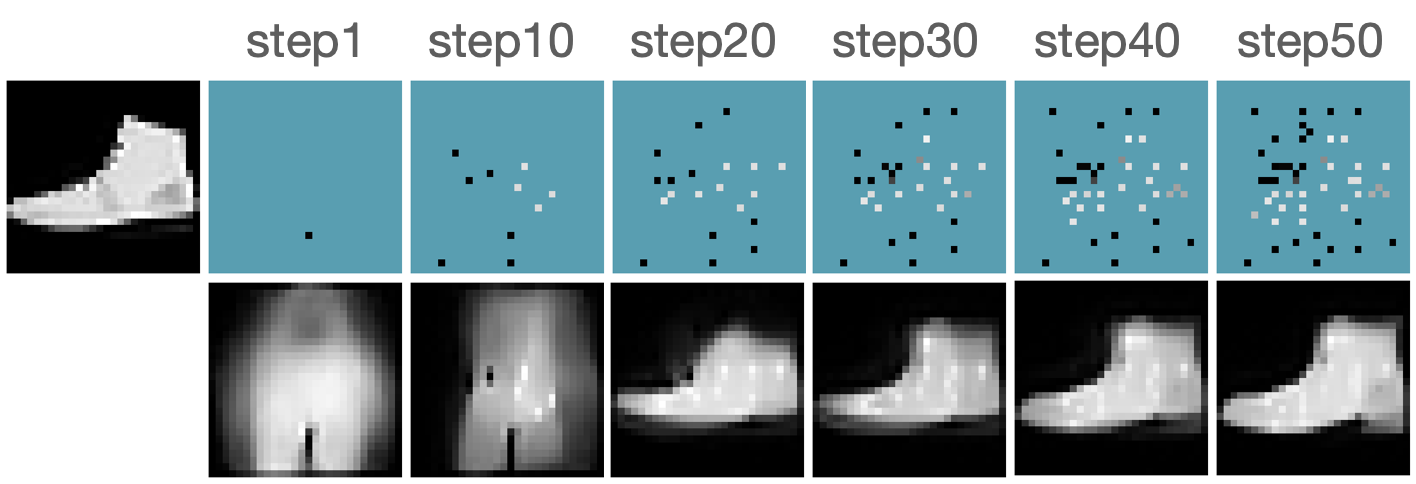}
    \vspace{-5pt}
    \caption{Examples of the acquisition process for unsupervised robust AFA.}
    \label{fig:rair}
    \vspace{-15pt}
\end{minipage}
\end{figure}

\paragraph{Unsupervised Robust AFA}
Next, we evaluate the unsupervised AFA task using MNIST and FashionMNSIT datasets. Figure~\ref{fig:mse} and \ref{fig:rec} report the reconstruction MSE and OOD detection performance respectively using the acquired features. We can see our unsupervised robust AFA framework improves the OOD detection performance significantly, especially when the acquisition budget is low, while the reconstruction MSEs are comparable to GSMRL. Figure~\ref{fig:rair} presents several examples of the acquisition process for unsupervised robust AFA.



\section{Conclusions}\label{sec:conclusion}
In this work, we develop methodology that is able to interact with the environment to obtain new features on-the-fly to better one's prediction. We note that this task is inherently tied to the \emph{information} and \emph{conditional dependencies} that are present in the data. As such, our proposed approach for actively acquiring features is guided by a learned surrogate model over features. First, we showed that it is possible to train a single model, ACFlow, that estimates the exponentially many conditional likelihoods. After, we leveraged these conditional likelihoods in a greedy approach to sequentially acquire features that are anticipated to be informative according to their expected information gain. We further connect the AFA dynamics with the arbitrary conditional distributions and propose a model-based RL formulation, where the dynamics model assists the agent by providing intermediate reward and auxiliary information. Furthermore, we propose a hierarchical acquisition policy, where the candidate features are grouped based on their informativeness, to deal with the potentially large action space of the AFA MDP. Moreover, we increase the interpretability of acquisitions by explicitly setting sub-goals that acquisitions are responsible for. That is, we answer \emph{why} acquisitions are made by explicitly defining sub-targets that acquisitions are disambiguating between. Lastly, we also consider that when applied in real-world scenarios, our AFA agent may be applied on instances that are out-of-distribution w.r.t. the training data. In order to fail gracefully in these cases where predictions may not generalize, we develop a system to jointly perform OOD and AFA. In doing so, we develop a partially observed OOD detector and show that it can be leveraged (without a significant loss to the original accuracy) to detect when features are being acquired on the fly. 
We hope that the totality of this work serves as a foundation for developing intelligent agents that may interact with environments to obtain pertinent information at inference time.
Our future efforts shall include the study of: acquisitions in spatiotemporal domains; acquisitions in natural language domains; and acquisitions in support of general decision making.

\acks{This research was partly funded by grants NSF IIS2133595 and by NIH 1R01AA02687901A1. }


\newpage
\appendix

\renewcommand\thefigure{\thesection.\arabic{figure}}
\renewcommand\thetable{\thesection.\arabic{table}}
\renewcommand{\theequation}{\thesection.\arabic{equation}}

\setcounter{figure}{0} 
\setcounter{subfigure}{0}
\setcounter{equation}{0}

\section{Policy Invariance under Intermediate Rewards}\label{sec:invariance}
Assume the original Markov Decision Process (MDP) without the intermediate rewards is defined as $M = (S,A,T,\gamma,R)$, where $S$ and $A$ are state and action spaces, $T$ is the state transition probabilities, $\gamma$ is the discount factor, and $R$ is the rewards. When we introduce the intermediate rewards $R_m$, the MDP is modified to $M'=(S,A,T,\gamma,R')$, where $R'=R+R_m$. The following theory provides a sufficient and necessary condition for the modified MDP $M'$ to achieve the same optimal policy as the original MDP $M$.

\noindent
{\bf Theorem 1}
The modified MDP $M'=(S,A,T,\gamma,R+F)$ with any shaping reward function $F$ is guaranteed to be consistent with the optimal policy of the original MDP $M=(S,A,T,\gamma,R)$ if the shaping function $F$ have the following form
\begin{equation}
    F(s,a,s') = \gamma \Phi(s') - \Phi(s),
\end{equation}
where $\Phi: S \rightarrow \mathbb{R}$ is a potential function evaluated on states. For infinite-state case (i.e., the state space is an infinite set) the potential function is additionally required to be bounded.

\noindent
{\bf Proof}
Please refer to \citet{ng1999policy} for detailed proof.

From the above theorem, we can see our intermediate rewards in \eqref{eq:interm_reward} is a potential based shaping function and the potential function is $\Phi(s) = - H(y \mid s)$.
For classification task where $y \in \mathcal{Y} = \{c_1, _2,\dots,c_{|\mathcal{Y}|}\}$ is a discrete variable, the entropy is naturally bounded, i.e., $0 \leq H(y \mid s) \leq log \vert \mathcal{Y} \vert$, where $\vert \mathcal{Y} \vert$ is the cardinality of the label space.
For regression task where $y \in \mathbb{R}$, the entropy is bounded by $0 \leq H(y \mid s) \leq H(y \mid \emptyset)$. The upper bound $H(y \mid \emptyset)$ is determined by the given surrogate model. Similarly, for the intermediate rewards in \eqref{eq:interm_reward_unsup}, the potential function $\Phi(s) = \frac{\log p(x_u \mid s)}{|u|}$ is also bounded for a given surrogate model.

\vskip 0.2in
\bibliography{main}

\begin{thebibliography}{87}
\providecommand{\natexlab}[1]{#1}
\providecommand{\url}[1]{\texttt{#1}}
\expandafter\ifx\csname urlstyle\endcsname\relax
  \providecommand{\doi}[1]{doi: #1}\else
  \providecommand{\doi}{doi: \begingroup \urlstyle{rm}\Url}\fi

\bibitem[Abadi et~al.(2016)Abadi, Barham, Chen, Chen, Davis, Dean, Devin,
  Ghemawat, Irving, Isard, et~al.]{abadi2016tensorflow}
Mart{\'\i}n Abadi, Paul Barham, Jianmin Chen, Zhifeng Chen, Andy Davis, Jeffrey
  Dean, Matthieu Devin, Sanjay Ghemawat, Geoffrey Irving, Michael Isard, et~al.
\newblock $\{$TensorFlow$\}$: A system for $\{$Large-Scale$\}$ machine
  learning.
\newblock In \emph{12th USENIX symposium on operating systems design and
  implementation (OSDI 16)}, pages 265--283, 2016.

\bibitem[Aloimonos et~al.(1988)Aloimonos, Weiss, and
  Bandyopadhyay]{aloimonos1988active}
John Aloimonos, Isaac Weiss, and Amit Bandyopadhyay.
\newblock Active vision.
\newblock \emph{International journal of computer vision}, 1\penalty0
  (4):\penalty0 333--356, 1988.

\bibitem[Bagnall et~al.(2017)Bagnall, Lines, Bostrom, Large, and
  Keogh]{bagnall2017great}
Anthony Bagnall, Jason Lines, Aaron Bostrom, James Large, and Eamonn Keogh.
\newblock The great time series classification bake off: a review and
  experimental evaluation of recent algorithmic advances.
\newblock \emph{Data Mining and Knowledge Discovery}, 31\penalty0 (3):\penalty0
  606--660, 2017.

\bibitem[Bajcsy(1988)]{bajcsy1988active}
Ruzena Bajcsy.
\newblock Active perception.
\newblock \emph{Proceedings of the IEEE}, 76\penalty0 (8):\penalty0 966--1005,
  1988.

\bibitem[Belghazi et~al.(2019)Belghazi, Oquab, and
  Lopez-Paz]{belghazi2019learning}
Mohamed Belghazi, Maxime Oquab, and David Lopez-Paz.
\newblock Learning about an exponential amount of conditional distributions.
\newblock \emph{Advances in Neural Information Processing Systems}, 32, 2019.

\bibitem[Bergamin et~al.(2022)Bergamin, Mattei, Havtorn, Senetaire, Schmutz,
  Maal{\o}e, Hauberg, and Frellsen]{bergamin2022model}
Federico Bergamin, Pierre-Alexandre Mattei, Jakob Havtorn, Hugo Senetaire, Hugo
  Schmutz, Lars Maal{\o}e, S{\o}ren Hauberg, and Jes Frellsen.
\newblock Model-agnostic out-of-distribution detection using combined
  statistical tests.
\newblock In \emph{AISTATS 2022-25th International Conference on Artificial
  Intelligence and Statistics}, volume 151, 2022.

\bibitem[Bishop(1994)]{bishop1994novelty}
Christopher~M Bishop.
\newblock Novelty detection and neural network validation.
\newblock \emph{IEE Proceedings-Vision, Image and Signal processing},
  141\penalty0 (4):\penalty0 217--222, 1994.

\bibitem[Blundell et~al.(2015)Blundell, Cornebise, Kavukcuoglu, and
  Wierstra]{blundell2015weight}
Charles Blundell, Julien Cornebise, Koray Kavukcuoglu, and Daan Wierstra.
\newblock Weight uncertainty in neural network.
\newblock In \emph{International Conference on Machine Learning}, pages
  1613--1622. PMLR, 2015.

\bibitem[Breiman(2001)]{breiman2001random}
Leo Breiman.
\newblock Random forests.
\newblock \emph{Machine learning}, 45\penalty0 (1):\penalty0 5--32, 2001.

\bibitem[Cai et~al.(2018)Cai, Luo, Wang, and Yang]{cai2018feature}
Jie Cai, Jiawei Luo, Shulin Wang, and Sheng Yang.
\newblock Feature selection in machine learning: A new perspective.
\newblock \emph{Neurocomputing}, 300:\penalty0 70--79, 2018.

\bibitem[Casalicchio et~al.(2018)Casalicchio, Molnar, and
  Bischl]{casalicchio2018visualizing}
Giuseppe Casalicchio, Christoph Molnar, and Bernd Bischl.
\newblock Visualizing the feature importance for black box models.
\newblock In \emph{Joint European Conference on Machine Learning and Knowledge
  Discovery in Databases}, pages 655--670. Springer, 2018.

\bibitem[Chai et~al.(2004)Chai, Deng, Yang, and Ling]{chai2004test}
Xiaoyong Chai, Lin Deng, Qiang Yang, and Charles~X Ling.
\newblock Test-cost sensitive naive bayes classification.
\newblock In \emph{Fourth IEEE International Conference on Data Mining
  (ICDM'04)}, pages 51--58. IEEE, 2004.

\bibitem[Chebotar et~al.(2017)Chebotar, Hausman, Zhang, Sukhatme, Schaal, and
  Levine]{chebotar2017combining}
Yevgen Chebotar, Karol Hausman, Marvin Zhang, Gaurav Sukhatme, Stefan Schaal,
  and Sergey Levine.
\newblock Combining model-based and model-free updates for trajectory-centric
  reinforcement learning.
\newblock In \emph{International conference on machine learning}, pages
  703--711. PMLR, 2017.

\bibitem[Cheng et~al.(2018)Cheng, Agarwal, and
  Fragkiadaki]{cheng2018reinforcement}
Ricson Cheng, Arpit Agarwal, and Katerina Fragkiadaki.
\newblock Reinforcement learning of active vision for manipulating objects
  under occlusions.
\newblock In \emph{Conference on Robot Learning}, pages 422--431. PMLR, 2018.

\bibitem[Choi et~al.(2018)Choi, Jang, and Alemi]{choi2018waic}
Hyunsun Choi, Eric Jang, and Alexander~A Alemi.
\newblock Waic, but why? generative ensembles for robust anomaly detection.
\newblock \emph{arXiv preprint arXiv:1810.01392}, 2018.

\bibitem[Dinh et~al.(2014)Dinh, Krueger, and Bengio]{dinh2014nice}
Laurent Dinh, David Krueger, and Yoshua Bengio.
\newblock Nice: Non-linear independent components estimation.
\newblock \emph{arXiv preprint arXiv:1410.8516}, 2014.

\bibitem[Dinh et~al.(2016)Dinh, Sohl-Dickstein, and Bengio]{dinh2016density}
Laurent Dinh, Jascha Sohl-Dickstein, and Samy Bengio.
\newblock Density estimation using real nvp.
\newblock \emph{arXiv preprint arXiv:1605.08803}, 2016.

\bibitem[Douglas et~al.(2017)Douglas, Zarov, Gourgoulias, Lucas, Hart, Baker,
  Sahani, Perov, and Johri]{douglas2017marginalizer}
Laura Douglas, Iliyan Zarov, Konstantinos Gourgoulias, Chris Lucas, Chris Hart,
  Adam Baker, Maneesh Sahani, Yura Perov, and Saurabh Johri.
\newblock A universal marginalizer for amortized inference in generative
  models.
\newblock \emph{arXiv preprint arXiv:1711.00695}, 2017.

\bibitem[Dua and Graff(2017)]{Dua:2019}
Dheeru Dua and Casey Graff.
\newblock {UCI} machine learning repository, 2017.
\newblock URL \url{http://archive.ics.uci.edu/ml}.

\bibitem[Dulac-Arnold et~al.(2015)Dulac-Arnold, Evans, van Hasselt, Sunehag,
  Lillicrap, Hunt, Mann, Weber, Degris, and Coppin]{dulac2015deep}
Gabriel Dulac-Arnold, Richard Evans, Hado van Hasselt, Peter Sunehag, Timothy
  Lillicrap, Jonathan Hunt, Timothy Mann, Theophane Weber, Thomas Degris, and
  Ben Coppin.
\newblock Deep reinforcement learning in large discrete action spaces.
\newblock \emph{arXiv preprint arXiv:1512.07679}, 2015.

\bibitem[Durkan et~al.(2019)Durkan, Bekasov, Murray, and
  Papamakarios]{durkan2019neural}
Conor Durkan, Artur Bekasov, Iain Murray, and George Papamakarios.
\newblock Neural spline flows.
\newblock \emph{Advances in neural information processing systems}, 32, 2019.

\bibitem[{FICO Community}(2018)]{fico}
{FICO Community}.
\newblock Explainable machine learning challenge, 2018.
\newblock URL
  \url{https://community.fico.com/s/explainable-machine-learning-challenge}.

\bibitem[Friedman and Popescu(2008)]{friedman2008predictive}
Jerome~H Friedman and Bogdan~E Popescu.
\newblock Predictive learning via rule ensembles.
\newblock \emph{The annals of applied statistics}, 2\penalty0 (3):\penalty0
  916--954, 2008.

\bibitem[Fu et~al.(2013)Fu, Zhu, and Li]{fu2013survey}
Yifan Fu, Xingquan Zhu, and Bin Li.
\newblock A survey on instance selection for active learning.
\newblock \emph{Knowledge and information systems}, 35\penalty0 (2):\penalty0
  249--283, 2013.

\bibitem[Gal and Ghahramani(2016)]{gal2016dropout}
Yarin Gal and Zoubin Ghahramani.
\newblock Dropout as a bayesian approximation: Representing model uncertainty
  in deep learning.
\newblock In \emph{international conference on machine learning}, pages
  1050--1059. PMLR, 2016.

\bibitem[Goldberger et~al.(2000)Goldberger, Amaral, Glass, Hausdorff, Ivanov,
  Mark, Mietus, Moody, Peng, and Stanley]{goldberger2000physiobank}
Ary~L Goldberger, Luis~AN Amaral, Leon Glass, Jeffrey~M Hausdorff, Plamen~Ch
  Ivanov, Roger~G Mark, Joseph~E Mietus, George~B Moody, Chung-Kang Peng, and
  H~Eugene Stanley.
\newblock Physiobank, physiotoolkit, and physionet: components of a new
  research resource for complex physiologic signals.
\newblock \emph{circulation}, 101\penalty0 (23):\penalty0 e215--e220, 2000.

\bibitem[Gong et~al.(2019)Gong, Tschiatschek, Nowozin, Turner,
  Hern{\'a}ndez-Lobato, and Zhang]{gong2019icebreaker}
Wenbo Gong, Sebastian Tschiatschek, Sebastian Nowozin, Richard~E Turner,
  Jos{\'e}~Miguel Hern{\'a}ndez-Lobato, and Cheng Zhang.
\newblock Icebreaker: Element-wise efficient information acquisition with a
  bayesian deep latent gaussian model.
\newblock In \emph{Advances in Neural Information Processing Systems}, pages
  14820--14831, 2019.

\bibitem[Gu et~al.(2016)Gu, Lillicrap, Sutskever, and Levine]{gu2016continuous}
Shixiang Gu, Timothy Lillicrap, Ilya Sutskever, and Sergey Levine.
\newblock Continuous deep q-learning with model-based acceleration.
\newblock In \emph{International Conference on Machine Learning}, pages
  2829--2838, 2016.

\bibitem[He et~al.(2012)He, Eisner, and Daume]{he2012imitation}
He~He, Jason Eisner, and Hal Daume.
\newblock Imitation learning by coaching.
\newblock In \emph{Advances in Neural Information Processing Systems}, pages
  3149--3157, 2012.

\bibitem[He et~al.(2016)He, Mineiro, and Karampatziakis]{he2016active}
He~He, Paul Mineiro, and Nikos Karampatziakis.
\newblock Active information acquisition.
\newblock \emph{arXiv preprint arXiv:1602.02181}, 2016.

\bibitem[Heess et~al.(2015)Heess, Wayne, Silver, Lillicrap, Erez, and
  Tassa]{heess2015learning}
Nicolas Heess, Gregory Wayne, David Silver, Timothy Lillicrap, Tom Erez, and
  Yuval Tassa.
\newblock Learning continuous control policies by stochastic value gradients.
\newblock In \emph{Advances in Neural Information Processing Systems}, pages
  2944--2952, 2015.

\bibitem[Hendrycks et~al.(2019)Hendrycks, Mazeika, and
  Dietterich]{hendrycks2018deep}
Dan Hendrycks, Mantas Mazeika, and Thomas Dietterich.
\newblock Deep anomaly detection with outlier exposure.
\newblock In \emph{International Conference on Learning Representations}, 2019.
\newblock URL \url{https://openreview.net/forum?id=HyxCxhRcY7}.

\bibitem[Islam et~al.(2021)Islam, Eberle, Ghafoor, and
  Ahmed]{islam2021explainable}
Sheikh~Rabiul Islam, William Eberle, Sheikh~Khaled Ghafoor, and Mohiuddin
  Ahmed.
\newblock Explainable artificial intelligence approaches: A survey.
\newblock \emph{arXiv preprint arXiv:2101.09429}, 2021.

\bibitem[Ivanov et~al.(2018)Ivanov, Figurnov, and
  Vetrov]{ivanov2018variational}
Oleg Ivanov, Michael Figurnov, and Dmitry Vetrov.
\newblock Variational autoencoder with arbitrary conditioning.
\newblock \emph{arXiv preprint arXiv:1806.02382}, 2018.

\bibitem[Jaini et~al.(2019)Jaini, Selby, and Yu]{JainiSY19}
P.~Jaini, K.~Selby, and Y.~Yu.
\newblock Sum-of-squares polynomial flow.
\newblock In \emph{International Conference on Machine Learning {(ICML)}},
  2019.

\bibitem[Jayaraman and Grauman(2018)]{jayaraman2018learning}
Dinesh Jayaraman and Kristen Grauman.
\newblock Learning to look around: Intelligently exploring unseen environments
  for unknown tasks.
\newblock In \emph{Proceedings of the IEEE Conference on Computer Vision and
  Pattern Recognition}, pages 1238--1247, 2018.

\bibitem[Kingma and Dhariwal(2018)]{kingma2018glow}
Durk~P Kingma and Prafulla Dhariwal.
\newblock Glow: Generative flow with invertible 1x1 convolutions.
\newblock In \emph{Advances in Neural Information Processing Systems}, pages
  10215--10224, 2018.

\bibitem[Konyushkova et~al.(2017)Konyushkova, Sznitman, and
  Fua]{konyushkova2017learning}
Ksenia Konyushkova, Raphael Sznitman, and Pascal Fua.
\newblock Learning active learning from data.
\newblock In \emph{Advances in Neural Information Processing Systems}, pages
  4225--4235, 2017.

\bibitem[Kumar et~al.(2018)Kumar, Eslami, Rezende, Garnelo, Viola, Lockhart,
  and Shanahan]{kumar2018consistent}
Ananya Kumar, SM~Eslami, Danilo~J Rezende, Marta Garnelo, Fabio Viola, Edward
  Lockhart, and Murray Shanahan.
\newblock Consistent generative query networks.
\newblock \emph{arXiv preprint arXiv:1807.02033}, 2018.

\bibitem[Kumar et~al.(2019)Kumar, Liang, and Ma]{kumar2019verified}
Ananya Kumar, Percy~S Liang, and Tengyu Ma.
\newblock Verified uncertainty calibration.
\newblock \emph{Advances in Neural Information Processing Systems}, 32, 2019.

\bibitem[Kybic(2007)]{kybic2007high}
Jan Kybic.
\newblock High-dimensional entropy estimation for finite accuracy data: R-nn
  entropy estimator.
\newblock In \emph{Biennial International Conference on Information Processing
  in Medical Imaging}, pages 569--580. Springer, 2007.

\bibitem[Lakshminarayanan et~al.(2017)Lakshminarayanan, Pritzel, and
  Blundell]{lakshminarayanan2017simple}
Balaji Lakshminarayanan, Alexander Pritzel, and Charles Blundell.
\newblock Simple and scalable predictive uncertainty estimation using deep
  ensembles.
\newblock \emph{Advances in neural information processing systems}, 30, 2017.

\bibitem[Larochelle and Murray(2011)]{larochelle2011neural}
Hugo Larochelle and Iain Murray.
\newblock The neural autoregressive distribution estimator.
\newblock In \emph{Proceedings of the Fourteenth International Conference on
  Artificial Intelligence and Statistics}, pages 29--37, 2011.

\bibitem[LeCun(1998)]{lecun1998mnist}
Yann LeCun.
\newblock The mnist database of handwritten digits.
\newblock \emph{http://yann.lecun.com/exdb/mnist/}, 1998.

\bibitem[Li et~al.(2017)Li, Cheng, Wang, Morstatter, Trevino, Tang, and
  Liu]{li2017feature}
Jundong Li, Kewei Cheng, Suhang Wang, Fred Morstatter, Robert~P Trevino,
  Jiliang Tang, and Huan Liu.
\newblock Feature selection: A data perspective.
\newblock \emph{ACM Computing Surveys (CSUR)}, 50\penalty0 (6):\penalty0 1--45,
  2017.

\bibitem[Li and Oliva(2021)]{li2021active}
Yang Li and Junier Oliva.
\newblock Active feature acquisition with generative surrogate models.
\newblock In \emph{International Conference on Machine Learning}, pages
  6450--6459. PMLR, 2021.

\bibitem[Li et~al.(2020{\natexlab{a}})Li, Akbar, and Oliva]{li2019flow}
Yang Li, Shoaib Akbar, and Junier Oliva.
\newblock Acflow: Flow models for arbitrary conditional likelihoods.
\newblock In \emph{International Conference on Machine Learning}, pages
  5831--5841. PMLR, 2020{\natexlab{a}}.

\bibitem[Li et~al.(2020{\natexlab{b}})Li, Gao, and Oliva]{li2019forest}
Yang Li, Tianxiang Gao, and Junier Oliva.
\newblock A forest from the trees: Generation through neighborhoods.
\newblock In \emph{Proceedings of the AAAI Conference on Artificial
  Intelligence}, volume~34, pages 4755--4762, 2020{\natexlab{b}}.

\bibitem[Li(2017)]{li2017deep}
Yuxi Li.
\newblock Deep reinforcement learning: An overview.
\newblock \emph{arXiv preprint arXiv:1701.07274}, 2017.

\bibitem[Ling et~al.(2004)Ling, Yang, Wang, and Zhang]{ling2004decision}
Charles~X Ling, Qiang Yang, Jianning Wang, and Shichao Zhang.
\newblock Decision trees with minimal costs.
\newblock In \emph{Proceedings of the twenty-first international conference on
  Machine learning}, page~69, 2004.

\bibitem[Liu et~al.(2020)Liu, Lin, Padhy, Tran, Bedrax~Weiss, and
  Lakshminarayanan]{liu2020simple}
Jeremiah Liu, Zi~Lin, Shreyas Padhy, Dustin Tran, Tania Bedrax~Weiss, and
  Balaji Lakshminarayanan.
\newblock Simple and principled uncertainty estimation with deterministic deep
  learning via distance awareness.
\newblock \emph{Advances in Neural Information Processing Systems},
  33:\penalty0 7498--7512, 2020.

\bibitem[Ma et~al.(2019)Ma, Tschiatschek, Palla, Hernandez-Lobato, Nowozin, and
  Zhang]{ma2018eddi}
Chao Ma, Sebastian Tschiatschek, Konstantina Palla, Jose~Miguel
  Hernandez-Lobato, Sebastian Nowozin, and Cheng Zhang.
\newblock Eddi: Efficient dynamic discovery of high-value information with
  partial vae.
\newblock In \emph{International Conference on Machine Learning}, pages
  4234--4243. PMLR, 2019.

\bibitem[Madumal et~al.(2020)Madumal, Miller, Sonenberg, and
  Vetere]{madumal2020explainable}
Prashan Madumal, Tim Miller, Liz Sonenberg, and Frank Vetere.
\newblock Explainable reinforcement learning through a causal lens.
\newblock In \emph{Proceedings of the AAAI conference on artificial
  intelligence}, volume~34, pages 2493--2500, 2020.

\bibitem[Mahmood et~al.(2021)Mahmood, Oliva, and Styner]{mahmood2021multiscale}
Ahsan Mahmood, Junier Oliva, and Martin~Andreas Styner.
\newblock Multiscale score matching for out-of-distribution detection.
\newblock In \emph{International Conference on Learning Representations}, 2021.
\newblock URL \url{https://openreview.net/forum?id=xoHdgbQJohv}.

\bibitem[Majeed and Hutter(2021)]{majeed2020exact}
Sultan~J Majeed and Marcus Hutter.
\newblock Exact reduction of huge action spaces in general reinforcement
  learning.
\newblock In \emph{Proceedings of the AAAI Conference on Artificial
  Intelligence}, volume~35, pages 8874--8883, 2021.

\bibitem[Miao and Niu(2016)]{miao2016survey}
Jianyu Miao and Lingfeng Niu.
\newblock A survey on feature selection.
\newblock \emph{Procedia Computer Science}, 91:\penalty0 919--926, 2016.

\bibitem[Minsky(1961)]{minsky1961steps}
Marvin Minsky.
\newblock Steps toward artificial intelligence.
\newblock \emph{Proceedings of the IRE}, 49\penalty0 (1):\penalty0 8--30, 1961.

\bibitem[Morningstar et~al.(2021)Morningstar, Ham, Gallagher, Lakshminarayanan,
  Alemi, and Dillon]{morningstar2021density}
Warren Morningstar, Cusuh Ham, Andrew Gallagher, Balaji Lakshminarayanan, Alex
  Alemi, and Joshua Dillon.
\newblock Density of states estimation for out of distribution detection.
\newblock In \emph{International Conference on Artificial Intelligence and
  Statistics}, pages 3232--3240. PMLR, 2021.

\bibitem[Nalisnick et~al.(2019{\natexlab{a}})Nalisnick, Matsukawa, Teh, Gorur,
  and Lakshminarayanan]{nalisnick2018do}
Eric Nalisnick, Akihiro Matsukawa, Yee~Whye Teh, Dilan Gorur, and Balaji
  Lakshminarayanan.
\newblock Do deep generative models know what they don't know?
\newblock In \emph{International Conference on Learning Representations},
  2019{\natexlab{a}}.
\newblock URL \url{https://openreview.net/forum?id=H1xwNhCcYm}.

\bibitem[Nalisnick et~al.(2019{\natexlab{b}})Nalisnick, Matsukawa, Teh, and
  Lakshminarayanan]{nalisnick2019detecting}
Eric Nalisnick, Akihiro Matsukawa, Yee~Whye Teh, and Balaji Lakshminarayanan.
\newblock Detecting out-of-distribution inputs to deep generative models using
  typicality.
\newblock \emph{arXiv preprint arXiv:1906.02994}, 2019{\natexlab{b}}.

\bibitem[Nan et~al.(2014)Nan, Wang, Trapeznikov, and Saligrama]{nan2014fast}
Feng Nan, Joseph Wang, Kirill Trapeznikov, and Venkatesh Saligrama.
\newblock Fast margin-based cost-sensitive classification.
\newblock In \emph{2014 IEEE International Conference on Acoustics, Speech and
  Signal Processing (ICASSP)}, pages 2952--2956. IEEE, 2014.

\bibitem[Ng et~al.(1999)Ng, Harada, and Russell]{ng1999policy}
Andrew~Y Ng, Daishi Harada, and Stuart Russell.
\newblock Policy invariance under reward transformations: Theory and
  application to reward shaping.
\newblock In \emph{Icml}, volume~99, pages 278--287, 1999.

\bibitem[Norouzi et~al.(2020)Norouzi, Fleet, and Norouzi]{norouzi2020exemplar}
Sajad Norouzi, David~J Fleet, and Mohammad Norouzi.
\newblock Exemplar vae: Linking generative models, nearest neighbor retrieval,
  and data augmentation.
\newblock \emph{Advances in Neural Information Processing Systems},
  33:\penalty0 8753--8764, 2020.

\bibitem[Oliva et~al.(2018)Oliva, Dubey, Zaheer, Poczos, Salakhutdinov, Xing,
  and Schneider]{oliva2018transformation}
Junier Oliva, Avinava Dubey, Manzil Zaheer, Barnabas Poczos, Ruslan
  Salakhutdinov, Eric Xing, and Jeff Schneider.
\newblock Transformation autoregressive networks.
\newblock In \emph{International Conference on Machine Learning}, pages
  3898--3907. PMLR, 2018.

\bibitem[Ovadia et~al.(2019)Ovadia, Fertig, Ren, Nado, Sculley, Nowozin,
  Dillon, Lakshminarayanan, and Snoek]{ovadia2019can}
Yaniv Ovadia, Emily Fertig, Jie Ren, Zachary Nado, David Sculley, Sebastian
  Nowozin, Joshua Dillon, Balaji Lakshminarayanan, and Jasper Snoek.
\newblock Can you trust your model's uncertainty? evaluating predictive
  uncertainty under dataset shift.
\newblock \emph{Advances in neural information processing systems}, 32, 2019.

\bibitem[Papamakarios et~al.(2017)Papamakarios, Pavlakou, and
  Murray]{papamakarios2017masked}
George Papamakarios, Theo Pavlakou, and Iain Murray.
\newblock Masked autoregressive flow for density estimation.
\newblock \emph{Advances in neural information processing systems}, 30, 2017.

\bibitem[Papamakarios et~al.(2021)Papamakarios, Nalisnick, Rezende, Mohamed,
  and Lakshminarayanan]{papamakarios2021normalizing}
George Papamakarios, Eric Nalisnick, Danilo~Jimenez Rezende, Shakir Mohamed,
  and Balaji Lakshminarayanan.
\newblock Normalizing flows for probabilistic modeling and inference.
\newblock \emph{Journal of Machine Learning Research}, 22\penalty0
  (57):\penalty0 1--64, 2021.

\bibitem[Pong et~al.(2018)Pong, Gu, Dalal, and Levine]{pong2018temporal}
Vitchyr Pong, Shixiang Gu, Murtaza Dalal, and Sergey Levine.
\newblock Temporal difference models: Model-free deep rl for model-based
  control.
\newblock In \emph{International Conference on Learning Representations}, 2018.

\bibitem[Puiutta and Veith(2020)]{puiutta2020explainable}
Erika Puiutta and Eric Veith.
\newblock Explainable reinforcement learning: A survey.
\newblock In \emph{International cross-domain conference for machine learning
  and knowledge extraction}, pages 77--95. Springer, 2020.

\bibitem[Ren et~al.(2019)Ren, Liu, Fertig, Snoek, Poplin, Depristo, Dillon, and
  Lakshminarayanan]{ren2019likelihood}
Jie Ren, Peter~J Liu, Emily Fertig, Jasper Snoek, Ryan Poplin, Mark Depristo,
  Joshua Dillon, and Balaji Lakshminarayanan.
\newblock Likelihood ratios for out-of-distribution detection.
\newblock \emph{Advances in Neural Information Processing Systems}, 32, 2019.

\bibitem[Ribeiro et~al.(2016)Ribeiro, Singh, and Guestrin]{ribeiro2016should}
Marco~Tulio Ribeiro, Sameer Singh, and Carlos Guestrin.
\newblock " why should i trust you?" explaining the predictions of any
  classifier.
\newblock In \emph{Proceedings of the 22nd ACM SIGKDD international conference
  on knowledge discovery and data mining}, pages 1135--1144, 2016.

\bibitem[R{\"u}ckstie{\ss} et~al.(2011)R{\"u}ckstie{\ss}, Osendorfer, and
  van~der Smagt]{ruckstiess2011sequential}
Thomas R{\"u}ckstie{\ss}, Christian Osendorfer, and Patrick van~der Smagt.
\newblock Sequential feature selection for classification.
\newblock In \emph{Australasian Joint Conference on Artificial Intelligence},
  pages 132--141. Springer, 2011.

\bibitem[Schulman et~al.(2017)Schulman, Wolski, Dhariwal, Radford, and
  Klimov]{schulman2017proximal}
John Schulman, Filip Wolski, Prafulla Dhariwal, Alec Radford, and Oleg Klimov.
\newblock Proximal policy optimization algorithms.
\newblock \emph{arXiv preprint arXiv:1707.06347}, 2017.

\bibitem[Shapley(1953)]{shapley1953value}
Lloyd~S Shapley.
\newblock A value for n-person games, contributions to the theory of games, 2,
  307--317, 1953.

\bibitem[Shim et~al.(2018)Shim, Hwang, and Yang]{shim2018joint}
Hajin Shim, Sung~Ju Hwang, and Eunho Yang.
\newblock Joint active feature acquisition and classification with
  variable-size set encoding.
\newblock In \emph{Advances in neural information processing systems}, pages
  1368--1378, 2018.

\bibitem[Song and Ermon(2019)]{song2019generative}
Yang Song and Stefano Ermon.
\newblock Generative modeling by estimating gradients of the data distribution.
\newblock \emph{Advances in Neural Information Processing Systems}, 32, 2019.

\bibitem[Strauss and Oliva(2021)]{strauss2021arbitrary}
Ryan Strauss and Junier~B Oliva.
\newblock Arbitrary conditional distributions with energy.
\newblock \emph{Advances in Neural Information Processing Systems}, 34, 2021.

\bibitem[Strauss and Oliva(2022)]{strauss2022any}
Ryan~R Strauss and Junier~B Oliva.
\newblock Any variational autoencoder can do arbitrary conditioning.
\newblock \emph{arXiv preprint arXiv:2201.12414}, 2022.

\bibitem[Sutton(1988)]{sutton1988learning}
Richard~S Sutton.
\newblock Learning to predict by the methods of temporal differences.
\newblock \emph{Machine learning}, 3\penalty0 (1):\penalty0 9--44, 1988.

\bibitem[Tjoa and Guan(2020)]{tjoa2020survey}
Erico Tjoa and Cuntai Guan.
\newblock A survey on explainable artificial intelligence (xai): Toward medical
  xai.
\newblock \emph{IEEE transactions on neural networks and learning systems},
  32\penalty0 (11):\penalty0 4793--4813, 2020.

\bibitem[van Amersfoort et~al.(2021)van Amersfoort, Smith, Jesson, Key, and
  Gal]{van2021improving}
Joost van Amersfoort, Lewis Smith, Andrew Jesson, Oscar Key, and Yarin Gal.
\newblock Improving deterministic uncertainty estimation in deep learning for
  classification and regression.
\newblock \emph{arXiv preprint arXiv:2102.11409}, 2021.

\bibitem[Van~den Oord et~al.(2016)Van~den Oord, Kalchbrenner, Espeholt,
  Vinyals, Graves, et~al.]{van2016conditional}
Aaron Van~den Oord, Nal Kalchbrenner, Lasse Espeholt, Oriol Vinyals, Alex
  Graves, et~al.
\newblock Conditional image generation with pixelcnn decoders.
\newblock \emph{Advances in neural information processing systems}, 29, 2016.

\bibitem[Verma et~al.(2018)Verma, Murali, Singh, Kohli, and
  Chaudhuri]{verma2018programmatically}
Abhinav Verma, Vijayaraghavan Murali, Rishabh Singh, Pushmeet Kohli, and Swarat
  Chaudhuri.
\newblock Programmatically interpretable reinforcement learning.
\newblock In \emph{International Conference on Machine Learning}, pages
  5045--5054. PMLR, 2018.

\bibitem[Xiao et~al.(2017)Xiao, Rasul, and Vollgraf]{xiao2017fashion}
Han Xiao, Kashif Rasul, and Roland Vollgraf.
\newblock Fashion-mnist: a novel image dataset for benchmarking machine
  learning algorithms.
\newblock \emph{arXiv preprint arXiv:1708.07747}, 2017.

\bibitem[Yoo and Kweon(2019)]{yoo2019learning}
Donggeun Yoo and In~So Kweon.
\newblock Learning loss for active learning.
\newblock In \emph{Proceedings of the IEEE Conference on Computer Vision and
  Pattern Recognition}, pages 93--102, 2019.

\bibitem[Zannone et~al.(2019)Zannone, Hernandez~Lobato, Zhang, and
  Palla]{zannone2019odin}
Sara Zannone, Jose~Miguel Hernandez~Lobato, Cheng Zhang, and Konstantina Palla.
\newblock Odin: Optimal discovery of high-value information using model-based
  deep reinforcement learning.
\newblock In \emph{Real-world Sequential Decision Making Workshop, ICML}, June
  2019.

\bibitem[Zubek and Dietterich(2002)]{zubek2002pruning}
Valentina~Bayer Zubek and Thomas~G Dietterich.
\newblock Pruning improves heuristic search for cost-sensitive learning.
\newblock In \emph{ICML}, 2002.

\end{thebibliography}

\end{document}